\pdfoutput=1

\documentclass[11pt]{article}

\usepackage[dvipsnames,table,xcdraw]{xcolor}
\usepackage[preprint]{acl}

\usepackage{times}
\usepackage{latexsym}

\usepackage[T1]{fontenc}

\usepackage[utf8]{inputenc}

\usepackage{microtype}

\usepackage{inconsolata}

\usepackage{graphicx}
\usepackage{xspace}
\usepackage{booktabs}
\usepackage{multirow}
\usepackage{colortbl}
\usepackage{tcolorbox}
\usepackage{amssymb}
\usepackage{amsmath}
\usepackage{pifont}
\usepackage{bbding}
\usepackage{makecell}
\usepackage{scalerel}
\usepackage{highlight}
\usepackage{fontawesome5}
\usepackage{subcaption}
\usepackage{titletoc}

\newcommand{\ourbench}{\texttt{HumanEval-V}\xspace}
\newcommand{\grayline}{\rowcolor[gray]{.90}}

\newcommand{\logo}{\scalerel*{\includegraphics{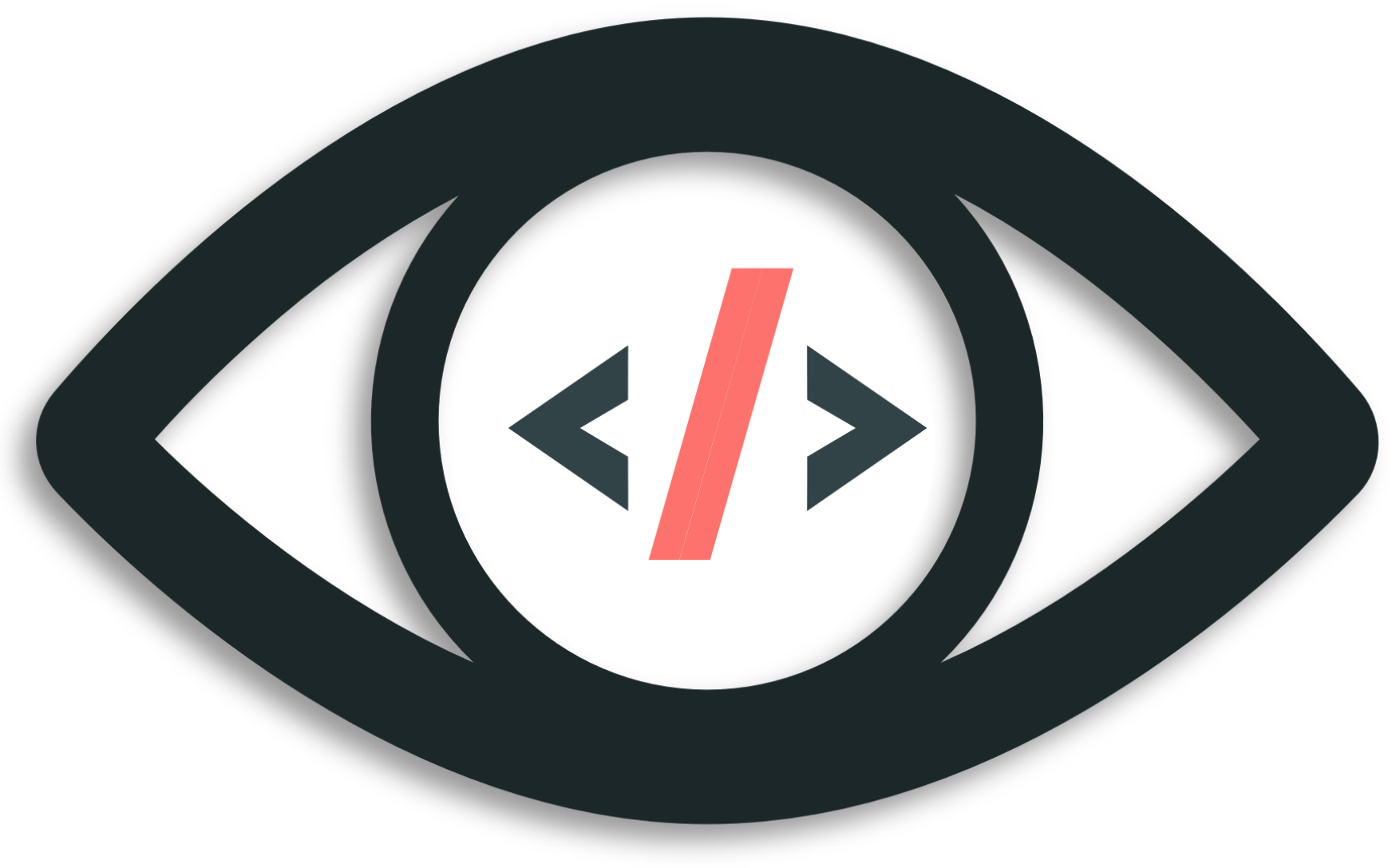}}{\textrm{\textbigcircle}}\xspace}

\title{\logo{} \ourbench: Benchmarking High-Level Visual Reasoning \\ with Complex Diagrams in Coding Tasks}

\author{
\centerline{Fengji Zhang\textsuperscript{*\ 1}
\quad Linquan Wu\textsuperscript{*\ 1}
\quad Huiyu Bai\textsuperscript{*\ 1}
\quad Guancheng Lin\textsuperscript{*\ 2}}\\[0.15cm]
\textbf{
\centerline{Xiao Li\textsuperscript{3}
\quad Xiao Yu\textsuperscript{4}
\quad Yue Wang\textsuperscript{5}
\quad Bei Chen \textsuperscript{5}
\quad Jacky Keung \textsuperscript{1}
}} \\[0.15cm]
\centerline{\textsuperscript{1}{City University of Hong Kong}
\quad \textsuperscript{2}{Wuhan University}} \\[0.15cm]
\centerline{\textsuperscript{3}{Tsinghua University}
\quad \textsuperscript{4}{Zhejiang University}
\quad \textsuperscript{5}{Rhymes AI}
} \\[0.15cm]
}

\makeatletter
\def\blfootnote{\gdef\@thefnmark{}\@footnotetext}
\makeatother

\begin{document}
\maketitle
\begin{abstract}
Understanding and reasoning over diagrams is a fundamental aspect of human intelligence. While Large Multimodal Models (LMMs) have demonstrated impressive capabilities across various tasks, existing benchmarks lack comprehensive evaluation of their diagram interpretation and reasoning abilities, particularly in coding contexts. We present \ourbench, a rigorous benchmark of human-annotated coding tasks that spans six task types and evaluates diverse visual reasoning capabilities. Each task features carefully crafted diagrams paired with function signatures and test cases, employing novel code generation tasks to thoroughly assess models' diagram comprehension. Through extensive experiments with 22 LMMs, we find that even top-performing models achieve modest success rates, with Claude 3.5 Sonnet reaching only 36.8\% pass@1, highlighting substantial room for improvement. Our analysis reveals that current LMMs struggle with spatial transformations, topological relationships, and dynamic patterns that humans find intuitive. These findings provide valuable insights for advancing LMMs' visual reasoning abilities. We have open-sourced our code and benchmark at \url{https://github.com/HumanEval-V/HumanEval-V-Benchmark}.
\end{abstract}

\blfootnote{\hspace{-1.85em}$^*$ denotes equal contribution.\\ Correspondence to \texttt{\href{mailto:fengji.zhang@my.cityu.edu.hk}{fengji.zhang@my.cityu.edu.hk}}}

\vspace{-1em}
\section{Introduction}

\vspace{-0.5em}
High-level intelligence, whether in humans or advanced AI systems, requires the ability to understand and reason over visual information represented in diagrams. Diagrams are essential in many domains, including science, engineering, and mathematics, as they serve as a powerful medium for abstracting and communicating complex data, relationships, and processes, encoding rich information in a visual and structured format. The abilities required to comprehend diagrams extend beyond simple pattern recognition; they necessitate sophisticated cognitive capabilities, including interpreting
transformation patterns, recognizing hierarchical structures, and integrating multiple visual cues such as arrows, symbols, and their relative positions to perform spatial or logical reasoning. The rapid development of Large Multimodal Models (LMMs) has led to the creation of various benchmarks designed to assess the alignment between LMMs' capabilities and human intelligence. However, there remains a significant gap in benchmarks that specifically evaluate the ability to understand and reason over complex diagrams.

\begin{figure}[t]
\centering
\includegraphics[width=\linewidth]{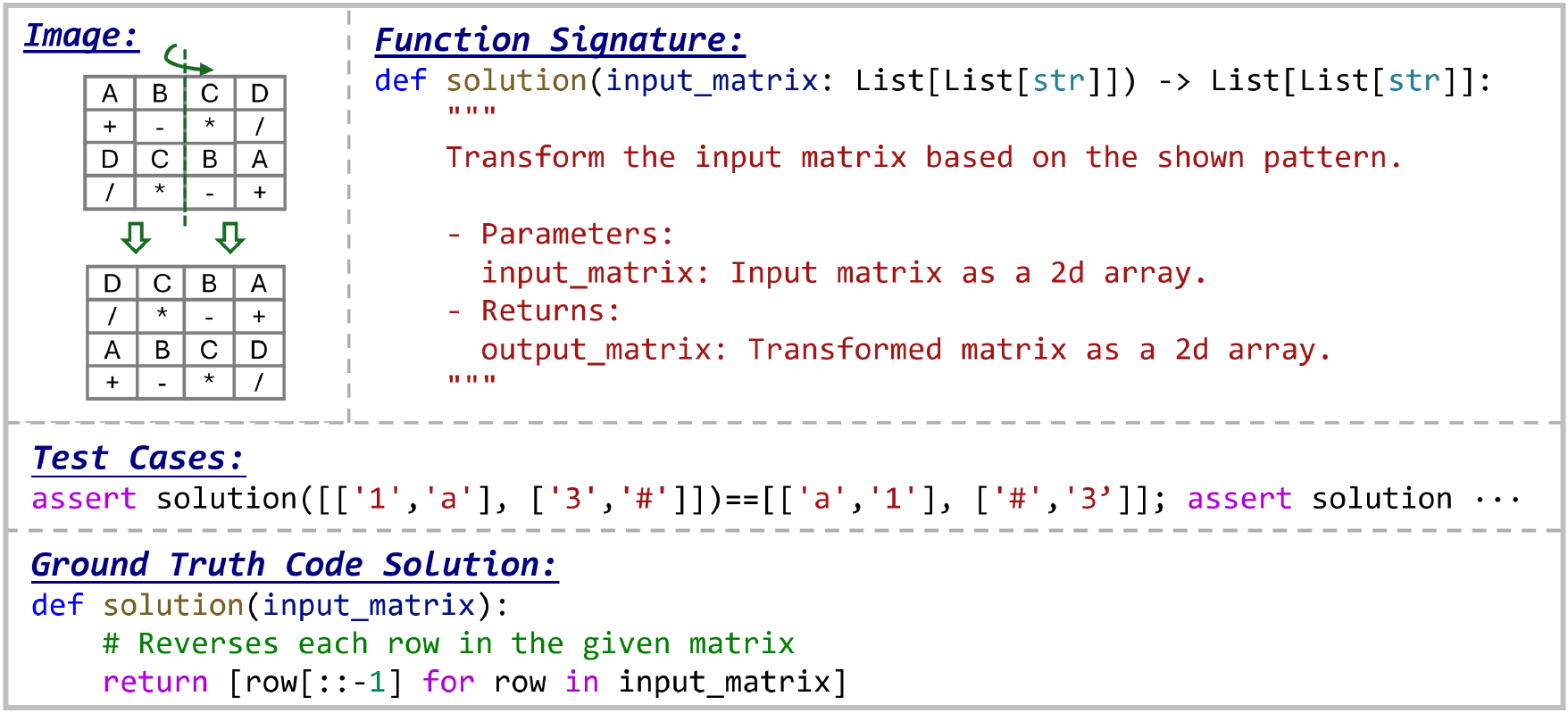}
\vspace{-1em}
\caption{A task example from \ourbench. LMMs are required to figure out the facts and patterns in the diagram and complete the function body.}
\label{fig:intro_example}
\end{figure}
\begin{figure*}[t]
\begin{minipage}[b]{0.6\linewidth}
    \centering
    \includegraphics[width=\linewidth]{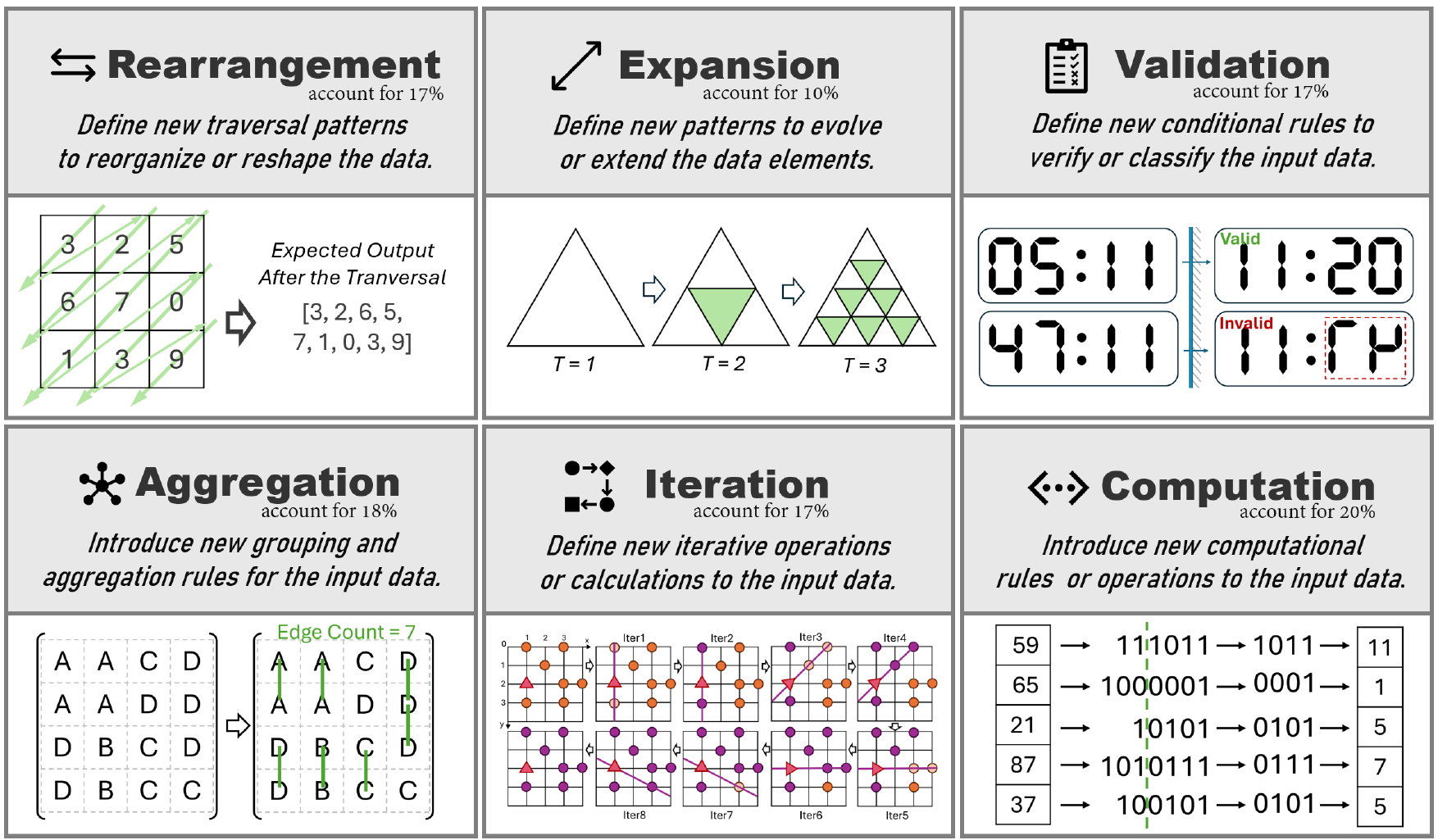}
    \vspace{-0.5em}
    \caption{The six task categories in \ourbench, along with their quantitative distribution and representative diagram examples.}
 \label{fig:task_types}

\end{minipage}
\hspace{0.2cm}
\begin{minipage}[b]{0.38\linewidth}
    \centering
    \includegraphics[width=\linewidth]{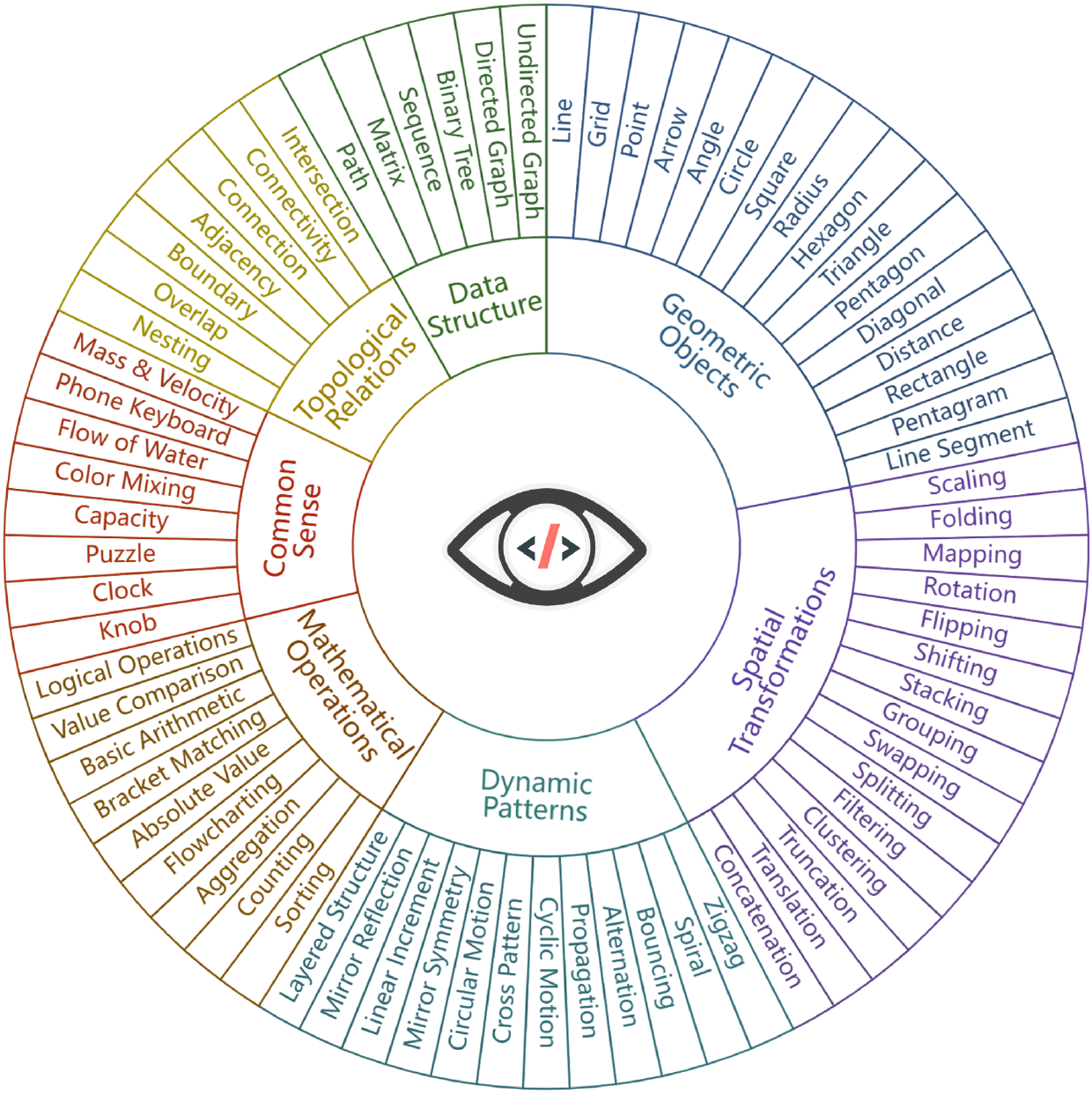}
    \caption{Core knowledge required for understanding diagrams in \ourbench.}
    \label{fig:diverse_labels}
\end{minipage}
\end{figure*}

Popular multimodal benchmarks, such as MMMU~\citep{yue2024mmmu}, MathVista~\citep{lu2023mathvista}, and ChartQA~\citep{masry2022chartqa}, focus primarily on scientific, mathematical, and chart-based analytical questions over various domains of images, testing LMMs' multidiscipline knowledge
rather than  diagram understanding. 
While abstract visual reasoning tasks~\cite{zhang2019raven, nie2020bongard} from IQ tests typically feature static patterns based on visual analogies or numerical inference over diagrams, they often lack the complexity and diversity in visual patterns and diagram types. This gap highlights the need for benchmarks that assess more intricate diagram reasoning abilities.  
The field of coding presents an underexplored opportunity, as developers frequently use various diagrams to illustrate data structures, algorithms, and problem constraints. A recent study, MMCode~\citep{li2024mmcode}, evaluated LMMs on coding problems with visual contexts directly crawled from competition platforms. However, competition coding problems often include comprehensive textual descriptions, making the visual information supplementary. Their results, which showed similar LMM performance with and without visual information, further underscore the gap in evaluating genuine diagram understanding abilities.

To address this gap, we introduce \ourbench, a novel benchmark designed to provide a focused evaluation of complex diagram understanding and reasoning abilities in programming contexts. Unlike MMCode, our benchmark is dedicated to assessing visual capabilities through a rigorous annotation pipeline that creates coding tasks capturing the essence of real-world problems. Each task features an \textit{indispensable}, \textit{self-explanatory} diagram with minimal textual clues, as demonstrated by our experiments where top LMMs failed all tasks without the provided diagrams.
\ourbench consists of 253 human-annotated coding tasks. Each task features (1) a diagram encoding the problem context, (2) a function signature defining the task's input-output structures, and (3) test cases to verify solution correctness. 
Figure~\ref{fig:intro_example} provides an example task, where the diagram illustrates spatial transformation patterns, requiring the model to comprehend fine-grained visual elements such as matrices, arrow directions, and spatially ordered data points. This task aligns with the ARC-AGI~\citep{chollet2019measure} benchmark in inferring transformation patterns from limited visual examples. However, unlike ARC’s matrix-formatted diagrams, \ourbench offers a more diverse and complex set of diagrams spanning six task types (Figure~\ref{fig:task_types}), demanding versatile capabilities (Figure~\ref{fig:diverse_labels}) for diagram understanding and reasoning. For a comparison of diagrams from existing benchmarks and ours, see Figures~\ref{fig:diagram_comparison} and \ref{fig:our_diagram_examples}.

Another novelty of \ourbench lies in using code generation tasks for evaluation instead of the multiple-choice or short-answer questions commonly used in other multimodal benchmarks. This approach offers compelling benefits: code generation is more challenging, requiring comprehensive logical thinking and visual understanding with minimal chance of correct guesses, and test cases could rigorously verify whether the model captures all critical visual information, rather than relying on similarity matching with ground truth. 
Additionally, we utilize a two-stage evaluation pipeline that supports LMMs with limited coding abilities by first prompting them to generate a structured diagram description summarizing the visual context, then using a more capable coder model to implement the solution, ensuring the evaluation prioritizes visual understanding over coding proficiency.

Through extensive experiments with 22 LMMs, we observe the following key findings:
(1) Our benchmark presents unique challenges not addressed by other multimodal benchmarks. The top-performing model, Claude 3.5 Sonnet, achieves 36.8\% pass@1, while the best open-weight model, Pixtral 124B, reaches 21.3\%.
(2) Current LMMs exhibit stronger vision-to-language alignment than vision-to-code. Their best performance occurs when they serve as diagram describers, with GPT-4o acting as the coder model.
(3) LMMs' performance can be further enhanced through sampling or iterative self-refinement. For instance, Claude 3.5 Sonnet achieves a 74.3\% pass rate with 100 samples, and it can reach 55.3\% pass@1 with four self-refining iterations based on test case execution feedback.
(4) Current LMMs still have difficulty understanding diagrams that are trivial for humans, particularly understanding spatial transformations, topological relationships, and dynamic patterns.

\section{Benchmark Construction}
\label{sec:benchmark_construction}

\vspace{-0.4em}
\paragraph{Task Definition:} As shown in Figure~\ref{fig:intro_example}, each coding task in \ourbench includes: (1) a single diagram $D$ providing visual context, (2) a Python function signature $\sigma$ with input parameters, return type, and brief instructions, and (3) a set of test cases $T = {t_1, t_2, \dots, t_n}$ to validate the correctness of the generated output $O$, a complete Python function produced by the LMM.

\vspace{-0.6em}
\paragraph{Annotation Standards:} We establish rigorous standards to ensure high-quality coding tasks: (1) The visual context must be essential for solving the task, with all relevant information contained in a single image; (2) Tasks should be designed around the visual context with minimal textual description; (3) Unnecessary programming complexities, such as recursion, intricate constraints, or complex data structures, should be avoided.

\begin{figure}[t]
\centering
\includegraphics[width=\linewidth]{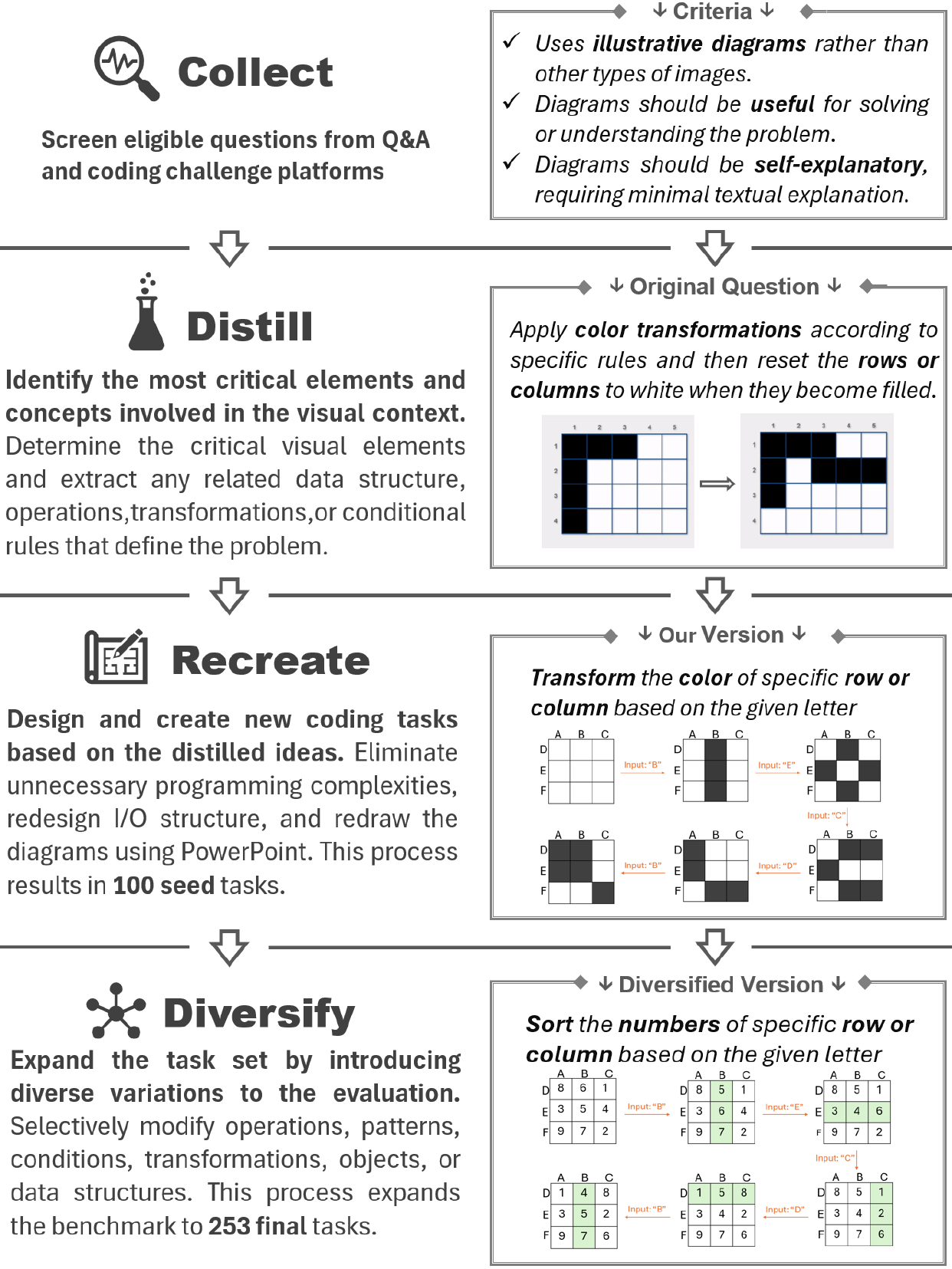}
\vspace{-0.5em}
\caption{\ourbench task construction pipeline.}
\label{fig:bench_construction}
\end{figure}

\vspace{-0.6em}
\paragraph{Task Annotation Pipeline:} We define the task annotation pipeline with \textit{four steps} as in Figure~\ref{fig:bench_construction}.

\textit{In the \textbf{first} step}, we collect a large set of coding problems from prominent Q\&A and coding challenge platforms that incorporate images, then screen them to exclude questions that (1) require specific programming frameworks or libraries, or contain images that (2) are not illustrative diagrams, (3) provide no useful information for solving the problem, or (4) require extensive textual context for interpretation.
\textit{For the \textbf{second} step}, we distill the screened problems to identify critical visual elements, along with the data structures, operations, transformations, or conditional rules involved, categorizing them into six task types (Figure~\ref{fig:task_types}) and outlining the key capabilities needed for diagram understanding.
\textit{For the \textbf{third} step}, we design new coding tasks based on these distilled ideas, eliminating unnecessary complexities, refining input/output structures and function signatures, and creating visual objects and layouts using basic shapes and a consistent color scheme in PowerPoint. We also generate tailored test cases, resulting in \textbf{\textit{100}} newly crafted \textit{seed tasks} after excluding tasks with design or formulation challenges.
\textit{For the \textbf{fourth} step}, we expand the task set by diversifying the seed tasks using a hybrid approach with GPT-4o. GPT-4o identifies relevant capability aspects for each seed task and suggests modifications involving new spatial transformations, mathematical operations, dynamic patterns, or variations in data structures and object attributes. Human annotators then refine and annotate these new tasks and diagrams, creating 0 to 2 diversified versions per seed task based on complexity, culminating in \textbf{\textit{253}} tasks in \ourbench and finalizing the capability aspects shown in Figure~\ref{fig:diverse_labels}. 
Further details and examples of the data collection and diversification processes are provided in Appendix~\ref{appendix:data_collection} and Appendix~\ref{appendix:task_annotation}.

\vspace{-0.8em}
\paragraph{Quality Assurance:} Our annotation team comprises four experienced programmers, each with over four years of Python programming experience. Initially, each annotator independently annotates their assigned tasks following pre-defined guidelines. Subsequently, all annotators review each other's work by annotating ground truth code solutions and diagram descriptions to ensure tasks are visually grounded, solvable with the provided information, and free of design or conceptual errors. Any identified issues are resolved collaboratively, with tasks finalized only after consensus is reached. Additionally, one annotator ensures consistent formatting and style across all visual representations and coding tasks. Each annotator contributes over 200 hours to the annotation process.

\begin{figure}[t]
    \centering
    \begin{subfigure}{0.48\linewidth}
        \centering
        \includegraphics[width=\linewidth]{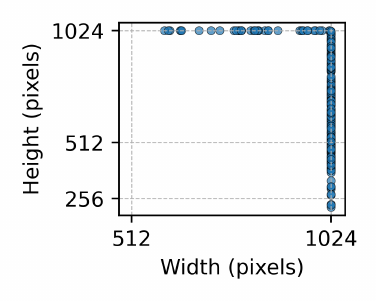}
        \caption{Diagram Size}
    \end{subfigure}
    \begin{subfigure}{0.48\linewidth}
        \centering
        \includegraphics[width=\linewidth]{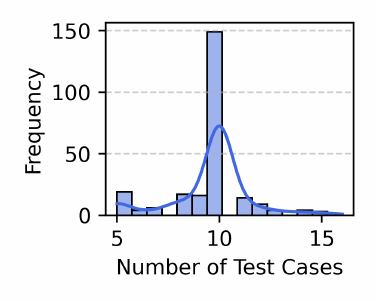}
        \caption{Test Cases Count}
    \end{subfigure}
    \begin{subfigure}{0.48\linewidth}
        \centering
        \includegraphics[width=\linewidth]{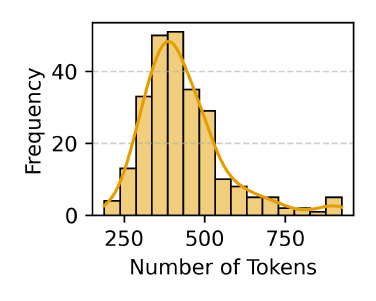}
        \caption{Description Length}
    \end{subfigure}
    \begin{subfigure}{0.48\linewidth}
        \centering
        \includegraphics[width=\linewidth]{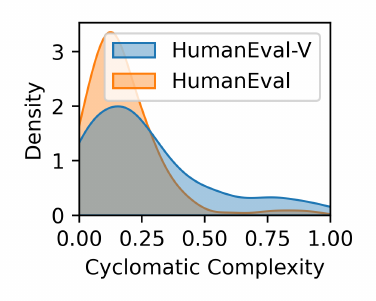}
        \caption{Coding Complexity}
    \end{subfigure}
    \caption{Distribution analysis of the benchmark data.}
    \label{fig:benchmark_statistics}
\end{figure}

\vspace{-0.6em}
\paragraph{Benchmark Statistics:} 
To further demonstrate the quality of our benchmark, we conduct statistical analyses on several key aspects and present the distribution of these statistics in Figure~\ref{fig:benchmark_statistics}. First, we strictly control diagram sizes, capping the maximum width or height at 1024 pixels to eliminate the need for high-resolution perception. Second, each task includes at least five test cases, with the majority containing ten, ensuring full statement and branch coverage over human-annotated code solutions. Third, the token length of human-annotated diagram descriptions is predominantly around 400 (measured using tiktoken~\citep{tiktoken}), demonstrating that our diagrams encapsulate rich visual context. Lastly, our human-annotated code solutions exhibit cyclomatic complexity~\citep{gill1991cyclomatic} levels comparable to HumanEval~\citep{chen2021evaluating}, a widely used coding benchmark designed for entry-level programming tasks.

\vspace{-0.3em}
\section{Benchmarking Setup}
\label{sec:benchmark_setup}
\begin{figure}[t]
\centering
\includegraphics[width=\linewidth]{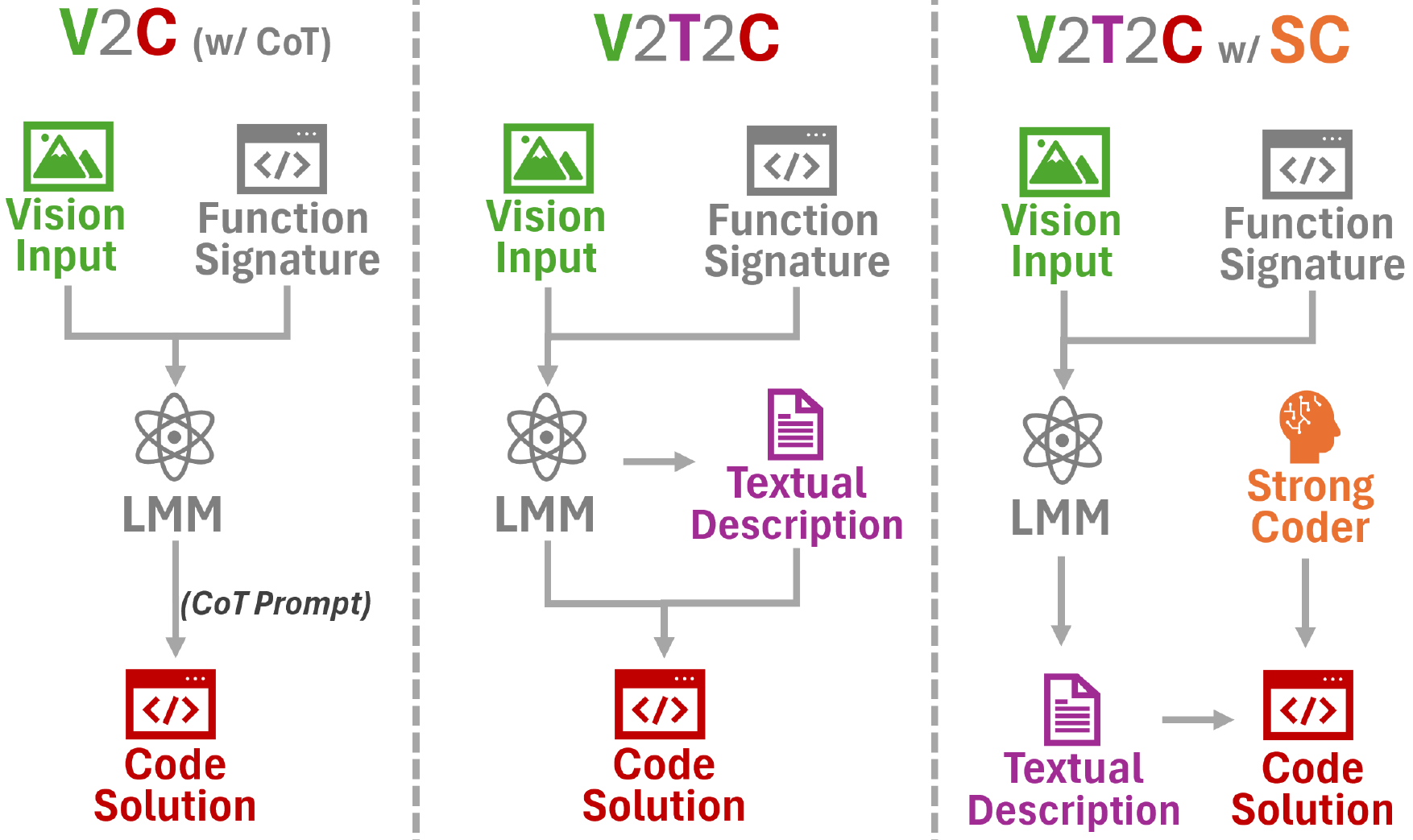}
\vspace{-0.5em}
\caption{Evaluation pipelines employed in the experiments: (1) Direct translation of visual context into code (V2C), (2) with an optional Chain-of-Thought prompt (V2C w/ CoT); (3) Translation of visual context into a textual description, which is then processed to generate code (V2T2C); and (4) A variant of the third pipeline, where a stronger coder model is used to generate the code solution (V2T2C w/ SC).
}
\label{fig:evaluation_pipeline}
\end{figure}

\begin{table*}[t]
  \centering
  \begin{minipage}[b]{0.77\linewidth}
  \scalebox{0.81}{
      \begin{tabular}{l|cc|ll|ll|ll}
      \toprule
      \multicolumn{1}{c}{\textbf{Models}}& \multicolumn{2}{c}{{\textbf{V2C}}} & \multicolumn{2}{c}{{\textbf{V2C w/ CoT}}} & \multicolumn{2}{c}{\textbf{V2T2C}} & \multicolumn{2}{c}{\textbf{V2T2C w/ \small{GPT-4o}}} \\
      \cmidrule(lr){1-1}
      \cmidrule(lr){2-3}
      \cmidrule(lr){4-5}
      \cmidrule(lr){6-7}
      \cmidrule(lr){8-9}
      \multicolumn{1}{c}{pass@$k$} & \multicolumn{1}{c}{$k$=1} & \multicolumn{1}{c}{$k$=3} & \multicolumn{1}{c}{$k$=1} & \multicolumn{1}{c}{$k$=3} & \multicolumn{1}{c}{$k$=1} & \multicolumn{1}{c}{$k$=3} & \multicolumn{1}{c}{$k$=1} & \multicolumn{1}{c}{$k$=3~\faMedal} \\
      \midrule
      \grayline \multicolumn{9}{c}{Proprietary LMMs} \\
      Claude 3.5 Sonnet & \textbf{28.1} & \textbf{37.9} & \textbf{36.8}\gradient{8.7} & \textbf{47.9}\gradient{10.0} & \textbf{33.2}\gradient{5.1} & \textbf{43.6}\gradient{5.7} & \textbf{31.6}\gradient{3.5} & \textbf{43.7}\gradient{5.8} \\
      GPT-4o  & \underline{24.1} & \underline{33.8} & \underline{27.7}\gradient{3.6} & \underline{40.0}\gradient{6.2} & 26.5\gradient{2.4} & \underline{40.5}\gradient{6.7} & 26.5\gradient{2.4} & \underline{40.5}\gradient{6.7} \\
      Gemini 1.5 Pro & 23.3 & 26.9 & 22.9\gradientb{0.4} & 34.1\gradient{7.2} & \underline{28.5}\gradient{5.2} & 36.4\gradient{9.5} & \underline{26.9}\gradient{3.6} & 37.3\gradient{10.4} \\
      Gemini 1.5 Flash & 15.4 & 20.5 & 17.4\gradient{2.0} & 24.9\gradient{4.4} & 15.8\gradient{0.4} & 22.0\gradient{1.5} & 18.6\gradient{3.2} & 27.2\gradient{6.7} \\
      GPT-4o-mini & 9.90 & 16.0 & 15.8\gradient{5.9} & 21.1\gradient{5.1} & 14.2\gradient{4.3} & 22.7\gradient{6.7} & 18.2\gradient{8.3} & 24.6\gradient{8.6} \\
      \midrule
      \grayline \multicolumn{9}{c}{Open-weight LMMs with more than 70B parameters } \\
      Pixtral 124B  & \textbf{12.6} & \textbf{20.3} & \textbf{16.6}\gradient{4} & \textbf{28.1}\gradient{7.8} & \textbf{21.3}\gradient{8.7} & \textbf{29.9}\gradient{9.6} & \underline{21.3}\gradient{8.7} & \textbf{31.6}\gradient{11.3} \\
      InternVL 2.5 78B & \underline{12.3} & \underline{19.7} & 13.4\gradient{1.1} & \underline{27.3}\gradient{7.6} & \underline{17.8}\gradient{5.5} & \underline{25.7}\gradient{6} & \textbf{21.7}\gradient{9.4} & \underline{31.4}\gradient{11.7} \\
      Qwen2 VL 72B & 9.10 & 15.7 & \underline{14.2}\gradient{5.1} & 19.4\gradient{3.7} & 10.7\gradient{1.6} & 19.1\gradient{3.4} & 16.6\gradient{7.5} & 25.1\gradient{9.4} \\
      LLaVA-OV 72B & 6.70 & 7.70 & 6.30\gradientb{0.4} & 11.4\gradient{3.7} & 10.7\gradient{4} & 13.1\gradient{5.4} & 13.8\gradient{7.1} & 19.7\gradient{12} \\
      Molmo-D 72B & 3.20 & 4.80 & 3.20\gradient{0.0} & 8.80\gradient{4.0} & 1.60\gradientb{1.6} & 7.00\gradient{2.2} & 5.10\gradient{1.9} & 14.2\gradient{9.4} \\
      Llama-3.2-V 90B & 4.30 & 6.10 & 4.00\gradientb{0.3} & 8.20\gradient{2.1} & 5.90\gradient{1.6} & 10.9\gradient{4.8} & 4.70\gradient{0.4} & 11.0\gradient{4.9} \\
      \midrule
      \grayline \multicolumn{9}{c}{Open-weight LMMs with fewer than 70B parameters} \\
      Pixtral 12B & \textbf{4.0} & \underline{5.9} & \textbf{6.3}\gradient{2.3} & \textbf{12.2}\gradient{6.3} & \textbf{5.5}\gradient{1.5} & \textbf{12.6}\gradient{6.7} & \textbf{13.8}\gradient{9.8} & \textbf{21.3}\gradient{15.4} \\
      InternVL 2.5 26B & \underline{3.6} & \textbf{6.6} & \underline{4.3}\gradient{0.7} & \underline{6.8}\gradient{0.2} & 2.8\gradientb{0.8} & \underline{6.7}\gradient{0.1} & \underline{8.3}\gradient{4.7} & \underline{16.7}\gradient{10.1} \\
      Qwen2 VL 7B & 0.8 & 3.3 & 1.6\gradient{0.8} & 3.9\gradient{0.6} & 2.4\gradient{1.6} & 6.3\gradient{3.0} & 6.3\gradient{5.5} & 14.7\gradient{11.4} \\
      InternVL 2.5 8B & 0.8 & 2.0 & 0.8\gradient{0.0} & 3.7\gradient{1.7} & 1.2\gradient{0.4} & 3.3\gradient{1.3} & 5.1\gradient{4.3} & 13.6\gradient{11.6} \\
      InternVL 2.5 4B & 1.2 & 4.1 & 3.2\gradient{2.0} & 3.4\gradientb{0.7} & \underline{3.2}\gradient{2.0} & 5.8\gradient{1.7} & 5.9\gradient{4.7} & 13.5\gradient{9.4} \\
      LLaVA-OV 7B & 2.0 & 1.9 & 1.6\gradientb{0.4} & 2.4\gradient{0.5} & 2.0\gradient{0.0} & 3.2\gradient{1.3} & 5.1\gradient{3.1} & 10.2\gradient{8.3} \\
      Phi-3.5-V 4B & 0.0 & 0.0 & 0.0\gradient{0.0} & 0.0\gradient{0.0} & 0.0\gradient{0.0} & 0.0\gradient{0.0} & 5.9\gradient{5.9} & 9.40\gradient{9.4} \\
      Llama-3.2-V 11B & 2.0 & 2.0 & 1.6\gradientb{0.4} & 3.9\gradient{1.9} & 2.0\gradient{0.0} & 5.2\gradient{3.2} & 4.0\gradient{2.0} & 8.80\gradient{6.8} \\
      Molmo-D 7B & 1.2 & 1.0 & 0.4\gradientb{0.8} & 1.8\gradient{0.8} & 0.8\gradientb{0.4} & 1.0\gradient{0.0} & 2.8\gradient{1.6} & 8.40\gradient{7.4} \\
      Chameleon 7B & 0.0 & 0.0 & 0.0\gradient{0.0} & 0.2\gradient{0.2} & 0.0\gradient{0.0} & 0.0\gradient{0.0} & 1.2\gradient{1.2} & 2.20\gradient{2.2} \\
      Chameleon 30B & 0.0 & 0.0 & 0.0\gradient{0.0} & 0.2\gradient{0.2} & 0.4\gradient{0.4} & 0.0\gradient{0.0} & 0.0\gradient{0.0} & 1.90\gradient{1.9} \\
      \bottomrule
      \end{tabular}
  }
  \caption{Performance of LMMs across different settings. Models are ranked based on the \faMedal ~column. The \textbf{best} and \underline{second-best} performances in each column are highlighted. The numerical values are color-coded to indicate performance changes relative to the corresponding pass@$k$ values in the \textbf{V2C} column: \textcolor{darkgreen}{green} represents improvement, \textcolor{darkred}{red} indicates decline, and \textcolor{lightorange}{yellow} denotes minimal change.}
  \label{tab:main_results}
  \end{minipage}
    \hspace{0.2cm}
    \begin{minipage}[b]{0.205\linewidth}
    \centering
    \captionsetup{type=figure}
    \includegraphics[width=\linewidth]{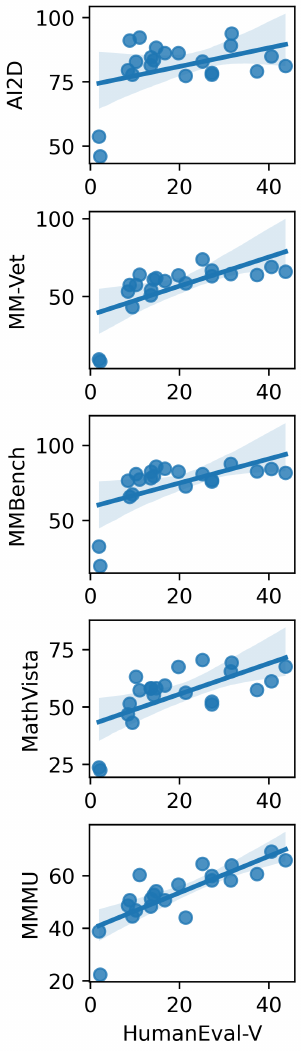}
    \caption{Comparison of LMM performance on \ourbench and other popular multimodal benchmarks.}
    \label{fig:correlation_with_other_bench}
  \end{minipage}
\end{table*}

\vspace{-0.4em}
\paragraph{Models:} We evaluate 22 state-of-the-art LMMs, including a representative mix of leading proprietary and open-weight models. Our evaluation covers five of the latest proprietary models: Claude 3.5 Sonnet, GPT-4o, GPT-4o-mini, Gemini 1.5 Pro, and Gemini 1.5 Flash. We also assess 17 top-performing open-weight models spanning various parameter sizes, including InternVL 2.5 (4/8/26/78B), Qwen2-VL (7/72B), Pixtral (12/124B), LLaVA-OV (7/72B), Llama-3.2-V (11/90B), Molmo-D (7/72B), Chameleon (7/30B), and Phi-3.5-V (4B). Further details are in Table~\ref{tab:model_details}.

\vspace{-0.6em}
\paragraph{Prompting: } we employ multiple strategies for our evaluation pipelines as illustrated in Figure~\ref{fig:evaluation_pipeline}, where LMMs may encounter four different prompting scenarios: 
(1) \textit{Direct Code Generation}. The model directly generates code based on the given diagram $D$ and function signature $\sigma$, denoted as $P_{V2C}(D, \sigma)$;
(2) \textit{Chain of Thought (CoT)}. This variant enhances the V2C pipeline by incorporating a zero-shot CoT instruction $I_{CoT}$~\citep{wei2022chain}, prompting the model to outline its reasoning process before generating the code. This is denoted as $P_{V2C}(D, \sigma, I_{CoT})$;
(3) \textit{Intermediate Textual Representation}. The model first produces a structured textual \textbf{\textit{problem specification $PS$}} based on $D$ and $\sigma$, denoted as $P_{V2T}(D, \sigma)$. The problem specification consists of three key sections: \textit{Problem Restatement}, \textit{Visual Facts}, and \textit{Visual Patterns}. This structured representation is derived from our benchmark annotation process, which we found to be effective in capturing a comprehensive description of the problem context;
(4) \textit{Code Generation from Text}. The model generates code based on $PS$ rather than the original diagram $D$, denoted as $P_{T2C}(PS, \sigma)$. 
The corresponding prompt templates are shown in Figure~\ref{fig:prompt_template}, with further details on prompt design available in Appendix~\ref{appendix:prompt_design}.

\paragraph{Hyper-parameters \& Post-processing:}
We apply two distinct decoding strategies for both code and description generation. First, we use greedy decoding to produce a single deterministic output, assessing model performance in a constrained setting. Additionally, we employ a sampling method with $Top\_p = 0.95$, $Top\_k = 20$, and a temperature of 0.8 to generate diverse outputs, allowing us to evaluate the models' ability to produce correct solutions when given multiple attempts.  
We set the maximum output length to 2048 tokens for both code and description generation. To facilitate extraction, we prompt the models to encapsulate their generated code within Markdown-style code blocks. We then apply an abstract syntax tree parser to detect and retrieve generated import statements, class definitions, and function definitions. These components are concatenated to form the final code solution. An additional ablation study on the temperature setting is presented in Appendix~\ref{appendix:ablation_on_temperature}.

\paragraph{Evaluation Metrics} 
Following established practices in code generation evaluation~\citep{chen2021evaluating, chen2022codet}, we use the pass@$k$ metric to assess functional correctness. A task is considered solved if at least one of the $k$ selected solutions passes all test cases, and pass@$k$ is the percentage of solved tasks. We report results for $k=1,3$. In the \textit{V2C} setting, we generate $n$ code samples per task and randomly select $k$ for evaluation. For greedy decoding, $n=1$ for pass@1, while for sampling-based evaluation, $n=6$ for pass@3. In the \textit{V2T2C} setting, we first sample six problem specifications per task, then use greedy decoding to generate one code solution per \textit{PS}, resulting in six solutions per task for pass@3 computation.

\section{Benchmarking Results}
\label{sec:benchmarking_results}
\begin{figure*}[t]
\begin{minipage}[b]{0.34\linewidth}
    \centering
    \includegraphics[width=\linewidth]{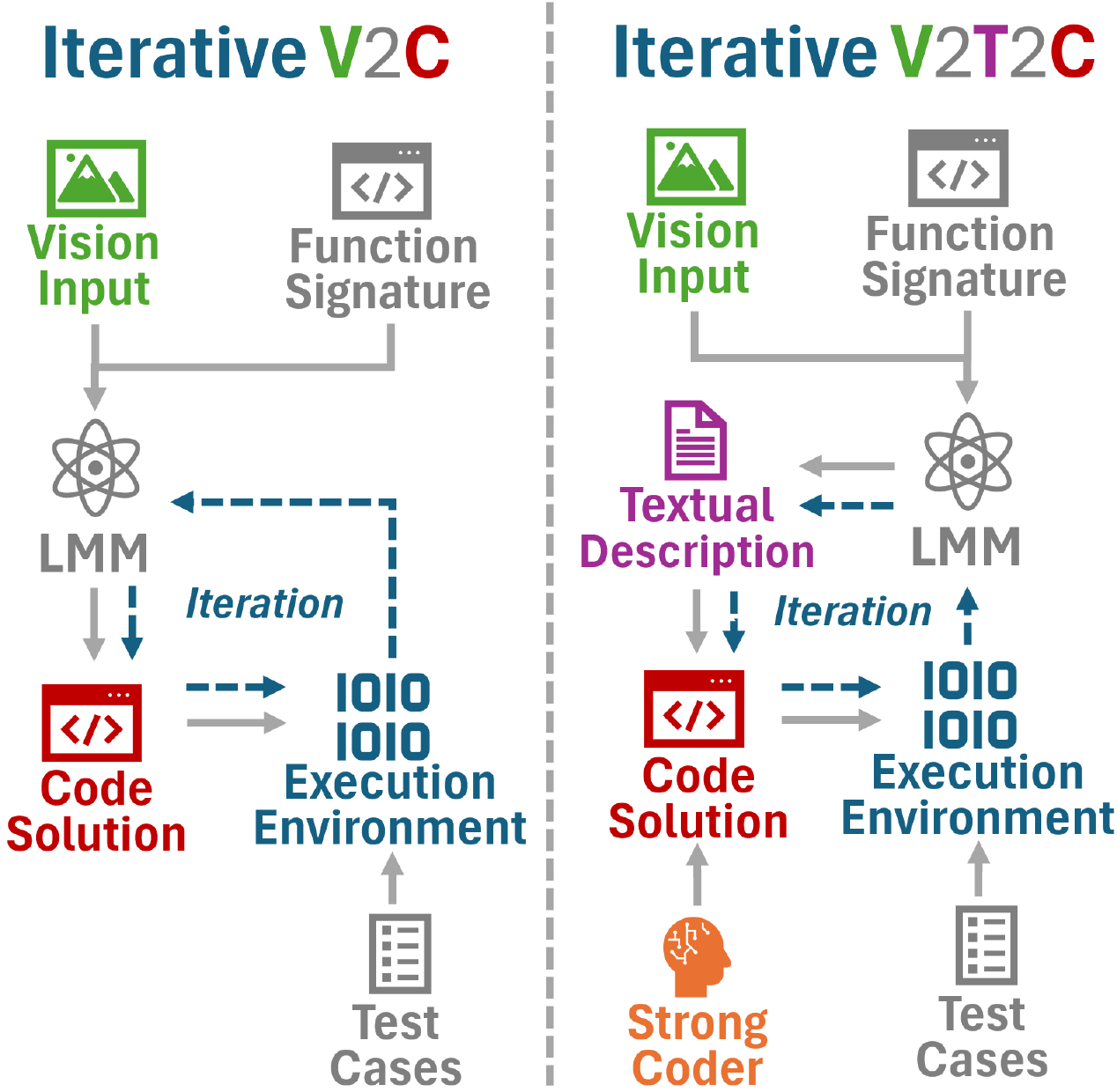}
    \vspace{-0.8em}
    \caption{Iterative evaluation pipelines.}
    \label{fig:iterative_pipeline}
\end{minipage}
\hspace{0.2cm}
\begin{minipage}[b]{0.65\linewidth}
    \centering
    \includegraphics[width=\linewidth]{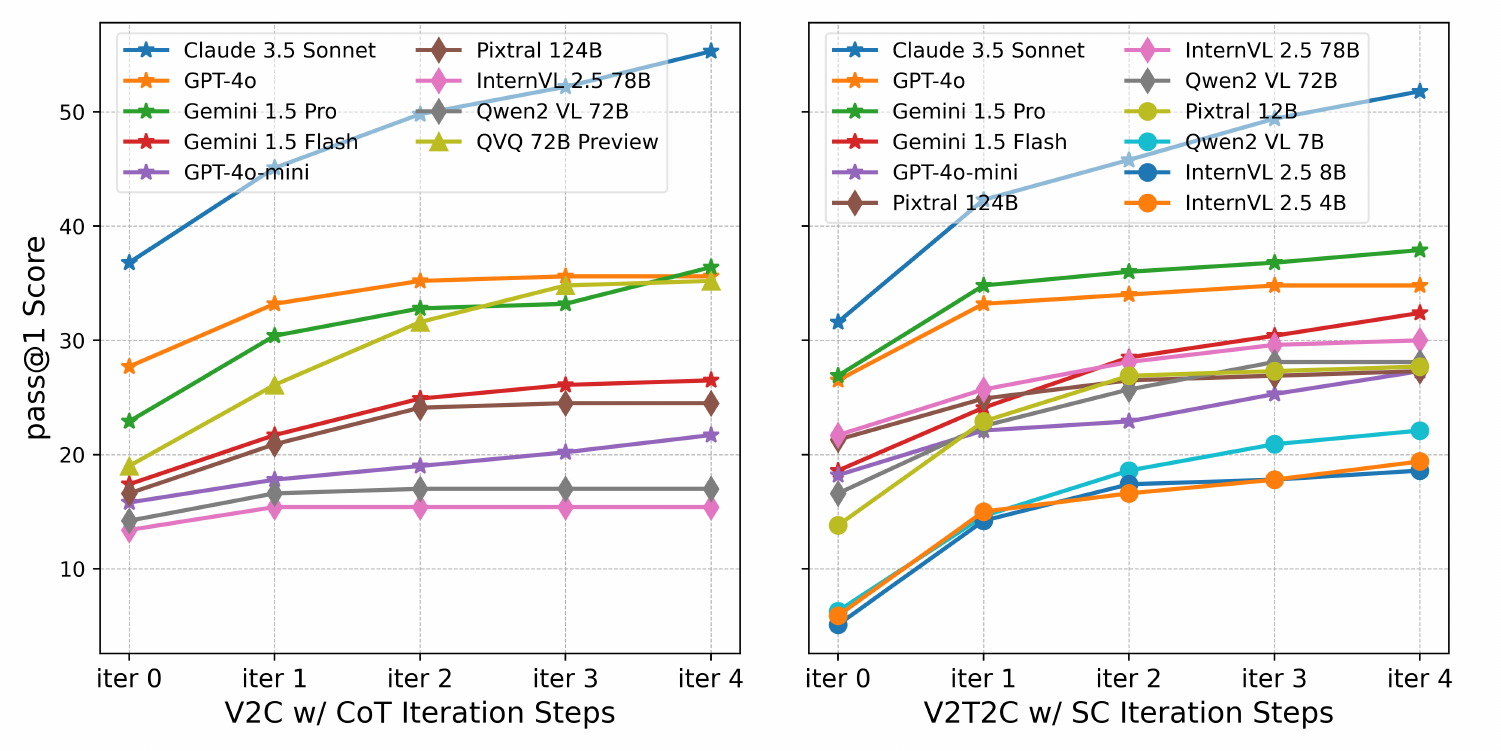}
    \vspace{-0.8em}
    \caption{Performance of LMMs under the iterative evaluation settings.}
    \label{fig:iteration_result}
\end{minipage}
\end{figure*}

\paragraph{Main Results:} We present the benchmarking results of 22 LMMs in Table~\ref{tab:main_results}, covering the four evaluation pipelines introduced in Figure~\ref{fig:evaluation_pipeline}. Additionally, Figure~\ref{fig:correlation_with_other_bench} provides a correlation analysis to illustrate the performance gap between the evaluated LMMs on \ourbench and other popular benchmarks (more details on the correlation analysis are in Appendix~\ref{appendix:correlation_with_other_bench}). Based on these results, we highlight the following key findings:
\textit{(1) Our benchmark presents unique challenges not captured by other benchmarks.}
As shown in Figure~\ref{fig:correlation_with_other_bench}, most evaluated LMMs exhibit significantly larger performance gaps on \ourbench compared to other benchmarks. While MMMU demonstrates the highest correlation with our benchmark, its results still lack sufficient discrimination between models.
\textit{(2) LMMs generally achieve their best performance under the V2T2C w/ GPT-4o setting.}
This is particularly evident for LMMs with fewer than 70B parameters, which struggle to complete tasks in the \textit{V2C} setting. These findings validate the importance of decoupling visual understanding from coding abilities.
Additionally, CoT prompting and the decoupled V2T2C pipeline show similar performance distributions, with more capable LMMs benefiting more from these enhancements than smaller models.
\textit{(3) Open-weight LMMs still lag behind top proprietary models.}
Although high-capacity open-weight LMMs (e.g., Pixtral 124B) outperform the mini/flash versions of proprietary models, they still fall short of the most capable proprietary LMMs. For smaller-scale models, Pixtral 12B, Qwen2 VL 7B, and InternVL 2.5 4B demonstrate a high performance-to-size ratio.
\textit{(4) Certain LMMs exhibit anomalously poor performance.}
Models such as Molmo-D, Llama-3.2-V, and the Chameleon series perform significantly worse than other LMMs of similar scale. Another case is Phi-3.5-V, which appears to lack coding ability, achieving a performance score of 0 in the \textit{V2C} settings, compared to 9.4\% pass@3 when assisted by GPT-4o for code generation.
\textit{(5) Additional Results of o1 and QVQ:} We also evaluated reasoning-enhanced LMMs that leverage test-time scaling by generating long chain-of-thought (CoT) reasoning. Specifically, we assessed OpenAI o1~\citep{openaio1} and QVQ-72B-Preview~\citep{qvq-72b-preview} under the \textit{V2C w/ CoT} setting, achieving pass@1 scores of 40.6\% and 19.0\%, respectively. Our case study reveals that both models still struggles with visual understanding, often failing to identify rules or patterns in the diagrams. Meanwhile, QVQ primarily fails due to excessively long CoT reasoning, with 35\% of cases unable to generate a valid code solution within the 20k token limit. These results underscore the complexity of the diagrams in our benchmark. Example cases for o1 and QVQ are shown in Figures~\ref{fig:error_case_q63_o1},~\ref{fig:error_case_q73_o1},~\ref{fig:error_case_q18_o1} and Figures~\ref{fig:error_case_q63_qvq},~\ref{fig:error_case_q73_qvq},~\ref{fig:error_case_q18_qvq}.

\begin{figure*}[t]
\begin{minipage}[b]{0.31\linewidth}
    \centering
    \includegraphics[width=\linewidth]{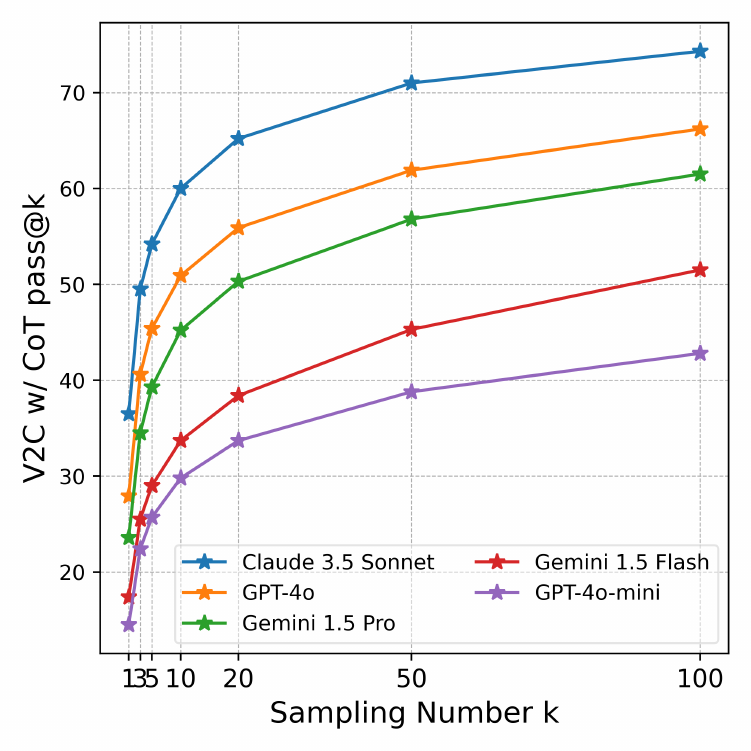}
    \vspace{-0.8em}
    \caption{Performance with increased sample size.}
    \label{fig:sampling_num}
\end{minipage}
\hspace{0.2cm}
\begin{minipage}[b]{0.31\linewidth}
    \centering
    \includegraphics[width=\linewidth]{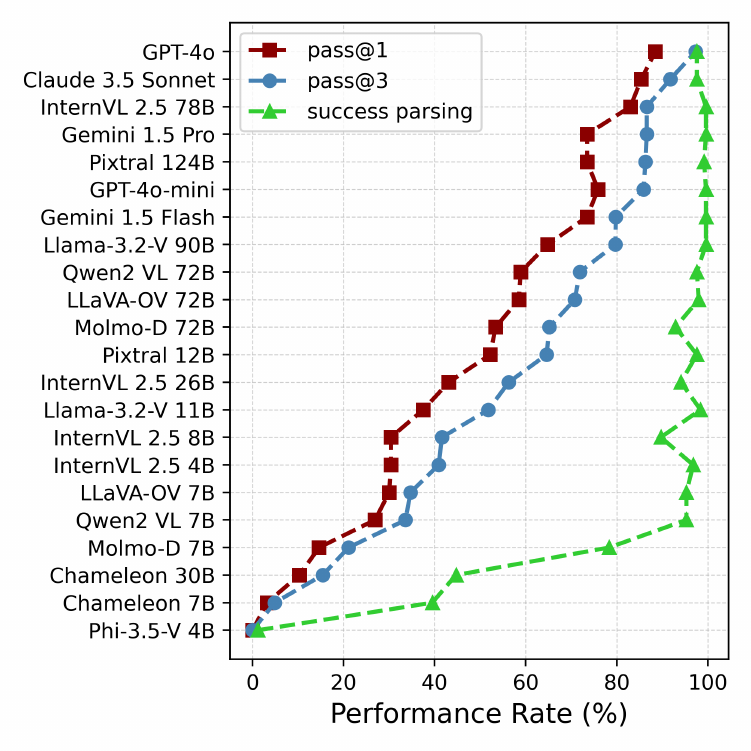}
    \vspace{-0.8em}
    \caption{LMMs' performance with human problem specifications.}
    \label{fig:gt_description_to_code}
\end{minipage}
\hspace{0.2cm}
\begin{minipage}[b]{0.31\linewidth}
    \centering
    \includegraphics[width=\linewidth]{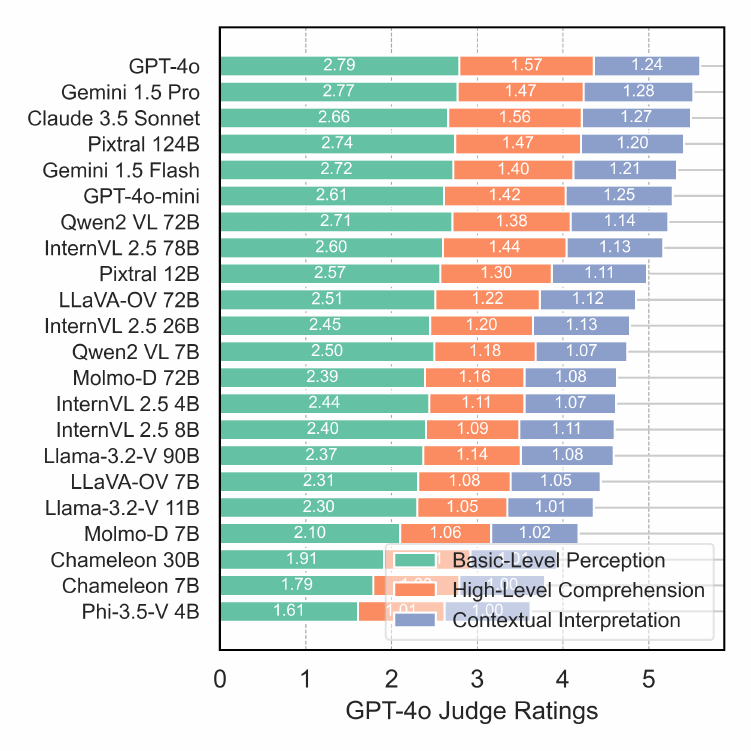}
    \vspace{-0.8em}
    \caption{LLM-as-Judge ratings for LMMs in the \textit{V2T2C} setting.}
    \label{fig:lmjudge_score}
\end{minipage}
\end{figure*}

\vspace{-0.6em}
\paragraph{Iterative Benchmarking:} We introduce an iterative benchmarking pipeline to evaluate LMMs' ability to reason over environmental feedback and perform self-refinement—an essential skill for real-world problem-solving. 
Figure~\ref{fig:iterative_pipeline} illustrates two types of iterative pipelines derived from the \textit{V2C} and \textit{V2T2C w/ SC} settings. In these pipelines, LMMs must refine either their generated code solutions or textual descriptions based on feedback from the execution environment. To support this process, we design new prompt templates (Figure~\ref{fig:iterative_prompt_template}) that guide the refinement steps. Specifically, for each task, LMMs perform an additional iteration if the generated code contains syntax errors or fails to pass all test cases. The feedback includes detailed error messages or the failed test cases' inputs and expected outputs.

For the iterative evaluation, we select the most capable LMMs across different parameter scales, using greedy decoding and the pass@1 metric. Figure~\ref{fig:iteration_result} presents the results, where \textit{iter 0} represents the first round of generation without feedback. We observe that LMMs generally improve across iterations, with more capable models achieving larger performance gains. Notably, some models, such as Claude and QVQ, exhibit stronger self-refinement capabilities. 
We also investigate the cases that are corrected after iterations and find that approximately 90\% of these cases are corrected due to the models’ improved understanding of the diagram and task. The remaining 10\% are cases that fix edge conditions highlighted by the test case feedback. None of these corrections result from hard-coding the exposed test cases into the code solutions.

\section{Experimental Analysis}
\label{sec:analysis_exp}
This section presents our analysis of model performance under various settings, including increased sampling sizes, human-annotated problem specifications, the use of GPT-4o judge for rating LMMs' diagram descriptions, and error pattern analysis. Our goal is to examine both the potential and limitations of the LMMs on \ourbench. 
Further analysis is provided in Appendix~\ref{appendix:deeper_analysis}, where we explore the co-occurrence of capability aspects required in our benchmark tasks, the stability of using QwenCoder-32B as a strong coder instead of GPT-4o, and experimental evidence supporting the value of tasks diversified from the seed tasks.

\paragraph{Performance with Increased Sample Size:} We scale up the number of samples for five proprietary LMMs to explore their potential performance. We increase the sampling number $n$ to 200, using the same $Top\_p$, $Top\_k$, and max token limitations outlined in Section~\ref{sec:benchmark_setup} to calculate the pass@100 score under the \textit{V2C w/ CoT} setting. As shown in Figure~\ref{fig:sampling_num}, we observe a consistent performance improvement across all models with larger sample sizes. Notably, Claude 3.5 Sonnet achieves a significant improvement, reaching 74.3\% pass@100, underscoring the strong potential for these models when scaling up sample sizes.

\vspace{-0.6em}
\paragraph{Coding Performance with Human-Annotated Problem Specifications:} We evaluate all LMMs on a new task where they \textit{generate code based on human-annotated problem specifications}, without direct access to the diagrams. This setup isolates their ability to perform visual reasoning and generate code. We also calculate the \textit{success parsing rate} using Pylint~\citep{wiki:Pylint}, which measures the syntactic correctness of the generated code, independent of its functional accuracy.  The results, presented in Figure~\ref{fig:gt_description_to_code}, show that \textit{most LMMs demonstrate strong coding capabilities}, generally outperforming their best results from Table~\ref{tab:main_results}. Notably, GPT-4o achieves 96.5\% pass@3, a significant improvement over its 40.5\% pass@3 in the \textit{V2T2C} setting. Smaller models, such as InternVL 2.5 4B, also show substantial improvement. 

We also evaluate a setting where LMMs generate code based solely on the function signature, without access to diagrams or descriptions, and find that \textit{none of the five proprietary models are able to pass any tasks}. This underscores the necessity of visual context in our benchmark. These results suggest that current LMMs face more challenges in visual reasoning than coding on \ourbench. 

\vspace{-0.6em}
\paragraph{LLM-as-Judge Ratings:} We evaluate the problem specifications (PS) generated by LMMs in the \textit{V2T2C} setting using GPT-4o as the judge. GPT-4o rates the PS in three dimensions: Basic-Level Perception (identifying basic visual elements), High-Level Comprehension (understanding objects, patterns, transformations, and operations), and Contextual Interpretation (clear description without vagueness or hallucinations) as outlined in the prompt template shown in Figure~\ref{fig:lmjudge_prompt_template}. Ratings are on a 1-3 scale, where 1 indicates severe errors and 3 reflects near perfection in the capability dimension.
The results, shown in Figure~\ref{fig:lmjudge_score}, reveal that while LMMs generally excel in basic perceptual abilities, they struggle with high-level comprehension and clarity of expression. Notably, the performance gap between models is small. We also find the difference in ratings between passed and failed tasks is minimal. For example, GPT-4o scores 2.9, 2.0, and 1.3 across the three dimensions on passed tasks, compared to average ratings of 2.8, 1.6, and 1.2 across all tasks, highlighting the limitations of using LLM-as-judge as an evaluation tool. This lack of robustness may stem from rigid comparisons to human-annotated PS, further emphasizing the importance of pass rates as the evaluation metric.

\vspace{-0.6em}
\paragraph{Error Analysis:} 
We conducted a comprehensive error analysis to understand the limitations of current LMMs in \ourbench, as detailed in Appendix~\ref{appendix:error_analysis}. Our analysis examined correlations between model performance and three key factors: task types, general capability dimensions, and specific capability aspects. The results reveal that LMMs particularly struggle with tasks involving \textit{Transformation} and \textit{Iterative Calculation}. And models show notable difficulties with specific capabilities such as understanding dynamic patterns (e.g., spirals, circular arrangements) and spatial transformations (e.g., stacking, translation).
Interestingly, our investigation of task difficulty metrics shows that LMM performance correlates poorly with human-perceived difficulty measures, including both programming complexity (measured by cyclomatic complexity) and visual comprehension difficulty (measured by description length). This suggests a fundamental gap in LMMs' visual reasoning capabilities, where even tasks considered trivial by humans can prove challenging for state-of-the-art models. For concrete examples of these challenges, we present representative error cases in Figures~\ref{fig:error_case_q63_sonnet} to~\ref{fig:error_case_q47_sonnet}.

\section{Related Work}

\paragraph{Benchmarks Involving Diagrams:} Prior work on multimodal benchmarks can be categorized into several groups: (1) General-purpose multimodal evaluation benchmarks \citep{yue2024mmmu,liu2023mmbench,yu2023mm,li2023seed,ying2024mmt,chen2024we} that assess models' broad multidisciplinary capabilities; (2) Scientific diagram understanding \citep{lu2022learn,kembhavi2016diagram}; (3) Mathematical visual reasoning \citep{lu2023mathvista,wang2024measuring,zhang2024mathverse}; (4) Data visualization comprehension \citep{masry2022chartqa,wang2024charxiv,chollet2019measure} that focus on plots and charts; (5) Abstract reasoning \citep{zhang2019raven,jiang2024marvel,nie2020bongard,chia2024puzzlevqa}; and (6) Specialized diagram understanding including abstract symbol interpretation and geometric spatial reasoning \citep{lu2021iconqa,rahmanzadehgervi2024vision}. While these benchmarks cover various aspects of visual understanding, they do not address the complex diagrams in the coding context.

\paragraph{Multimodal Code Generation:}
Recent work in multimodal code generation has focused on two main Categories. In the first category, researchers have explored derendering web pages into functional code ~\citep{si2024design2code, laurenccon2024unlocking} and converting scientific figures into their corresponding plotting code ~\citep{shi2024chartmimic, wu2024plot2code}. The second category includes Program-based VQA approaches, where models leverage pre-defined modules to answer visual questions ~\citep{suris2023vipergpt, subramanian2023modular}. 
MMCode~\citep{li2024mmcode} is the most related coding benchmark to ours, evaluating LMMs' coding abilities using problems with visual demonstrations from competition platforms. However, our benchmark differs in its dedicated focus on assessing the visual capabilities of LMMs.  We provide a detailed discussion in Appendix~\ref{appendix:mmcode}, highlighting the differences between \ourbench and MMCode in terms of visual indispensability, task complexity, and evaluation design.

\section{Conclusion}
In this paper, we introduced \ourbench, a novel benchmark designed to evaluate LMMs' capabilities in understanding and reasoning over  diagrams in programming contexts. Through comprehensive experiments, we demonstrated that current LMMs, while showing promising performance, still face significant challenges in complex diagram understanding and reasoning. Our extensive experimental results and analysis provide valuable insights for the future development of more sophisticated visual reasoning abilities in AI systems.

\newpage

\section{Limitations}
Despite the valuable contributions of our benchmark, several limitations remain that we aim to address in future work:

\paragraph{Limited Benchmark Size:} The size of our benchmark is constrained by the significant cost of human annotation, as we prioritize high-quality task design to ensure meaningful insights, with each annotator dedicating over 200 hours to constructing \ourbench. Nevertheless, our benchmark includes 253 tasks, comparable to many well-established human-annotated benchmarks in academia and industry, such as HumanEval~\citep{chen2021evaluating} with 164 tasks, MM-Vet~\citep{yu2023mm} with 218, and Vibe-Eval~\citep{padlewski2024vibe} with 269. Notably, none of the current popular multimodal benchmarks feature manually drawn diagrams, further distinguishing \ourbench. Furthermore, \ourbench offers a diverse and balanced set of task types covering a wide range of capability aspects, enabling us to uncover unique insights into the limitations of current LMMs.

\paragraph{Limited Model Coverage:} While our experiments evaluate a representative set of top-performing LMMs, the rapid pace of model development means newly released models may not be covered in our current evaluation. To address this, we plan to publicly release our evaluation toolkit and dataset, along with an up-to-date leaderboard to track ongoing advancements. This will enable benchmarking of new models as they become available, ensuring \ourbench remains relevant and continuously updated.

\paragraph{Limitations in Exploring Advanced Methods:} While our experiments cover various evaluation settings, including chain-of-thought (CoT), iterative refinement, and long-CoT-enhanced LMMs, our exploration of more advanced CoT techniques is limited. Methods such as supervised fine-tuning~\citep{chen2024m}, reinforcement learning~\citep{snell2024scaling}, or more complex CoT approaches~\citep{yao2024tree, mitra2024compositional} could further enhance LMM reasoning capabilities. However, these techniques are challenging to apply to diagram reasoning due to the lack of high-quality training data in this domain. As our primary objective is to bridge the gap in diagram reasoning benchmarks, we leave the exploration of more sophisticated reasoning-enhancing methods to future work.

\newpage

\bibliography{reference}

\newpage

\appendix

\section*{Appendix}
\startcontents[sections]
\printcontents[sections]{l}{1}{\setcounter{tocdepth}{2}}
\newpage

\section{Comparison with Other Benchmarks}
\subsection{Correlation Analysis}
\label{appendix:correlation_with_other_bench}
\begin{table*}[h]
  \centering
  \scalebox{0.85}{
      \begin{tabular}{lcccccccc}
      \toprule
      \textbf{Models} & \scriptsize \textbf{AI2D} & \scriptsize \textbf{MM-Vet} & \scriptsize \textbf{MMBench} & \scriptsize \textbf{MathVista} & \scriptsize \textbf{MMMU} & \scriptsize \textbf{MMStar} & \scriptsize \textbf{HallusionBench} & \scriptsize \textbf{HumanEval-V}~\faMedal \\
      \midrule
      \grayline \multicolumn{9}{c}{Proprietary LMMs} \\
        Claude 3.5 Sonnet & 81.2 & 66.0 & 81.7 & 67.7 & \textbf{65.9} & 65.1 & 55.1 & \textbf{43.7} \\
        GPT-4o & 84.9 & \textbf{69.1} & 84.3 & 61.3 & \textbf{69.2} & 65.1 & 56.2 & \textbf{40.5} \\
        Gemini 1.5 Pro & 79.1 & 64.0 & 82.8 & 57.5 & 60.6 & 59.1 & 45.6 & 37.3 \\
        Gemini 1.5 Flash & 78.5 & 63.2 & 76.9 & 51.2 & 58.2 & 55.8 & 48.5 & 27.2 \\
        GPT-4o-mini & 77.8 & 66.9 & 76.0 & 52.4 & 60.0 & 54.8 & 46.1 & 24.6 \\
      \midrule
      \grayline \multicolumn{9}{c}{Open-Weight LMMs} \\
        Pixtral 124B & \textbf{93.8} & - & - & \textbf{69.4} & 64.0 & - & - & 31.6 \\
        InternVL 2.5 78B & 89.2 & 64.4 & \textbf{87.7} & 65.6 & 58.3 & \textbf{72.1} & \textbf{57.4} & 31.4 \\
        Qwen2 VL 72B & 83.0 & \textbf{74.0} & 81.0 & \textbf{70.5} & 64.5 & 25.9 & \textbf{58.7} & 25.1 \\
        LLaVA-OV 72B & 86.2 & 63.7 & 82.6 & 67.5 & 56.6 & \textbf{65.8} & 47.9 & 19.7 \\
        Molmo-D 72B & 83.4 & 61.1 & 79.5 & 55.2 & 52.8 & 63.3 & 46.4 & 14.2 \\
        Llama-3.2-V 90B & \textbf{92.3} & 64.1 & 77.3 & 57.3 & 60.3 & 55.3 & 44.1 & 11.0 \\
        Pixtral 12B & 77.4 & 58.5 & 72.7 & 56.3 & 44.1 & 54.5 & 47.0 & 21.3 \\
        InternVL 2.5 26B & 86.2 & 60.0 & 84.6 & 59.4 & 50.7 & 66.5 & 55.8 & 16.7 \\
        Qwen2 VL 7B & 88.3 & 62.0 & \textbf{85.9} & 58.2 & 54.1 & 16.3 & 50.4 & 14.7 \\
        InternVL 2.5 8B & 84.6 & 54.3 & 82.5 & 58.3 & 51.2 & 63.2 & 49.0 & 13.6 \\
        InternVL 2.5 4B & 81.4 & 50.9 & 78.2 & 58.1 & 48.3 & 58.7 & 46.6 & 13.5 \\
        LLaVA-OV 7B & 82.8 & 57.5 & 80.9 & 63.2 & 46.8 & 61.9 & 31.6 & 10.2 \\
        Phi-3.5-V 4B & 77.8 & 43.2 & 67.4 & 43.2 & 44.6 & 47.5 & 40.5 & 9.4 \\
        Llama-3.2-V 11B & 91.1 & 57.6 & 65.8 & 51.5 & 50.7 & 49.8 & 40.3 & 8.8 \\
        Molmo-D 7B & 79.6 & 53.3 & 76.5 & 46.9 & 48.7 & 54.4 & 47.4 & 8.4 \\
        Chameleon 7B & 46.0 & 8.3 & 19.8 & 22.5 & 22.4 & 31.1 & 17.1 & 2.2 \\
        Chameleon 30B & 53.7 & 9.7 & 32.7 & 23.8 & 38.8 & 32.7 & 18.6 & 1.9 \\
      \bottomrule
      \end{tabular}
  }
  \caption{A performance comparison of 22 LMMs across \ourbench and seven popular multimodal benchmarks. Models are ranked according to the \faMedal~column. Results for \ourbench correspond to the \textit{V2T2C w/ GPT-4o} setting from Table~\ref{tab:main_results}. The top two results for each column are highlighted in \textbf{bold}.}
  \label{table:compare_with_other_bench}
\end{table*}

To assess whether \ourbench identifies specific weaknesses not captured by existing benchmarks, we select seven widely used multimodal benchmarks that cover a range of multidisciplinary abilities. These include AI2D~\citep{kembhavi2016diagram}, MMVet~\citep{yu2023mm}, MMBench~\citep{liu2023mmbench}, MathVista~\citep{lu2023mathvista}, MMMU~\citep{yue2024mmmu}, MMStar~\citep{chen2024we}, and HallusionBench~\citep{guan2023hallusionbench}. Performance results for the 22 LMMs evaluated in this paper are collected from the OpenVLM Leaderboard~\citep{duan2024vlmevalkit}, as well as corresponding papers and reports. These results are shown alongside the pass@3 scores for \ourbench under the \textit{V2T2C w/ GPT-4o} setting in Table~\ref{table:compare_with_other_bench}.

From the analysis, we observe that open-weight LMMs with more than 70B parameters generally perform well on the selected benchmarks, with models like Pixtral, InternVL 2.5, and Qwen2 VL even outperforming proprietary models such as GPT-4o and Claude 3.5 Sonnet in several cases. Llama-3.2-V also shows competitive performance. However, open-weight LMMs exhibit significantly lower performance on \ourbench, suggesting that our benchmark uncovers model weaknesses that may not be apparent in other benchmarks.

\begin{figure*}[t]
\centering
\includegraphics[width=\linewidth]{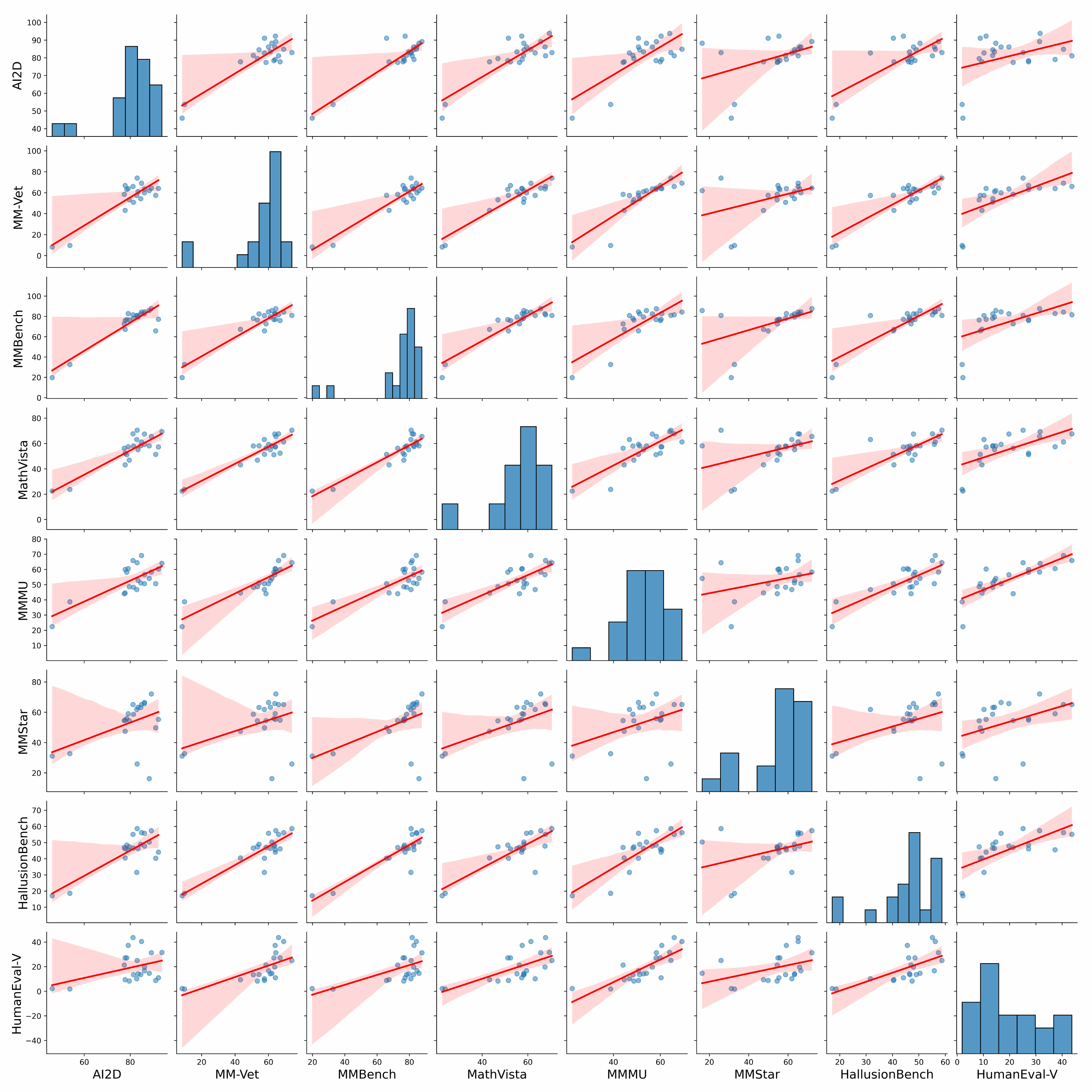}
\vspace{-0.5em}
\caption{Correlations between eight multimodal benchmarks, including \ourbench. Each subplot displays the relationship between two benchmarks, while the diagonal subplots show the performance distribution for the corresponding benchmark.}
\label{fig:full_correlation_with_other_bench}
\end{figure*}

To quantify the relationship between \ourbench and the other five benchmarks, we visualize the performance of the 22 LMMs across all benchmarks using regression plots for each benchmark pair in Figure~\ref{fig:full_correlation_with_other_bench}. The plots reveal low correlations between \ourbench and the other benchmarks, with notable differences in performance across models. Overall, the performance of all models remains lower on \ourbench compared to the other benchmarks.

\subsection{Diagrams in Other Benchmarks}
\begin{figure*}[t]
\centering
\includegraphics[width=\linewidth]{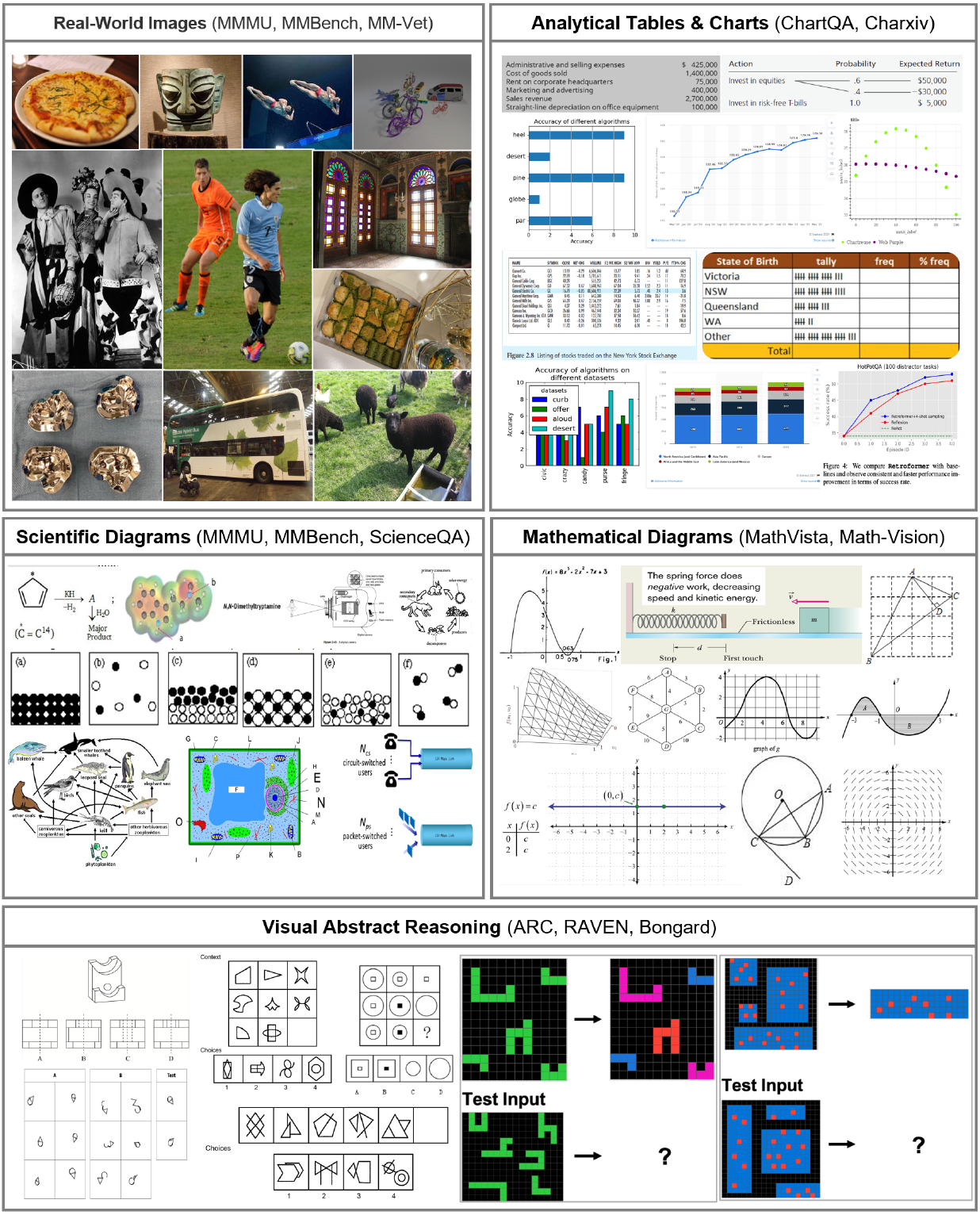}
\vspace{-0.5em}
\caption{A comparison between diagrams covered in popular multimodal benchmarks.}
\label{fig:diagram_comparison}
\end{figure*}

Figure~\ref{fig:diagram_comparison} presents a comprehensive comparison of five distinct categories of diagrams commonly used in various benchmarks and coding platforms, showcasing the diverse range of visual reasoning challenges in the open world. The first category consists of real-world images from benchmarks such as MMMU, MMBench, and MM-Vet, encompassing everyday photographs of food, sports, architecture, art, and wildlife in both color and monochrome formats. These images test general visual recognition and understanding capabilities, contrasting sharply with the more structured representations found in other categories.

The second and third categories focus on analytical and scientific visualization. Analytical tables and charts, evaluated through benchmarks like ChartQA~\citep{masry2022chartqa} and Charxiv~\citep{wang2024charxiv}, comprise business and scientific data visualizations including bar charts, line graphs, and frequency tables. Scientific diagrams featured in MMMU~\citep{yue2024mmmu}, MMBench~\citep{liu2023mmbench}, and ScienceQA~\citep{lu2022learn} present technical illustrations of molecular structures, particle dynamics, and ecosystem relationships. While both categories deal with data representation, they differ in their approach: analytical charts emphasize quantitative interpretation, whereas scientific diagrams focus on conceptual understanding.

Mathematical diagrams, assessed through benchmarks such as MathVista~\citep{lu2023mathvista} and Math-Vision~\citep{wang2024measuring}, represent another crucial category that bridges pure mathematics with practical applications. These include function graphs, geometric constructions, and physics diagrams, demonstrating complex mathematical concepts through visual means. This category shares some common ground with programming-related diagrams, particularly in their emphasis on logical relationships and systematic thinking.

The fifth category encompasses visual abstract reasoning, evaluated through benchmarks like ARC-AGI~\citep{chollet2019measure}, RAVEN~\citep{zhang2019raven}, and Bongard~\citep{nie2020bongard}. These tests feature grid-based patterns and geometric transformations that assess abstract thinking and pattern recognition skills. This category bears the closest resemblance to programming-related diagrams in terms of logical abstraction and systematic problem-solving approaches.

\subsection{Diagrams in \ourbench}
\begin{figure*}[t]
\centering
\includegraphics[width=\linewidth]{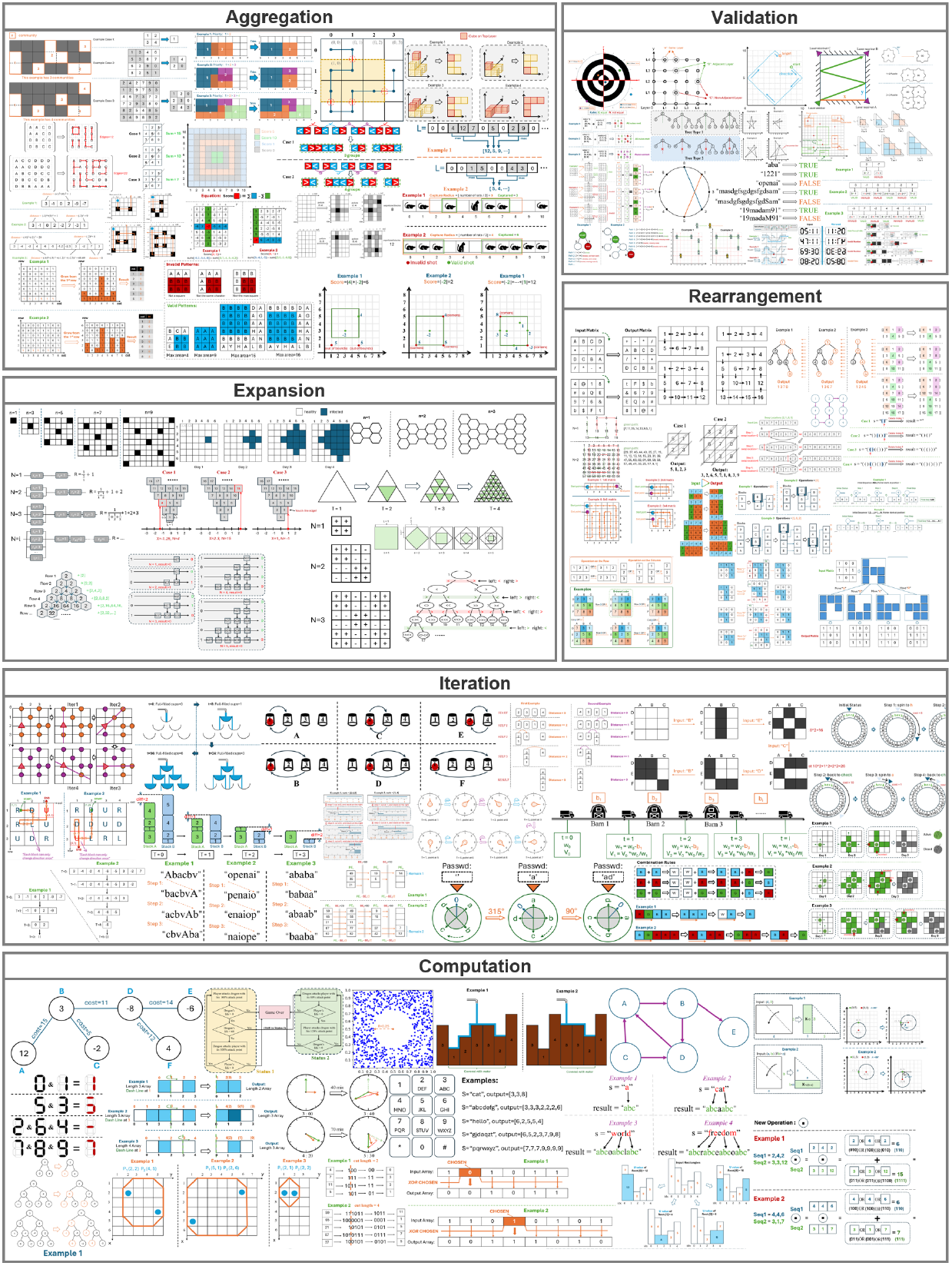}
\vspace{-0.5em}
\caption{A curated selection of diagrams representing the six task types in \ourbench.}
\label{fig:our_diagram_examples}
\end{figure*}

Figure~\ref{fig:our_diagram_examples} presents six fundamental task types in the \ourbench benchmark, each representing distinct cognitive challenges in visual reasoning. Our benchmark employs a rich variety of visual elements including geometric shapes, symbolic notations, matrices, and directed graphs. These representations are enhanced through connecting lines, arrows, color-coding, and numerical annotations to effectively capture relationships and transformations between components. The visual representations maintain clarity across all categories while scaling in complexity to accommodate different difficulty levels. Through careful design of visual elements and systematic progression of patterns, each task type provides a clear framework for evaluating specific aspects of visual reasoning and problem-solving abilities.

The six task categories demonstrate diverse problem-solving requirements: Aggregation tasks (18\% of the benchmark) introduce new grouping and aggregation rules for input data; Validation tasks (17\%) define conditional rules to verify or classify input data; Expansion tasks (16\%) focus on defining new patterns that evolve or extend data elements; Rearrangement tasks (17\%) establish new traversal patterns to reorganize data; Iteration tasks (17\%) define new iterative operations applied to input data; and Computation tasks (20\%) introduce new computational rules and operations.

What distinguishes our benchmark is not only its balanced distribution across task types but also the wide interconnection between categories. While each category emphasizes specific problem-solving skills, real-world scenarios often require combining multiple approaches. For instance, computation tasks may incorporate iterative processes, while aggregation problems might require validation steps. This interconnected design reflects the complexity of practical problem-solving scenarios where multiple cognitive skills must be applied simultaneously.

\section{More Details on Data Annotation}
\subsection{Data Collection and Screening}
\label{appendix:data_collection}
\begin{figure}[t]
\centering
\includegraphics[width=0.9\linewidth]{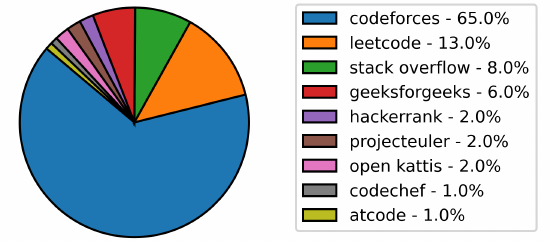}
\vspace{-0.5em}
\caption{Sources of the screened tasks for annotation.}
\label{fig:task_sources}
\end{figure}

Our data collection process involves two primary sources: coding challenge platforms, such as CodeForces, and the Q\&A platform Stack Overflow (SO). Each coding problem undergoes a rigorous screening process to ensure it aligns with the standards of \ourbench. Annotators are instructed to exclude problems that: (1) require knowledge of specific programming frameworks or libraries, (2) contain images that are not abstract diagrams, (3) provide no useful information for solving the problem, or (4) require excessive textual context for interpretation.

The majority of our tasks are sourced from coding challenge platforms, especially CodeForces, as shown in Figure~\ref{fig:task_sources}, where we display the distribution of screened tasks by platform.
For coding challenge platforms, we use the open-source MMCode dataset~\cite{li2024mmcode}, which includes coding problems from various platforms with visual elements in the problem descriptions. However, we find that most of these problems are unsuitable for \ourbench. Many images are non-essential, as they can be inferred from the textual problem descriptions. Some problems, though containing relevant visual information, are overly complex and require lengthy textual descriptions to interpret, violating our requirement for self-explanatory visual content. After careful screening, less than 5\% of the viewed problems pass our standards.

We select SO for its extensive repository of real-world programming problems. To identify relevant posts, we first filter questions from 2020 that have non-negative votes and accepted answers. Then, we focus on posts that include images in the question body and code blocks in the corresponding answers, narrowing down further to those tagged with Python. After this automated filtering, we manually review the remaining posts, excluding topics related to front-end, mobile, or UI development, as these often require external frameworks and libraries that do not align with the goals of our benchmark. We also exclude posts where the images provide information in textual nature, such as code snippets, error messages, or execution outputs. Ultimately, we identified suitable questions primarily covering topics like geometry, plotting, and image processing.

\begin{figure*}[t]
\centering
\includegraphics[width=\linewidth]{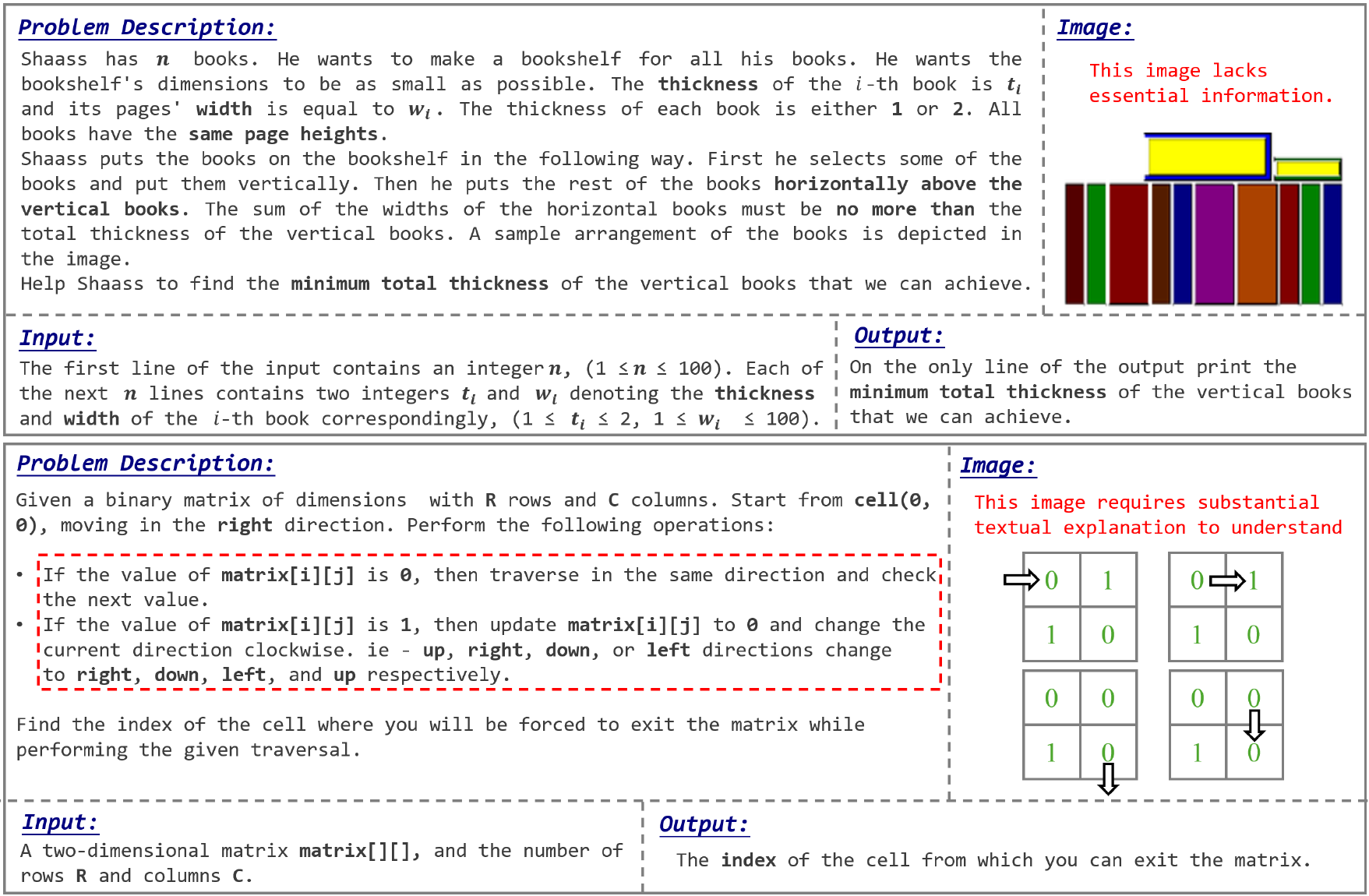}
\vspace{-0.5em}
\caption{Two negative examples in our data screening process: the first example is sourced from CodeForces (\url{https://codeforces.com/problemset/problem/294/B}), and the second from GeeksforGeeks (\url{https://www.geeksforgeeks.org/problems/last-cell-in-a-matrix/1}).}
\label{fig:negtive_screening_example}
\end{figure*}

To further illustrate our screening process, we present two negative examples that do not meet our standards in Figure~\ref{fig:negtive_screening_example}: (1) The first example is a coding problem from CodeForces, where the task is to determine an optimal stacking method for a set of books with identical heights, given their thickness and width, in order to minimize the total thickness. While the provided image shows a possible stacking configuration, it lacks critical information, such as constraints on the stacking method and precise book dimensions. Moreover, the core problem-solving details are conveyed primarily through text, making the image non-essential for understanding the solution. (2) The second example is a coding problem from GeeksForGeeks, which involves traversing a 2D matrix according to a specified pattern, starting from the top-left corner and identifying the traversal endpoint. Although the image offers a basic representation of the matrix, the traversal pattern is too complex to be effectively captured visually and requires significant textual explanation. As a result, the textual description carries more problem-solving information than the image itself, violating our requirement for the visual context to be self-explanatory and serve as the primary source of information.

\subsection{Recreation and Diversification}
\label{appendix:task_annotation}
\begin{table*}[t]
    \centering
    \resizebox{\linewidth}{!}{
    \begin{tabular}{ll}
    \toprule
    \textbf{Modification Aspects} & \textbf{Examples} \\ 
    \midrule
    Spatial Transformation & Adjust concatenation, swapping, grouping, or stacking order; modify the direction of translation, flipping, rotation, or folding. \\ 
    Mathematical Operations & Alter arithmetic operations, sorting order, or aggregation rules. \\ 
    Dynamic Patterns & Reverse alternation sequences; switch increments to decrements; change the direction of spirals or zigzags; adjust layer layouts. \\ 
    Object Attributes & Change the color, size, shape, angle or position of objects. \\ 
    Object Relations & Reverse nesting or overlap order; modify connection, intersection, or adjacency rules; adjust boundary interaction conditions. \\
    Data Structure & Modify graph direction, data type in array, matrix dimensions. \\ 
    \bottomrule
    \end{tabular}
    }
    \caption{Examples of modifications applied to diversify seed tasks. Modifications of a task may span many aspects.}
    \label{tab:diversify}
\end{table*}

\begin{figure*}[t]
\centering
\includegraphics[width=\linewidth]{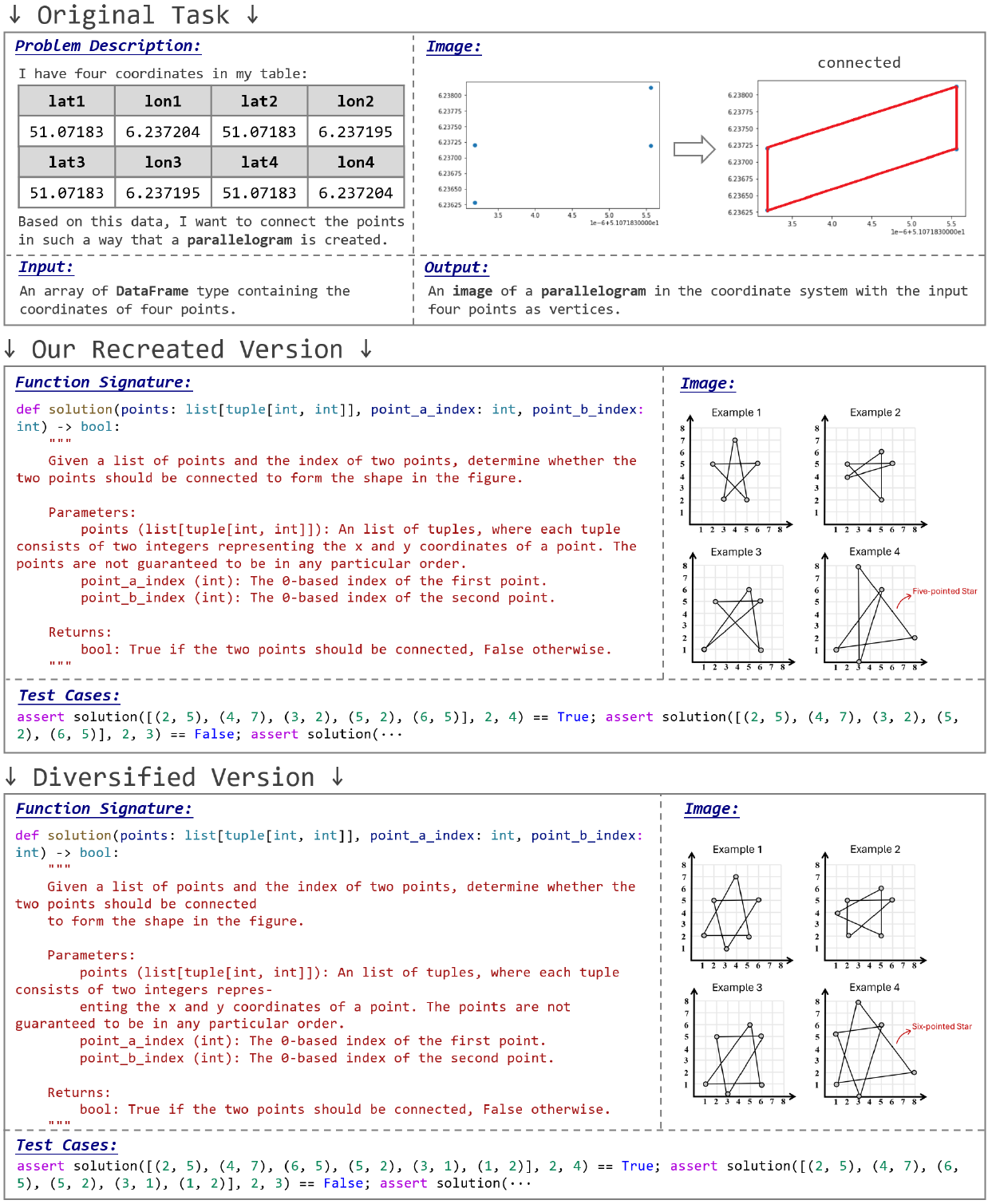}
\vspace{-0.5em}
\caption{Task annotation examples illustrating the recreation and diversification applied to the screened coding problem. The original problem is sourced from Stack Overflow (\url{https://stackoverflow.com/questions/69163515}).}
\label{fig:annotation_example_1}
\end{figure*}

\begin{figure*}[t]
\centering
\includegraphics[width=\linewidth]{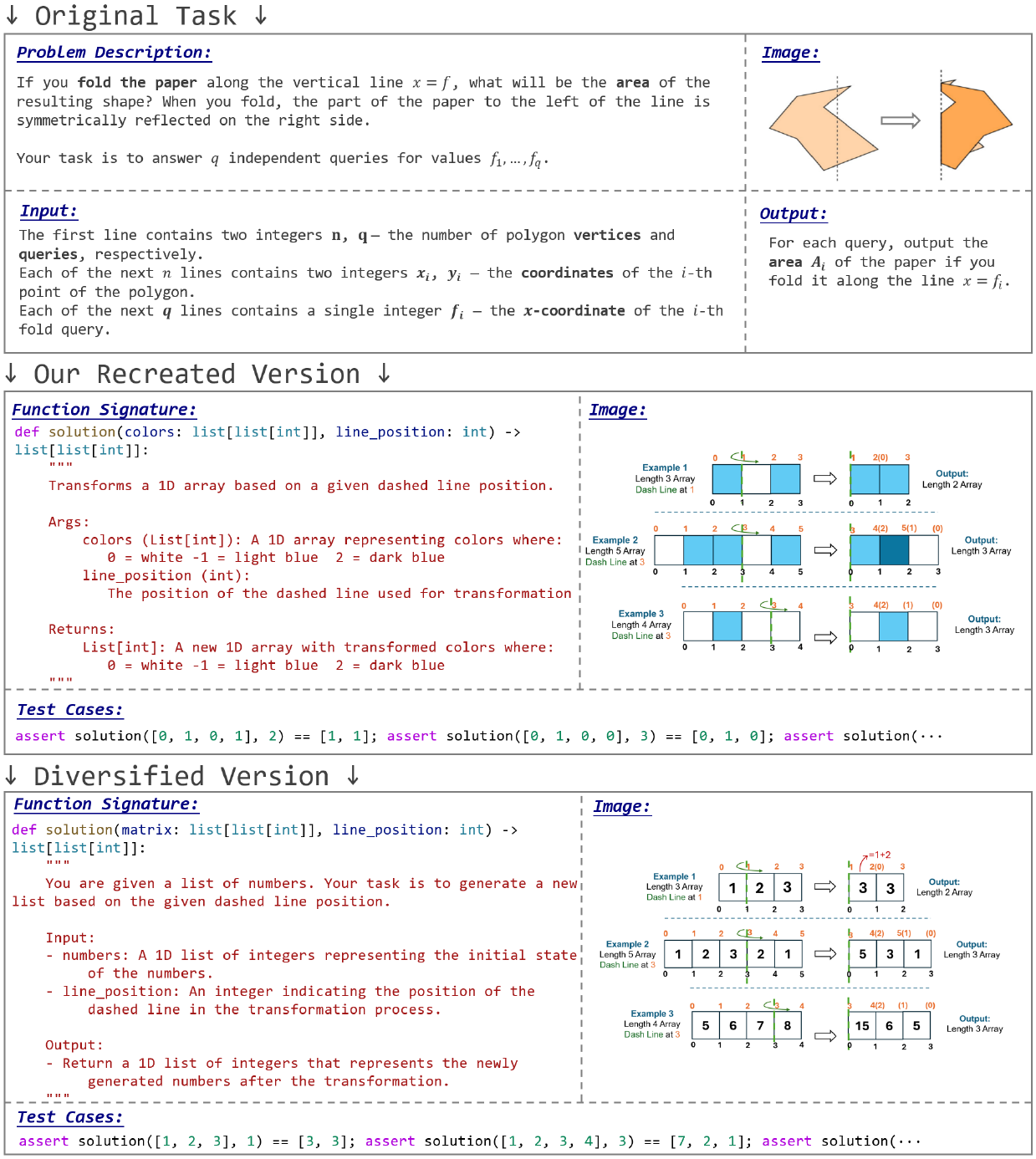}
\vspace{-0.5em}
\caption{Task annotation examples illustrating the recreation and diversification applied to the screened coding problem. The original problem is sourced from CodeForces (\url{https://codeforces.com/problemset/problem/1381/E}).}
\label{fig:annotation_example_2}
\end{figure*}

\begin{figure*}[t]
\centering
\includegraphics[width=\linewidth]{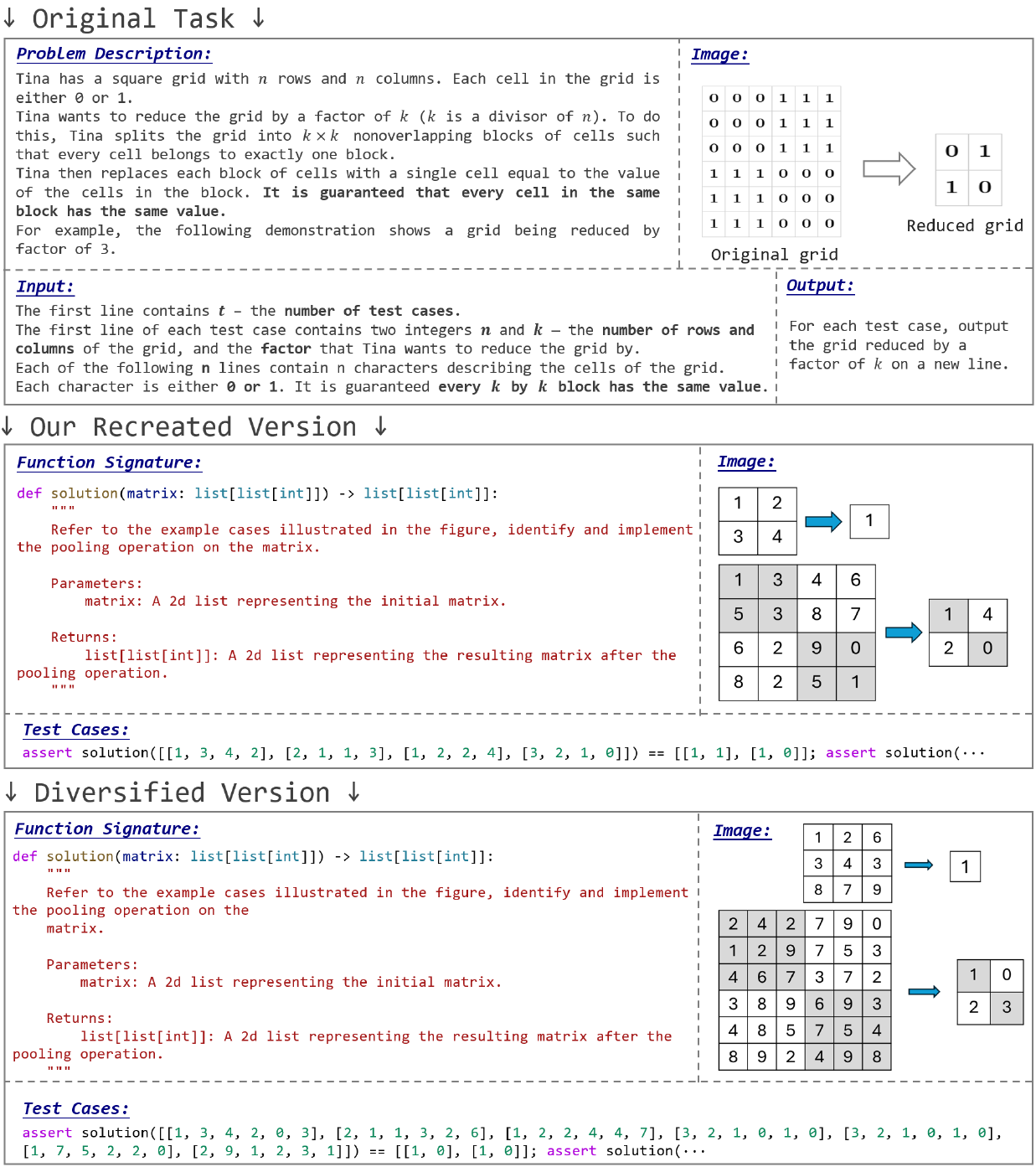}
\vspace{-0.5em}
\caption{Task annotation examples illustrating the recreation and diversification applied to the screened coding problem. The original problem is sourced from CodeForces (\url{https://codeforces.com/problemset/problem/1996/B}).}
\label{fig:annotation_example_3}
\end{figure*}

We present three examples in Figure~\ref{fig:annotation_example_1}, Figure~\ref{fig:annotation_example_2}, and Figure~\ref{fig:annotation_example_3} to demonstrate our recreation and diversification process. Each figure is divided into three parts: the original problem that meets our screening criteria (top), the recreated coding task based on the distilled ideas (middle), and the diversified variant (bottom). Below are detailed explanations of each example:

Figure \ref{fig:annotation_example_1} showcases a Stack Overflow problem where a developer needs to draw a parallelogram using four specified points. The image illustrates the connection between these points, providing the essential information needed to solve the task. Since the text merely restates the geometric properties shown in the image, we significantly reduce the textual content without losing crucial details. For recreation, we transform this into a five-pointed star problem, enriching the visual information with four examples showing different point connection patterns. The new function signature clearly defines the implementation requirements, including objectives, input parameters, and return value constraints. Instead of generating a parallelogram image, our task focuses on determining whether two specific points should be connected, simplifying the implementation while maintaining emphasis on visual reasoning. For diversification, we modify the visual pattern from a five-pointed to a six-pointed star while maintaining the same function signature.

Figure \ref{fig:annotation_example_2} presents a CodeForces problem involving polygon folding and area calculation. The image demonstrates the folding process along dashed lines, showing both initial and final states. For recreation, we simplify this into a matrix folding task where overlapping sections produce color changes. The input matrix uses two initial colors (white and light blue), which can result in three distinct outcomes after folding (white, light blue, and dark blue). Three illustrative examples clarify the folding mechanics. For diversification, we replace the color addition rule with numeric addition, requiring models to process numerical changes before and after folding.

Figure \ref{fig:annotation_example_3}, also from CodeForces, involves grid reduction following a specific pattern. The image effectively communicates the step-by-step transformation process. For recreation, we enhance the complexity by removing the reduction factor $k$ as a parameter, requiring models to deduce that $k=2$ from the provided examples. We transform the original binary scaling operation into a statistical pooling operation (e.g., minimum value computation), demanding both OCR capabilities and advanced visual reasoning. For diversification, we increase the pooling stride from 2 to 3, requiring models to analyze larger matrices. Test cases are adjusted accordingly to maintain consistency with the modified patterns.

In addition to the three examples above, we provide further examples of how we perform diversification across specific capability aspects in Table~\ref{tab:diversify}.

\section{More details on Experimental Setup}
\subsection{Evaluated Models}
\begin{table*}[t]
  \centering
  \scalebox{0.85}{
  \begin{tabular}{lll}
    \toprule
    \textbf{Models} & \textbf{Params} & \textbf{Links} \\
    \midrule

    \rowcolor[gray]{.90} \multicolumn{3}{c}{{Proprietary LMMs}} \\
    OpenAI o1~\citep{openaio1} & - & \href{https://openai.com/o1/}{OpenAI o1} \\
    GPT-4o (0806)~\citep{gpt4o} & - & \href{https://platform.openai.com/docs/models/gpt-4o}{OpenAI GPT-4o} \\
    GPT-4o-mini (0718)~\citep{gpt4o} & - & \href{https://platform.openai.com/docs/models/gpt-4o}{OpenAI GPT-4o-mini} \\
    Claude 3.5 Sonnet (1022)~\citep{claude35sonnet} & - & \href{https://docs.anthropic.com/claude/docs}{Anthropic Claude} \\
    Gemini 1.5 Pro (002)~\citep{gemini15pro} & - & \href{https://ai.google.dev/gemini-api/docs/models/gemini}{Google Gemini 1.5 Pro} \\
    Gemini 1.5 Flash (002)~\citep{gemini15pro} & - & \href{https://ai.google.dev/gemini-api/docs/models/gemini}{Google Gemini 1.5 Flash} \\
    \midrule
    \rowcolor[gray]{.90} \multicolumn{3}{c}{ {Open-weight LMMs with more than 70B parameters}} \\
    Pixtral~\citep{agrawal2024pixtral} & 124B & \href{https://huggingface.co/mistralai/Pixtral-Large-Instruct-2411}{mistralai/Pixtral-Large-Instruct-2411} \\
    Llama-3.2-V 90B~\citep{llama3d2v} & 88.8B & \href{https://huggingface.co/meta-llama/Llama-3.2-90B-Vision-Instruct}{meta-llama/Llama-3.2-90B-Vision-Instruct} \\
    InternVL 2.5 78B~\citep{chen2024expanding} & 78.4B & \href{https://huggingface.co/OpenGVLab/InternVL2_5-78B}{OpenGVLab/InternVL2-5-78B} \\
    Owen2 VL 72B~\citep{wang2024qwen2} & 73.4B & \href{https://huggingface.co/Qwen/Qwen2-VL-72B-Instruct}{Qwen/Qwen2-VL-72B-Instruct} \\ 
    QVQ-72B-Preview~\citep{qvq-72b-preview} & 73.4B & \href{https://huggingface.co/Qwen/QVQ-72B-Preview}{Qwen/QVQ-72B-Preview} \\
    Molmo-D 72B~\citep{deitke2024molmo} & 73.3B & \href{https://huggingface.co/allenai/Molmo-72B-0924}{allenai/Molmo-72B-0924} \\
    LLaVA-OV 72B~\citep{li2024llava} & 73.2B & \href{https://huggingface.co/llava-hf/llava-onevision-qwen2-72b-ov-chat-hf}{llava-hf/llava-onevision-qwen2-72b-ov-chat-hf} \\
    \midrule
    \rowcolor[gray]{.90} \multicolumn{3}{c}{ {Open-weight LMMs with fewer than 70B parameters}} \\
    Chameleon 30B~\citep{team2024chameleon} & 34.3B & \href{https://huggingface.co/facebook/chameleon-30b}{facebook/chameleon-30b} \\
    InternVL 2.5 26B~\citep{chen2024expanding} & 25.5B & \href{https://huggingface.co/OpenGVLab/InternVL2_5-26B}{OpenGVLab/InternVL2-5-26B} \\
    Pixtral 12B~\citep{agrawal2024pixtral} & 12.0B & \href{https://huggingface.co/mistralai/Pixtral-12B-2409}{mistralai/Pixtral-12B-2409} \\
    Llama-3.2-V 11B~\citep{llama3d2v} & 10.7B & \href{https://huggingface.co/meta-llama/Llama-3.2-11B-Vision-Instruct}{meta-llama/Llama-3.2-11B-Vision-Instruct} \\
    Qwen2 VL 7B ~\citep{wang2024qwen2} & 8.3B & \href{https://huggingface.co/Qwen/Qwen2-VL-7B-Instruct}{Qwen/Qwen2-VL-7B-Instruct} \\
    InternVl 2.5 8B~\citep{chen2024expanding} & 8.1B & \href{https://huggingface.co/OpenGVLab/InternVL2_5-8B}{OpenGVLab/InternVL2-5-8B} \\
    LLaVA-OV 7B~\citep{li2024llava} & 8.03B & \href{https://huggingface.co/llava-hf/llava-onevision-qwen2-7b-ov-chat-hf}{llava-hf/llava-onevision-qwen2-7b-ov-chat-hf} \\
    Molmo-D 7B~\citep{deitke2024molmo} & 8.02B & \href{https://huggingface.co/allenai/Molmo-7B-D-0924}{allenai/Molmo-7B-D-0924} \\
    Chameleon 7B ~\citep{team2024chameleon} & 7.04B & \href{https://huggingface.co/facebook/chameleon-7b}{facebook/chameleon-7b} \\
    Phi-3.5-V 4B ~\citep{phi3_5} & 4.2B & \href{https://huggingface.co/microsoft/Phi-3.5-vision-instruct}{microsoft/Phi-3.5-vision-instruct} \\
    InternVL 2.5 4B ~\citep{chen2024expanding} & 3.7B & \href{https://huggingface.co/OpenGVLab/InternVL2_5-4B}{OpenGVLab/InternVL2-5-4B} \\
    \midrule
    \rowcolor[gray]{.90} \multicolumn{3}{c}{ {Open-Weight Code LLM}} \\
    Qwen2.5 Coder 32B \citep{hui2024qwen2}  & 32.8B & \href{https://huggingface.co/Qwen/Qwen2.5-Coder-32B-Instruct}{Qwen/Qwen2.5-Coder-32B-Instruct} \\
    \bottomrule
    \end{tabular}
    }
    \caption{List of LMMs with their parameter sizes and links to the official reports or Huggingface repositories.}
    \label{tab:model_details}
\end{table*}
In Table~\ref{tab:model_details}, we provide a detailed list of Large Multimodal Models (LMMs) used in our experiments. For each model, we specify the number of parameters and include direct links to relevant reports or Huggingface repositories for further reference.

\subsection{Prompt Templates}
\label{appendix:prompt_design}
\begin{figure*}[t]
\centering
\includegraphics[width=\linewidth]{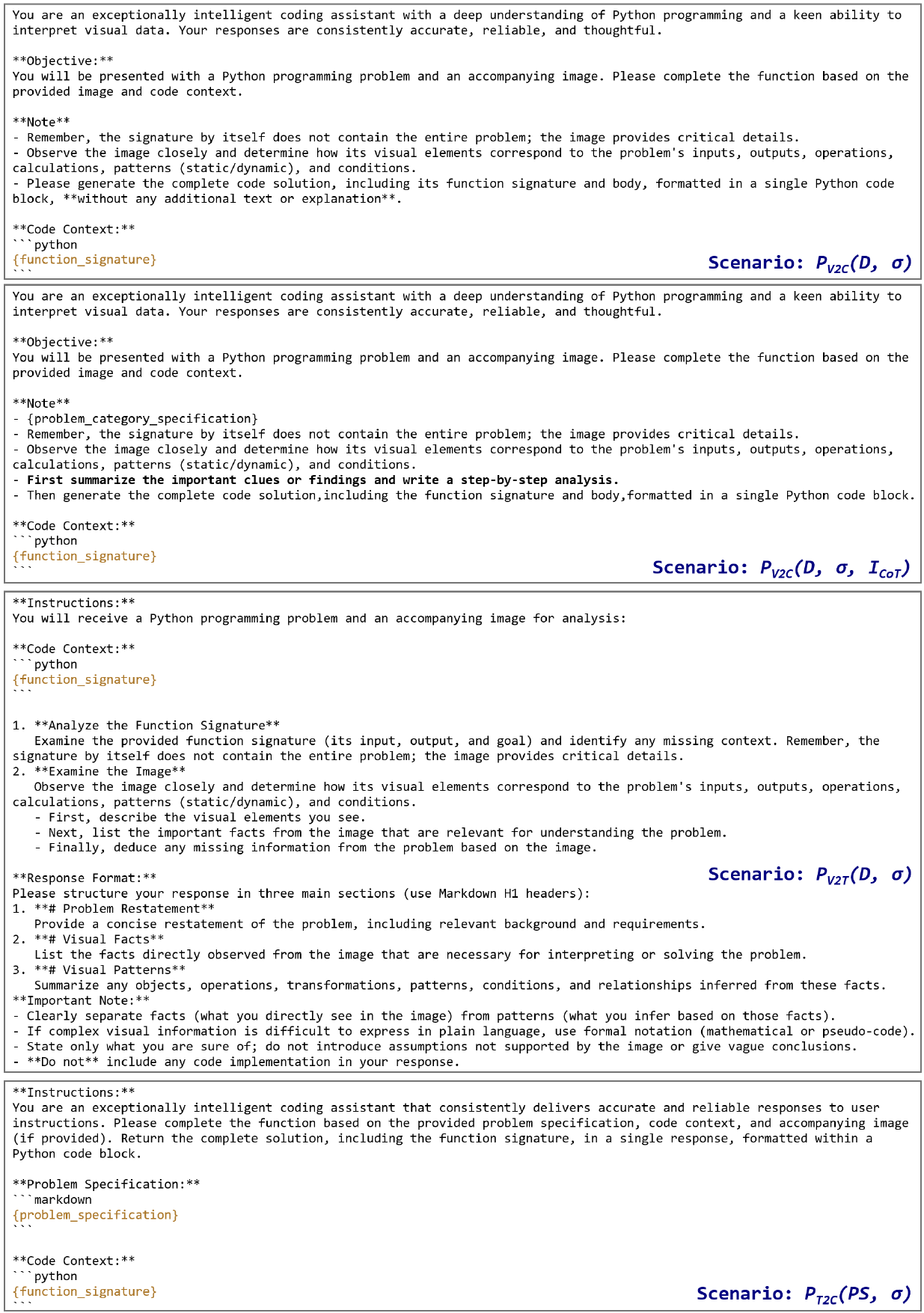}
\vspace{-0.5em}
\caption{Prompting templates used for the four scenarios introduced in Section~\ref{sec:benchmark_setup}. \{function\_signature\} and \{problem\_specification\} serve as placeholders for the respective content.}
\label{fig:prompt_template}
\end{figure*}

\begin{figure*}[t]
\centering
\includegraphics[width=0.9\linewidth]{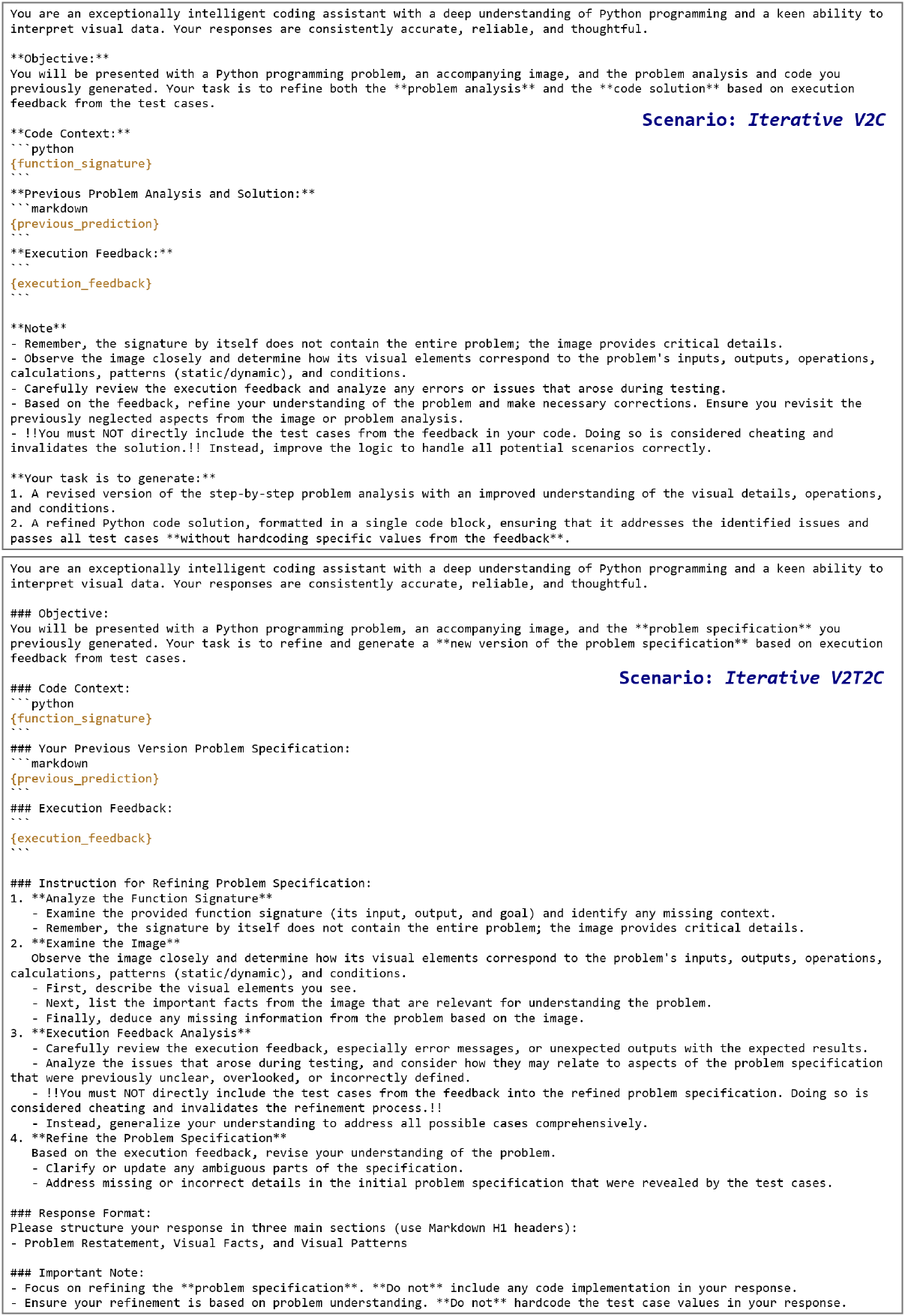}
\vspace{-0.5em}
\caption{Prompting templates used for the iterative benchmarking scenarios introduced in Section~\ref{sec:benchmarking_results}. \{function\_signature\}, \{previous\_prediction\}, and \{execution\_feedback\} serve as placeholders for the respective content.}
\label{fig:iterative_prompt_template}
\end{figure*}

\begin{figure*}[t]
\centering
\includegraphics[width=\linewidth]{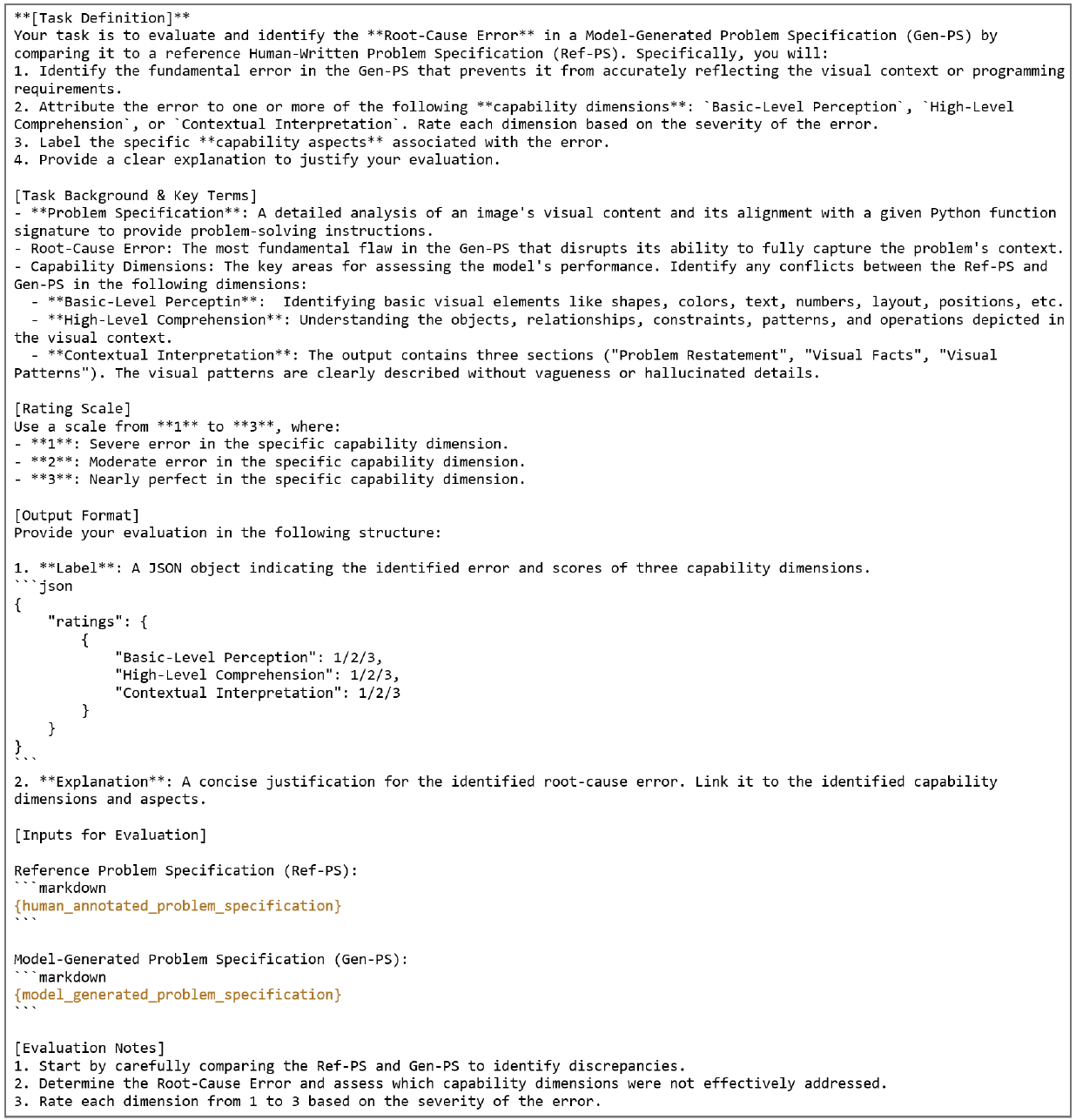}
\vspace{-0.5em}
\caption{Prompting templates used for the LLM-as-Judge rating experiment introduced in Section~\ref{sec:analysis_exp}. \{human\_annotated\_problem\_specification\} and \{model\_generated\_problem\_specification\} serve as placeholders for the respective content.}
\label{fig:lmjudge_prompt_template}
\end{figure*}

We designed three main sets of prompts for the experiments. The first set is used for the evaluation pipelines in the main benchmarking experiments, covering scenarios such as Vision-to-Code, Vision-to-Code with Chain-of-Thought (CoT), Vision-to-Text, and Text-to-Code, as described in Section~\ref{sec:benchmark_setup}. The corresponding prompts for these scenarios are listed in Figure~\ref{fig:prompt_template}.
The second set of prompts is used in the iterative refinement experiments, introduced in Section~\ref{sec:benchmarking_results}. These prompts address scenarios where code or previously generated textual problem specifications are refined based on feedback from the execution environment. The relevant prompts for this scenario are provided in Figure~\ref{fig:iterative_prompt_template}.
The third scenario involves using GPT-4o as a judge to rate the diagram descriptions generated by LMMs. The prompt used is shown in Figure~\ref{fig:lmjudge_prompt_template}.

Our prompt design has undergone multiple rounds of optimization to address specific issues we encountered. For instance, we instruct the model to follow a markdown format code block for generation and avoid generating multiple code blocks to streamline post-processing and improve parsing success rates. We also provide detailed instructions for prompting LMMs to generate diagram descriptions in a structured problem specification format, including \textit{Problem Restatement}, \textit{Visual Facts}, and \textit{Visual Patterns}, ensuring a comprehensive capture and expression of the visual context. In the iterative refinement setting, we specifically instruct LMMs not to hardcode test cases into their generated code to ensure that improvements stem from an enhanced understanding of the problem. Additionally, for LLM-as-Judge experiments, we list clear steps for rating the LMMs' outputs, promoting more robust and reliable rating results.

\subsection{Ablation on Temperature}
\label{appendix:ablation_on_temperature}
\begin{table}[t]
  \centering
  \scalebox{0.85}{
      \begin{tabular}{lcccc}
      \toprule
      \textbf{Models} & \textbf{T=0.4} & \textbf{T=0.6} & \textbf{T=0.8} &  \textbf{T=1} \\
      \midrule
      \grayline \multicolumn{5}{c}{Proprietary LMMs} \\
      Claude 3.5 Sonnet & 48.3 & 46.9 & 48.1 & 47.9 \\  
      GPT-4o & 44.9 & 42.2 & 43.6 & 44.9 \\ 
      Gemini 1.5 Pro & 41.1 & 39.2 & 39.4 & 41.6 \\ 
      Gemini 1.5 Flash & 27.1 & 30 & 28.4 & 29.1 \\ 
      GPT-4o-mini & 27.9 & 29 & 29.9 & 32.1 \\  
      \grayline \multicolumn{5}{c}{Open-weight LMMs} \\
      Pixtral 124B & 35.6 & 39.8 & 34.2 & 37 \\ 
      InternVL 2.5 78B & 37.2 & 36.1 & 35.9 & 36.2 \\ 
      Qwen2 VL 72B & 31.1 & 29.1 & 28.7 & 31.5 \\ 
      LLaVA-OV 72B & 25 & 22.7 & 24.1 & 22.2 \\ 
      Molmo-D 72B & 16.9 & 13.2 & 14.4 & 14.5 \\  
      Llama-3.2-V 90B & 12.5 & 10.7 & 12.4 & 13.7 \\  
      Pixtral 12B & 21.2 & 23.4 & 23.2 & 21.1 \\  
      InternVL 2.5 26B & 20.9 & 21 & 20.7 & 21.8 \\ 
      Qwen2 VL 7B & 13.3 & 17.2 & 16.6 & 18.5 \\  
      InternVL 2.5 8B & 13.6 & 16.6 & 17.3 & 16.3 \\  
      InternVL 2.5 4B & 16.9 & 17.6 & 15.2 & 20.5 \\  
      LLaVA-OV 7B & 15.1 & 15.2 & 13 & 15.2 \\  
      Phi-3.5-V 4B & 5.6 & 9.3 & 9.1 & 8.7 \\ 
      Llama-3.2-V 11B & 7.5 & 10.2 & 9.6 & 9.9 \\ 
      Molmo-D 7B & 9.2 & 10 & 11.2 & 11.2 \\  
      Chameleon 7B & 1 & 2.5 & 2.5 & 2.1 \\ 
      Chameleon 30B & 2 & 0.5 & 3.3 & 2 \\  
      \bottomrule
      \end{tabular}
  }
  \caption{Ablation on LMMs' sampling temperature for the pass@3 results under the \textit{V2T2C w/ GPT-4o} settings.}
  \label{table:temperature_ablation}
\end{table}
As described in Section~\ref{sec:benchmark_setup}, we set the sampling temperature to $T = 0.8$ for generating multiple predictions, following established practices in code generation benchmarking~\citep{chen2021evaluating, chen2022codet}. Given that LMMs may exhibit varying performance at different temperatures, we conduct an ablation study to assess the rationale behind this choice. Specifically, we evaluated all 22 LMMs under the \textit{V2T2C w/ GPT-4o} setting across a range of temperatures from 0.4 to 1.0. The results are presented in Table~\ref{table:temperature_ablation}, which indicate that LMMs generally demonstrate consistent performance across these settings, with a few models showing slight variations, further validating the rationale for our chosen temperature setting. 

\section{Deeper Analysis on \ourbench}
\label{appendix:deeper_analysis}
\subsection{Co-occurrence of Capability Aspects}
\begin{figure*}[t]
\centering
\includegraphics[width=\linewidth]{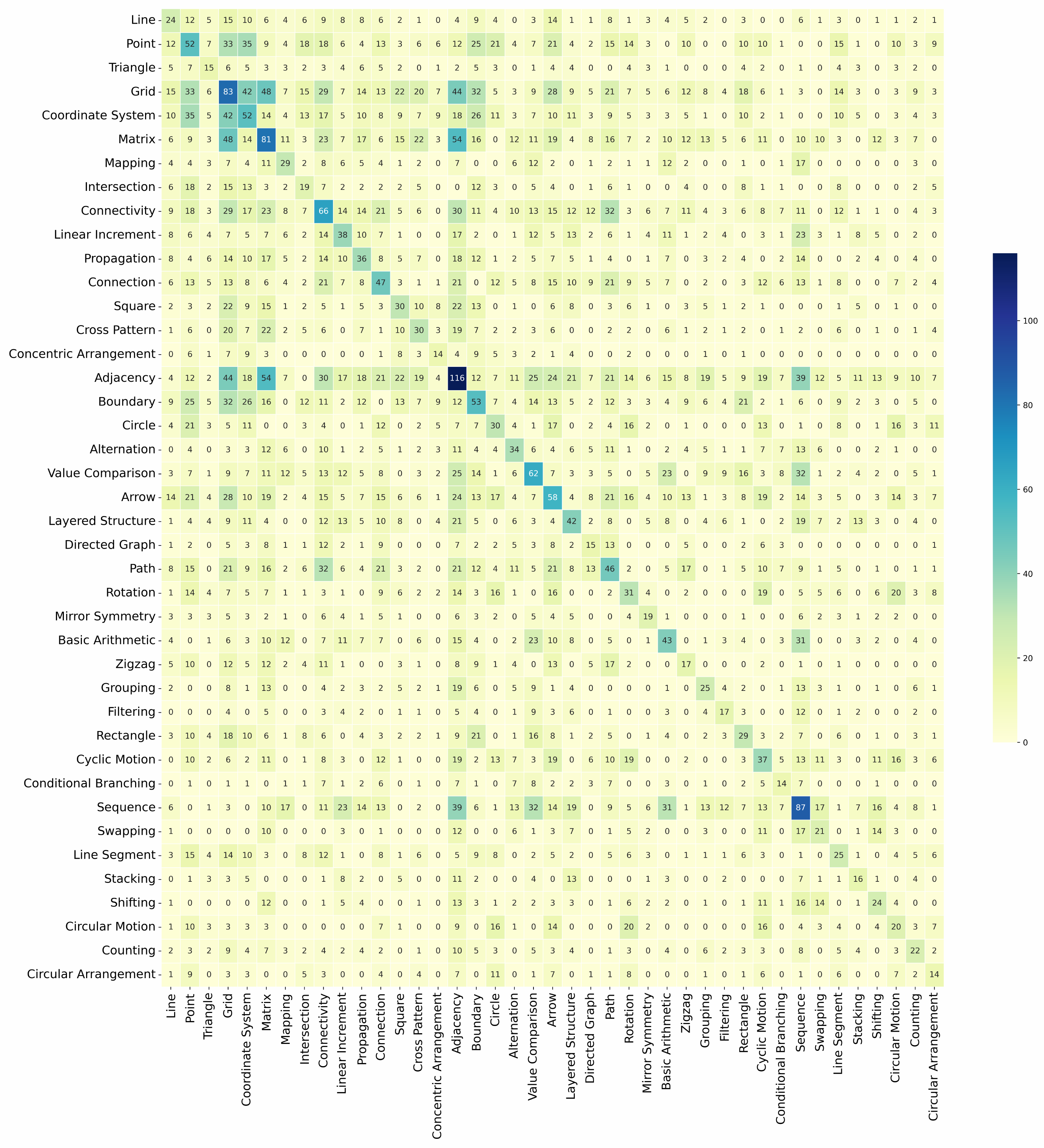}
\vspace{-0.5em}
\caption{Analysis of the capability aspect co-occurrences in \ourbench tasks.}
\label{fig:label_cooccurrence}
\end{figure*}

As illustrated in Figure~\ref{fig:diverse_labels}, the diagrams in \ourbench encompass a wide range of capability aspects that require human-level intelligence for interpretation. Each diagram typically involves multiple capability aspects to be fully understood.
To explore the relationships between these capability aspects, we conduct a co-occurrence analysis based on the aspect labels assigned by human annotators during task annotation. The results, presented in Figure~\ref{fig:label_cooccurrence}, show a heatmap where each value represents the number of tasks in \ourbench that involve both corresponding aspect labels.
Our analysis reveals that the diagrams in \ourbench exhibit a diverse distribution of capability aspects. Among them, \textit{adjacency}, \textit{grid}, \textit{matrix}, and \textit{sequence} are the most frequently occurring labels. Common co-occurrences include \textit{matrix-adjacency}, \textit{grid-boundary}, \textit{grid-path}, and \textit{sequence-linear increment}, highlighting the fundamental spatial and structural relationships embedded in these diagrams.

\subsection{Comparison of Strong Coder Models}
\begin{figure}[t]
    \centering
    \includegraphics[width=0.85\linewidth]{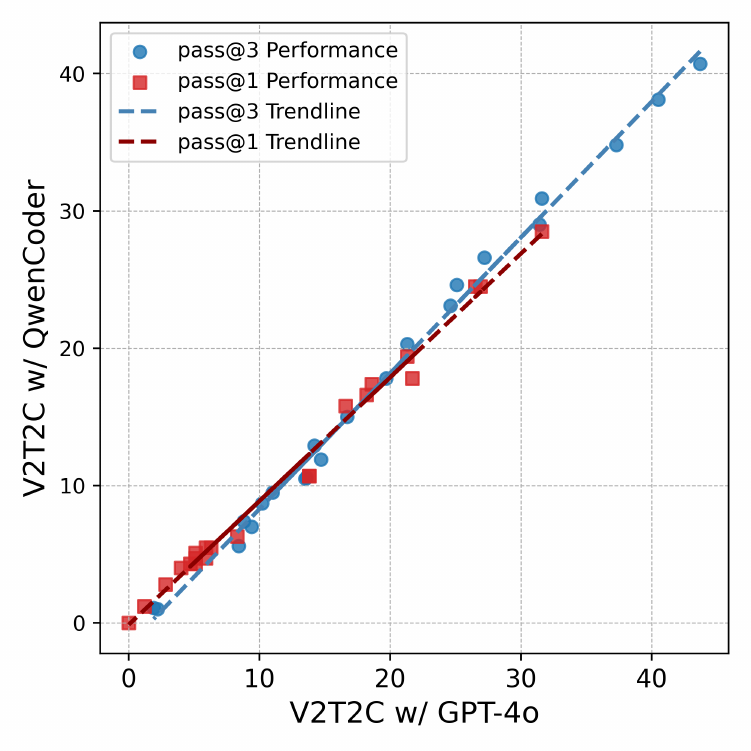}
    \vspace{-0.8em}
    \caption{Performance comparison between GPT-4o and QwenCoder-32B as strong coders under the \textit{V2T2C w/ SC} setting.}
    \label{fig:qwencoder_result}
\end{figure}

\begin{figure}[t]
    \centering
    \includegraphics[width=0.85\linewidth]{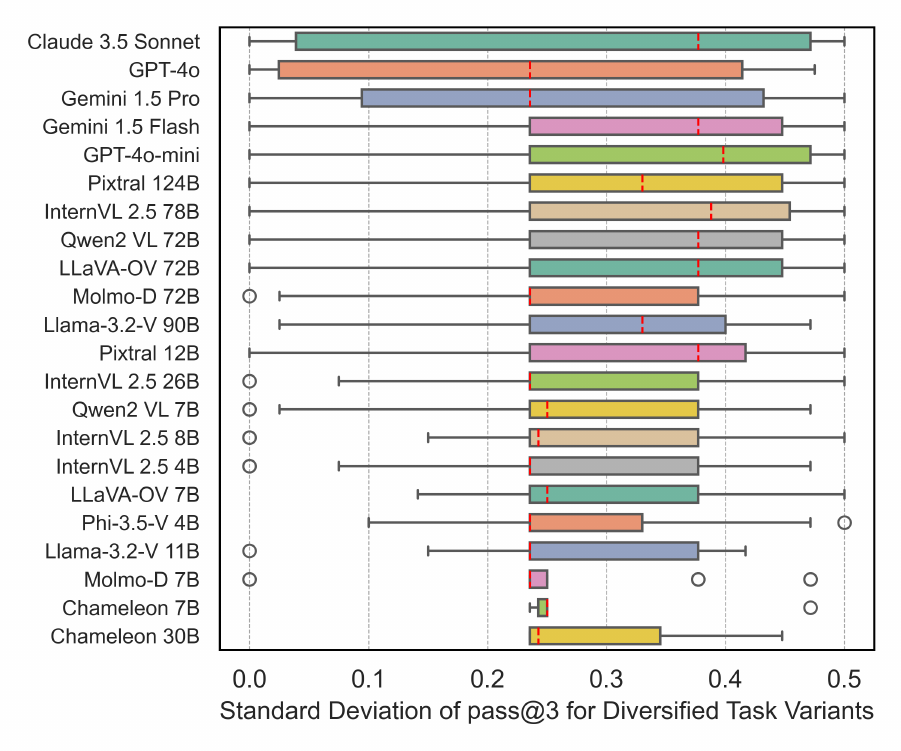}
    \vspace{-0.8em}
    \caption{Effect of task diversification on performance variation across models in the \textit{V2T2C w/ SC} setting.}
    \label{fig:std_diversified_version}
\end{figure}

We use GPT-4o as the primary strong coder for our benchmarking experiments, leveraging its superior coding capabilities to translate problem specifications generated by LMMs into code. This allows us to focus on the evaluation of LMMs' visual understanding ability in a more controllable manner. Our evaluation pipeline is designed to be robust, accommodating different strong coders. To test the stability of our results, we perform an ablation study by replacing GPT-4o with an open-weight LLM, Qwen2.5-Coder-32B-Instruct~\cite{hui2024qwen2}, referred to as QwenCoder-32B. QwenCoder-32B demonstrates comparable coding performance to GPT-4o, as evidenced by LiveCodeBench~\cite{jain2024livecodebench}. This ablation allows us to explore whether switching to a different strong coder leads to deviations in our findings.

The ablation is conducted under the \textit{V2T2C w/ SC} setting, where QwenCoder-32B replaces GPT-4o to generate code based on the problem specifications provided by LMMs. The results, shown in Figure~\ref{fig:qwencoder_result}, reveal that GPT-4o and QwenCoder-32B exhibit near-perfect correlations, demonstrating the strong stability of our evaluation methodology.

\subsection{The Effect of Diversified Tasks}
\begin{figure*}[t]
\begin{minipage}[b]{0.31\linewidth}
    \centering
    \includegraphics[width=\linewidth]{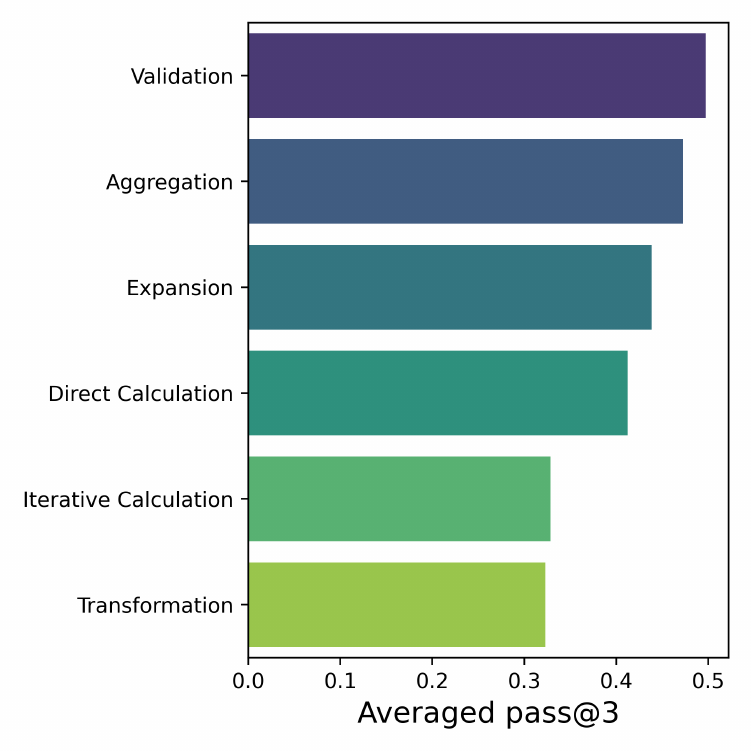}
    \vspace{-0.8em}
    \caption{Pass rates of LMMs across different task types.}
    \label{fig:psrate_by_task_type}
\end{minipage}
\hspace{0.2cm}
\begin{minipage}[b]{0.31\linewidth}
    \centering
    \includegraphics[width=\linewidth]{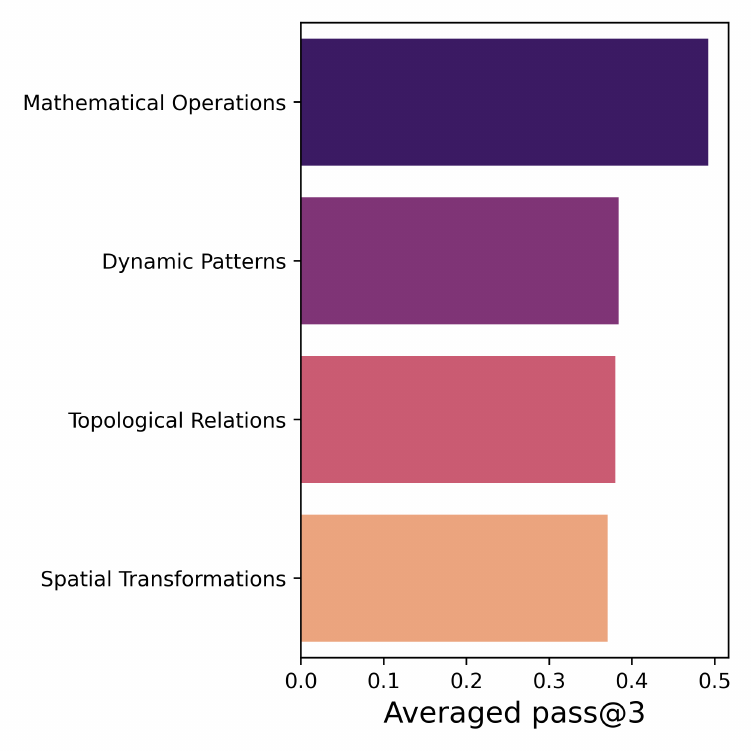}
    \vspace{-0.8em}
    \caption{Pass rates of LMMs across main capability dimensions.}
    \label{fig:psrate_by_cap_dim}
\end{minipage}
\hspace{0.2cm}
\begin{minipage}[b]{0.31\linewidth}
    \centering
    \includegraphics[width=\linewidth]{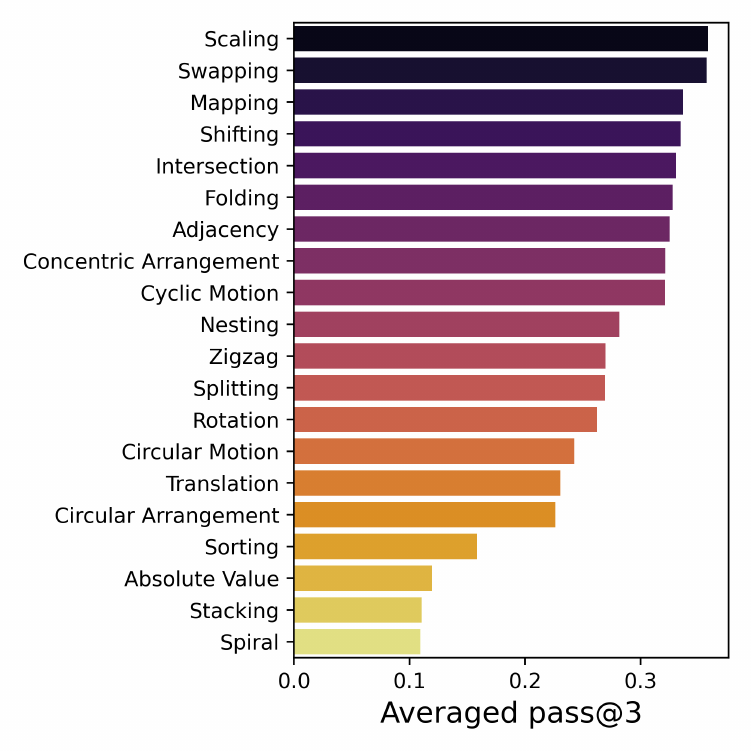}
    \vspace{-0.8em}
    \caption{Pass rates of LMMs across specific capability aspects.}
    \label{fig:psrate_by_cap_label}
\end{minipage}
\end{figure*}

As outlined in Section~\ref{sec:benchmark_construction} and Section~\ref{appendix:task_annotation}, our task annotation pipeline includes a crucial step to create diversified versions of the seed tasks, expanding both the volume and variety of tasks in \ourbench. To evaluate whether these diversified tasks introduce different challenges compared to the original seed tasks, we analyze the standard deviation of pass rates across the seed tasks and their diversified versions.

Specifically, we use the results of the 22 LMMs under the V2T2C w/ SC setting. We group tasks based on whether they are seed tasks or their variants, resulting in 100 task groups (since we have 100 seed tasks). We then calculate the standard deviation of the pass@3 results within each group, excluding groups where all tasks have a pass@3 rate of 0. The standard deviation for each group is computed as:

$$
\text{SD}_{\text{group}} = \sqrt{\frac{1}{N} \sum_{i=1}^{N} (\text{pass@3}_i - \mu_{\text{group}})^2}
$$

where $N$ is the number of tasks in the group, $\text{pass@3}_i$ is the pass rate of task $i$, and $\mu_{\text{group}}$ is the mean pass@3 for the group. The resulting distribution of pass@3 standard deviations (SD) is shown in Figure~\ref{fig:std_diversified_version} as box plots.

The results reveal that the 25th percentile of most models has an SD greater than 0.2, and the median SD is around 0.4. This demonstrates a notable performance gap between tasks within the same group, validating the effectiveness of our task diversification process.

\section{More Detailed Error Analysis}
\label{appendix:error_analysis}
\begin{figure}[t]
    \centering
    \includegraphics[width=\linewidth]{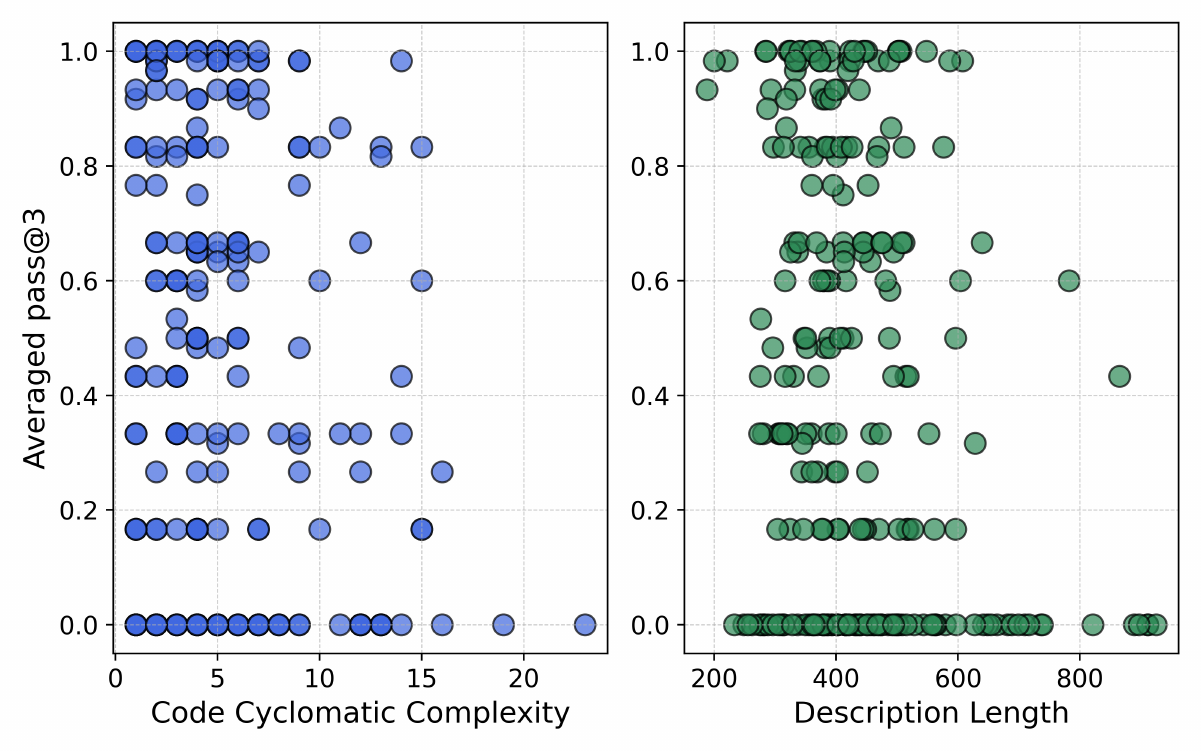}
    \vspace{-0.5em}
    \caption{Correlation Between LMM Pass Rates and Task Difficulty in Coding and Diagram Descriptions.}
    \label{fig:error_analysis_correlation}
\end{figure}

\subsection{Error Patterns and Taxonomy}

To better understand where current LMMs fall short in solving the coding tasks in \ourbench, we conduct a statistical analysis examining the correlation between pass rates and three key factors: task type, general capability dimensions, and specific capability aspects (illustrated in Figure~\ref{fig:diverse_labels}) required for understanding diagrams in \ourbench. The results are presented in Figure~\ref{fig:psrate_by_task_type}, Figure~\ref{fig:psrate_by_cap_dim}, and Figure~\ref{fig:psrate_by_cap_label}. The pass rate in our analysis is the averaged pass@3 for the five proprietary models under the \textit{V2T2C w/ SC} setting.

The analysis reveals that LMMs perform particularly poorly on tasks involving \textit{Transformation} and \textit{Iterative Calculation}, both achieving a pass@3 of approximately 32\%. This suggests that these models struggle with understanding spatial transformations and tracking state changes over iterative steps. In terms of general capability dimensions, the difference in pass rates across various categories is minimal. Specifically, \textit{Spatial Transformation}, \textit{Topological Relations}, and \textit{Dynamic Patterns} all yield an average pass@3 of 38\%.

\begin{figure*}[t]
\centering
\includegraphics[width=\linewidth]{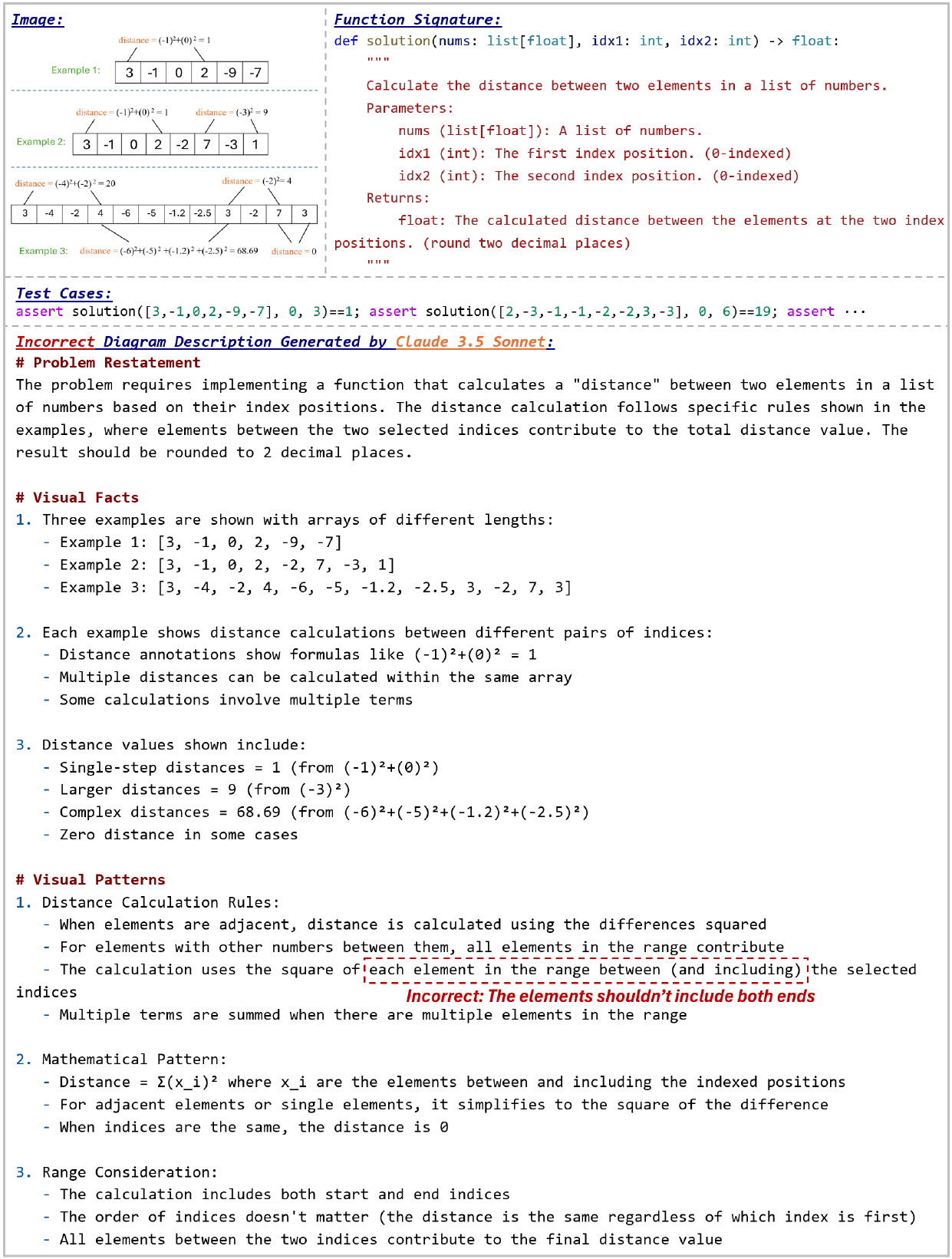}
\vspace{-0.5em}
\caption{Example error case demonstrating LMMs' challenges with data structure manipulation and mathematical operations. The case shown for Claude 3.5 Sonnet is under the \textit{V2T2C w/ SC} setting.}
\label{fig:error_case_q63_sonnet}
\end{figure*}

\begin{figure*}[t]
\centering
\includegraphics[width=0.9\linewidth]{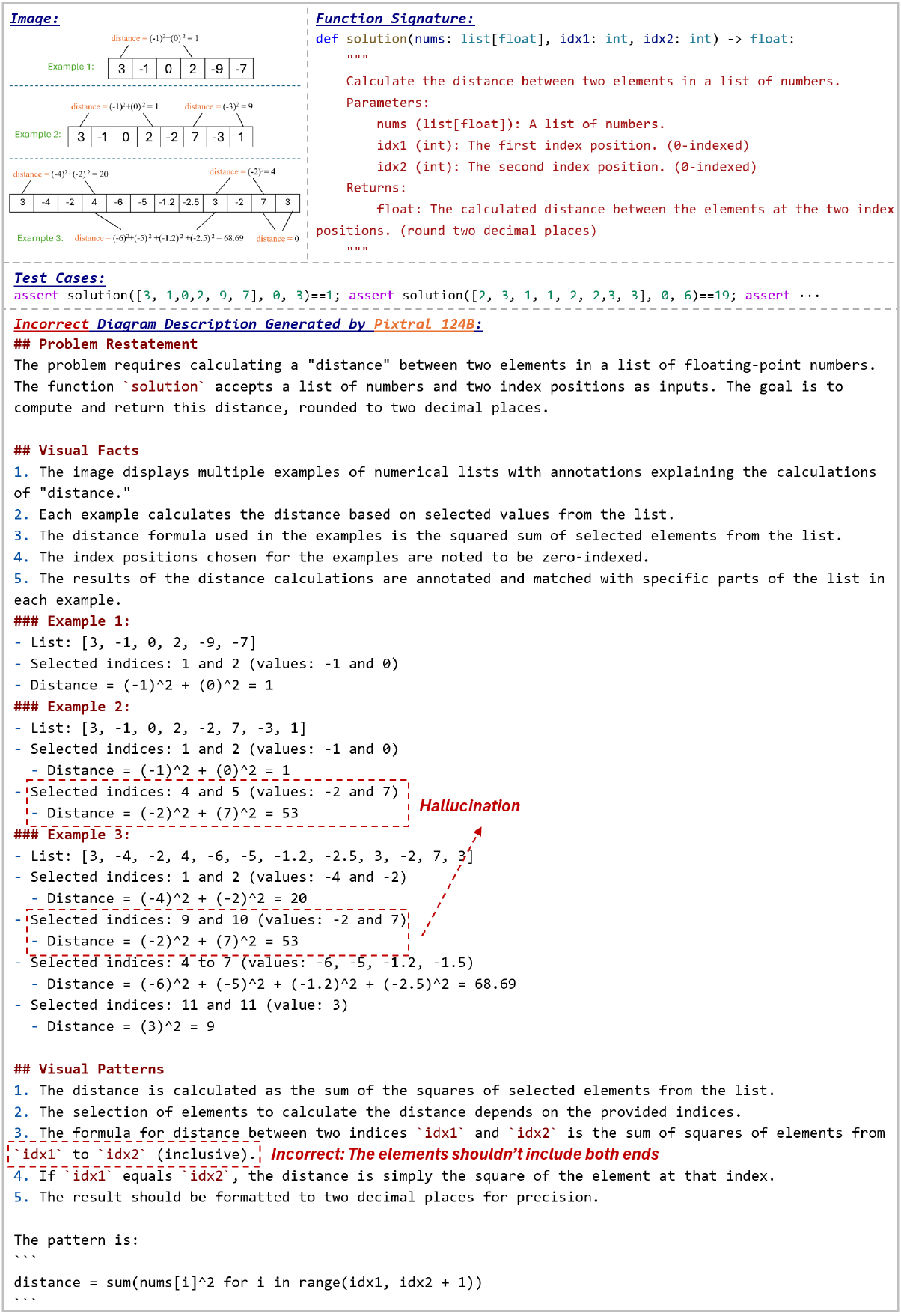}
\vspace{-0.5em}
\caption{Example case on the same task in Figure~\ref{fig:error_case_q63_sonnet}, demonstrated by Pixtral 124B under the \textit{V2T2C w/ SC} setting.}
\label{fig:error_case_q63_pixtral_large}
\end{figure*}

\begin{figure*}[t]
\centering
\includegraphics[width=\linewidth]{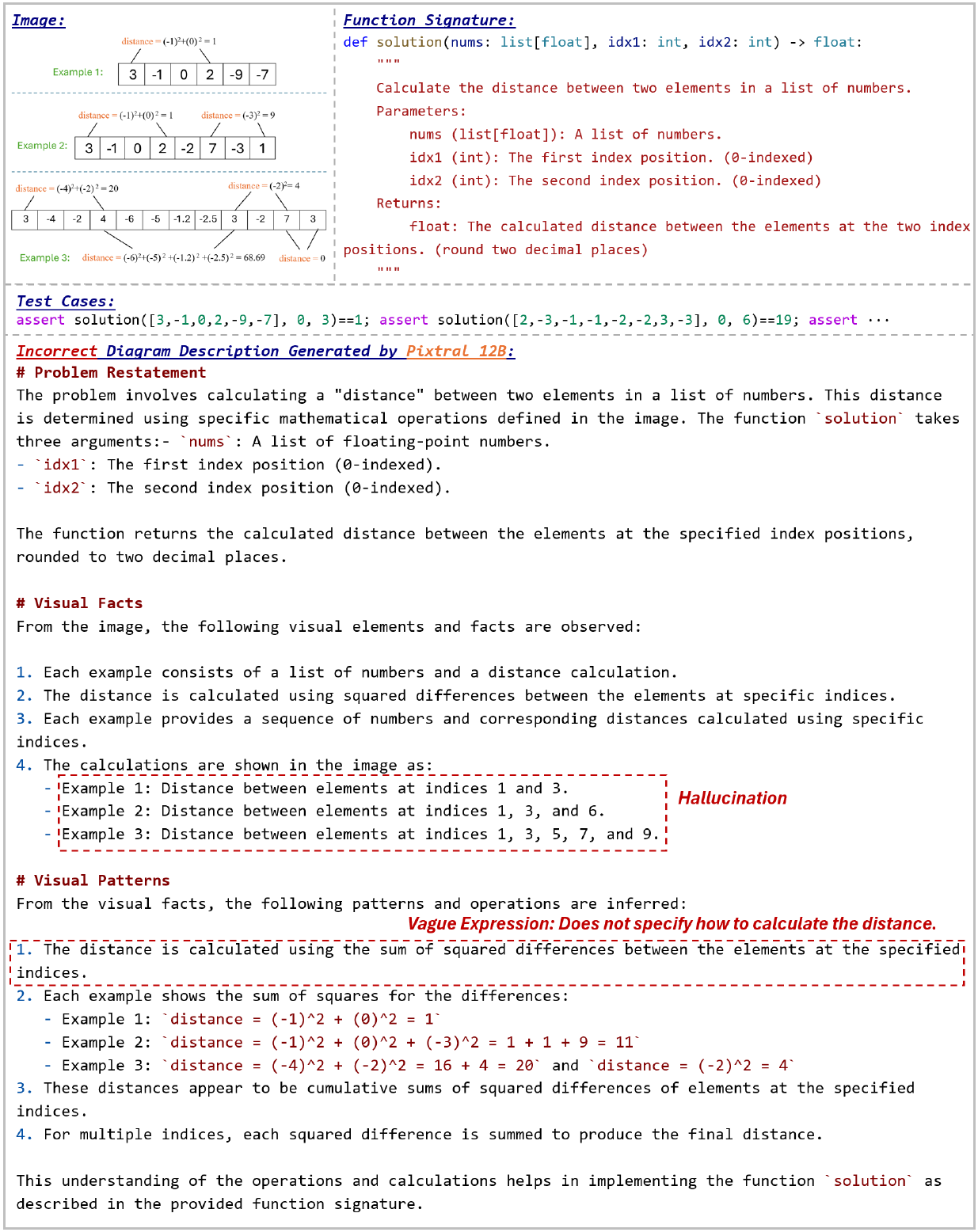}
\vspace{-0.5em}
\caption{Example case on the same task in Figure~\ref{fig:error_case_q63_sonnet}, demonstrated by Pixtral 12B under the \textit{V2T2C w/ SC} setting.}
\label{fig:error_case_q63_pixtral_12b}
\end{figure*}

\begin{figure*}[t]
\centering
\includegraphics[width=\linewidth]{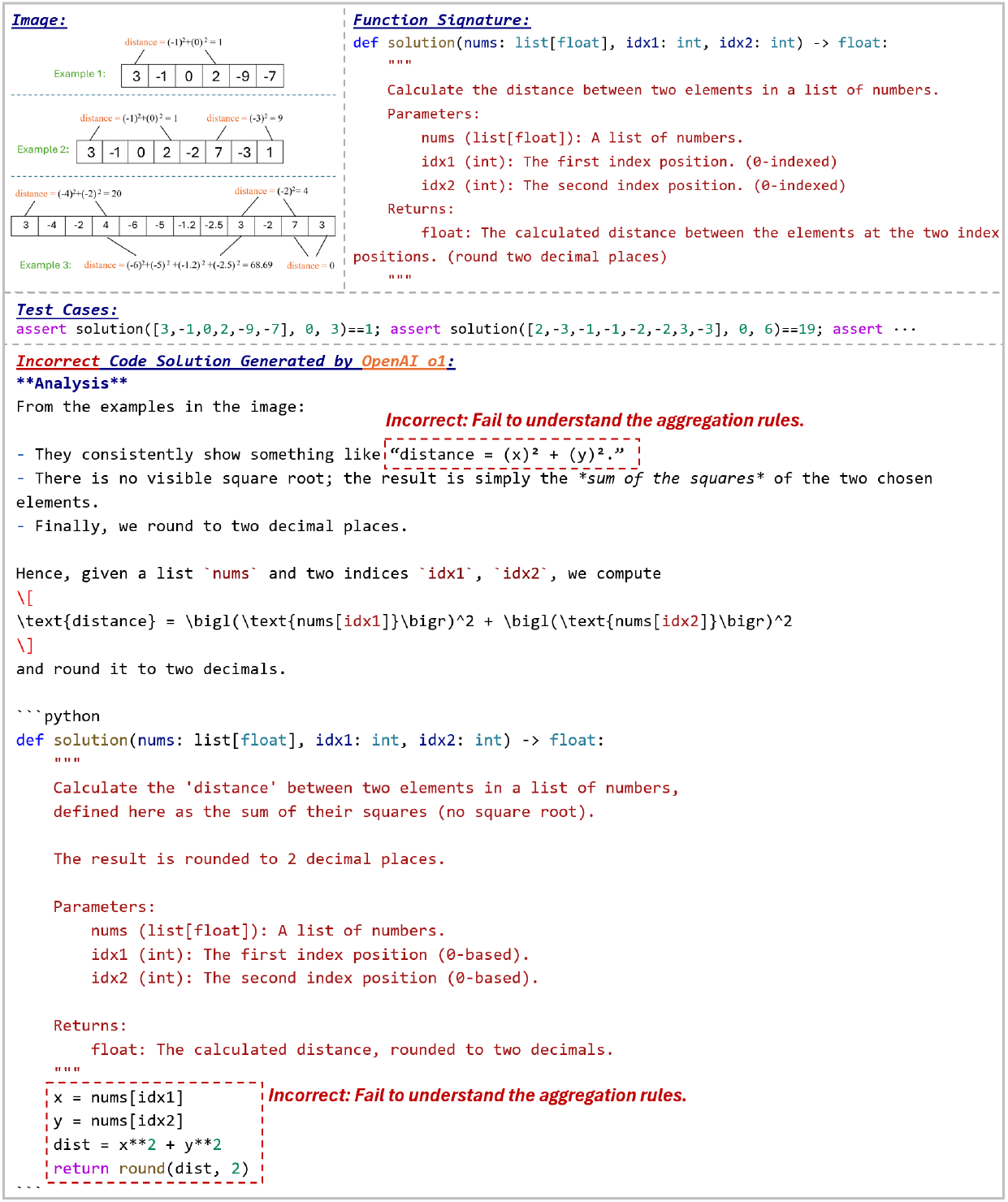}
\vspace{-0.5em}
\caption{Example case on the same task in Figure~\ref{fig:error_case_q63_sonnet}, demonstrated by OpenAI o1 under the textit{V2C w/ CoT} setting.}
\label{fig:error_case_q63_o1}
\end{figure*}

\begin{figure*}[t]
\centering
\includegraphics[width=\linewidth]{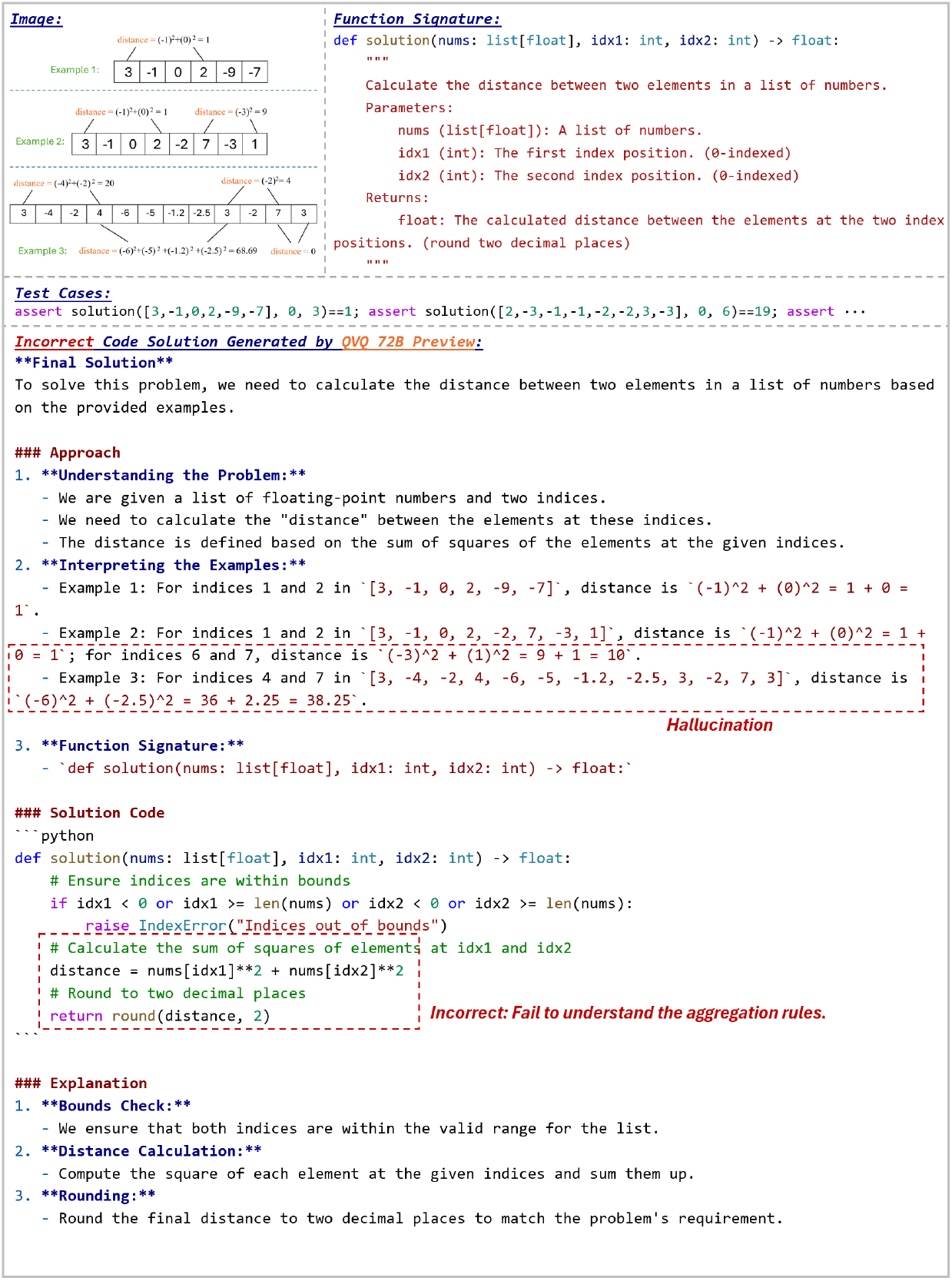}
\vspace{-0.5em}
\caption{Example case on the same task in Figure~\ref{fig:error_case_q63_sonnet}, demonstrated by QVQ 72B Preview under the textit{V2C w/ CoT} setting.}
\label{fig:error_case_q63_qvq}
\end{figure*}

\begin{figure*}[t]
\centering
\includegraphics[width=\linewidth]{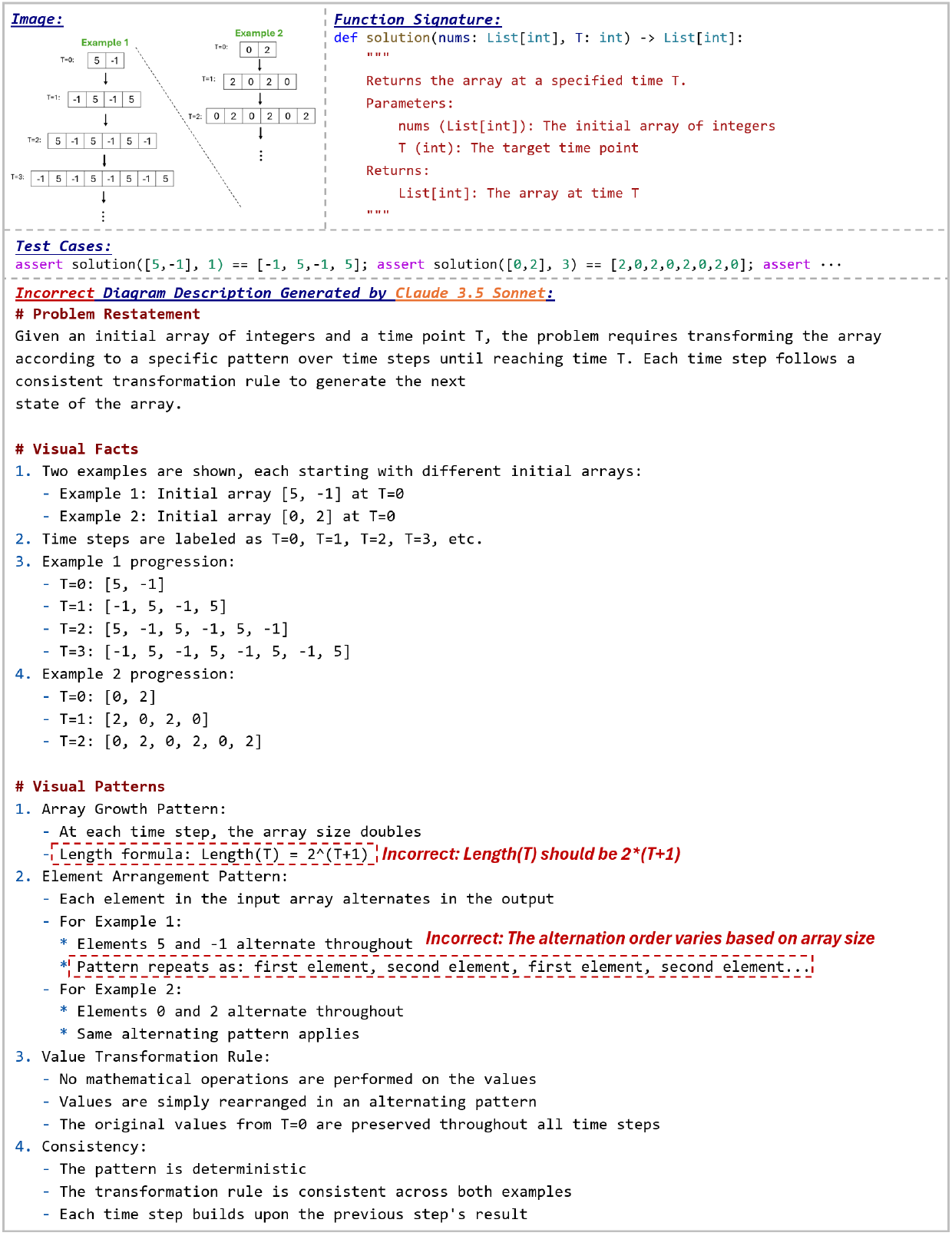}
\vspace{-0.5em}
\caption{Error case highlighting LMMs' limitations in understanding dynamic patterns with alternating elements and linear increments. The case shown for Claude 3.5 Sonnet is under the \textit{V2T2C w/ SC} setting.}
\label{fig:error_case_q73_sonnet}
\end{figure*}

\begin{figure*}[t]
\centering
\includegraphics[width=\linewidth]{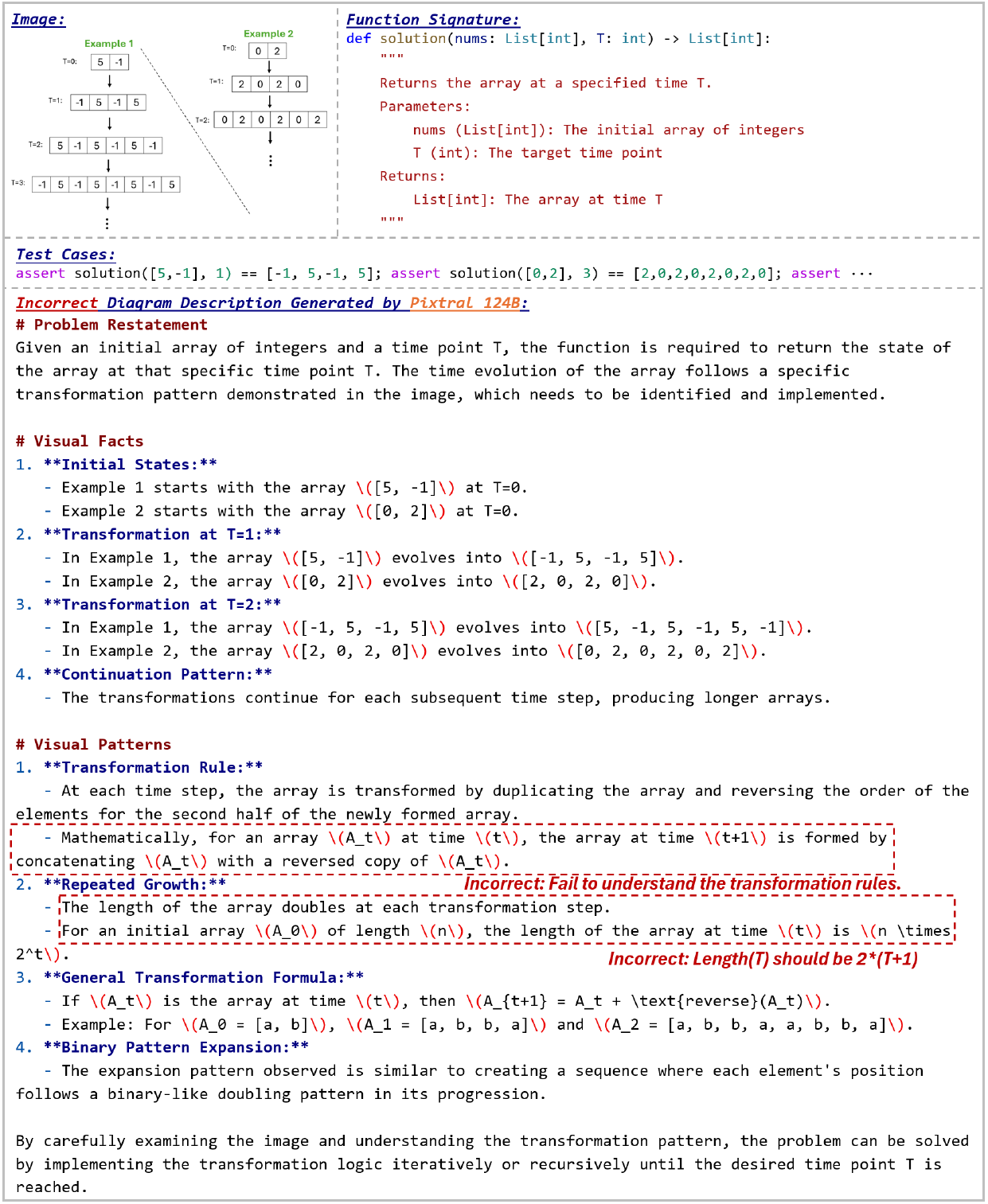}
\vspace{-0.5em}
\caption{Example case on the same task in Figure~\ref{fig:error_case_q73_sonnet}, demonstrated by Pixtral 124B under the \textit{V2T2C w/ SC} setting.}
\label{fig:error_case_q73_pixtral_large}
\end{figure*}

\begin{figure*}[t]
\centering
\includegraphics[width=\linewidth]{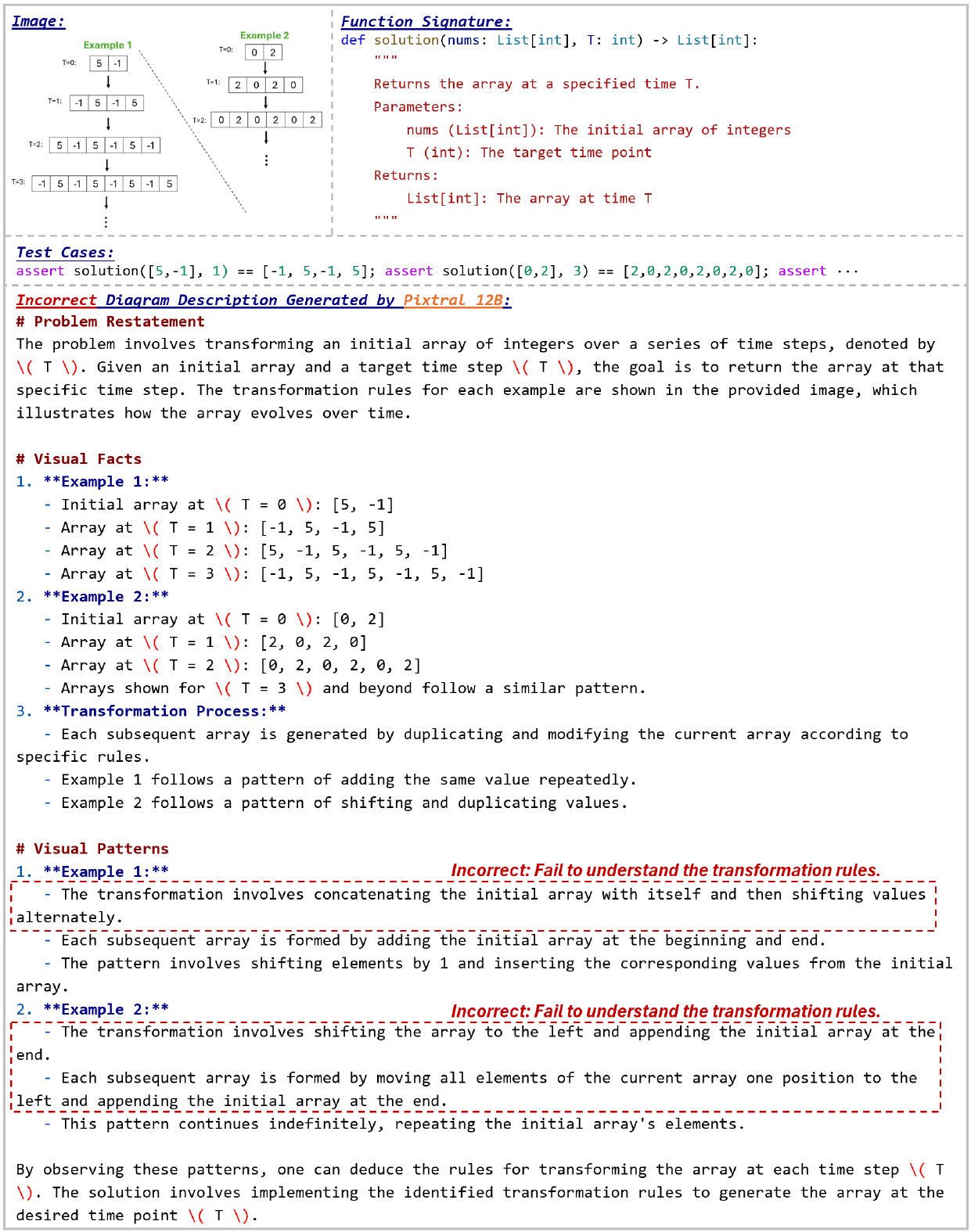}
\vspace{-0.5em}
\caption{Example case on the same task in Figure~\ref{fig:error_case_q73_sonnet}, demonstrated by Pixtral 12B under the \textit{V2T2C w/ SC} setting.}
\label{fig:error_case_q73_pixtral_12b}
\end{figure*}

\begin{figure*}[t]
\centering
\includegraphics[width=\linewidth]{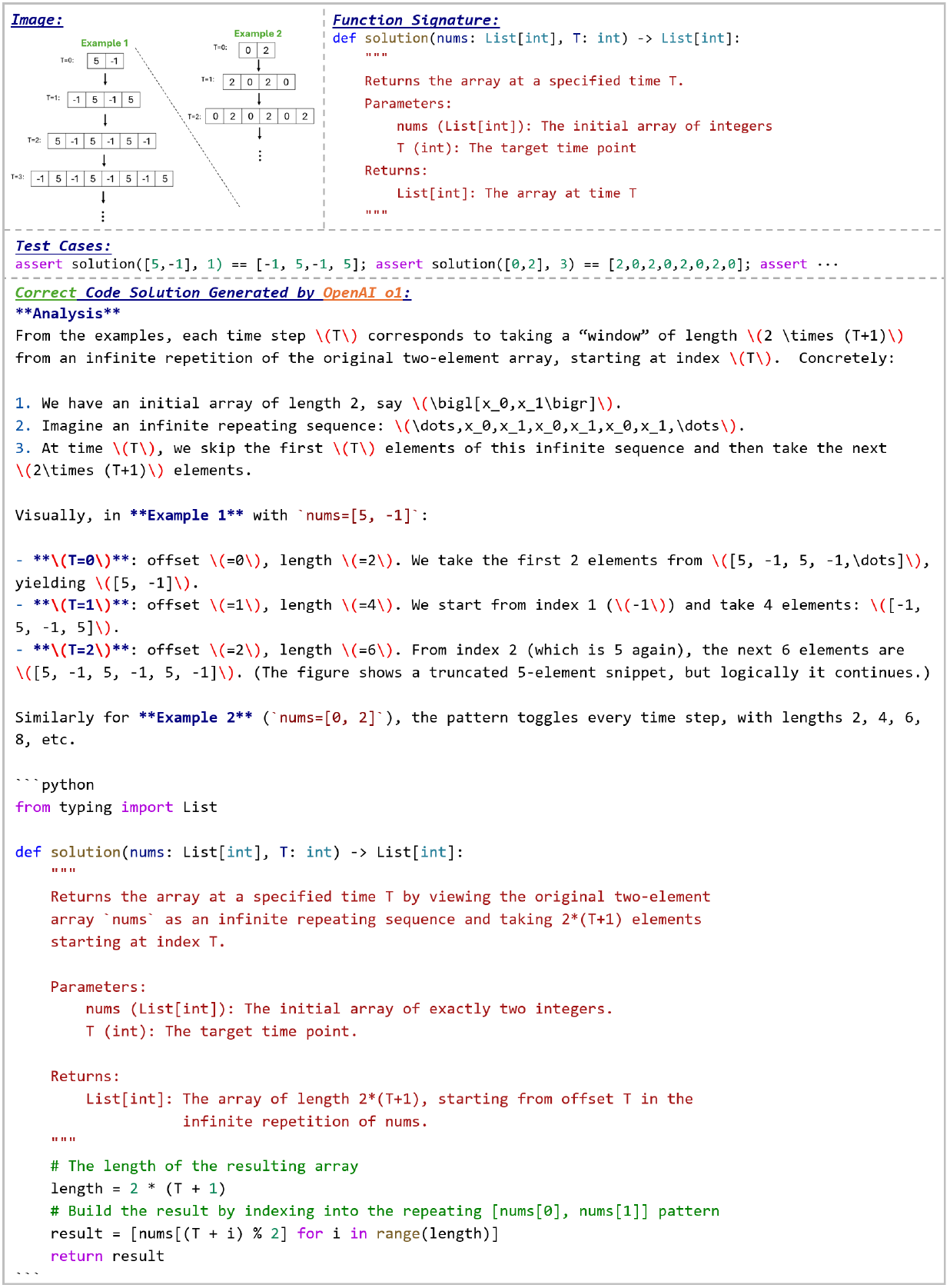}
\vspace{-0.5em}
\caption{Example case on the same task in Figure~\ref{fig:error_case_q73_sonnet}, demonstrated by OpenAI o1 under the \textit{V2C w/ CoT} setting.}
\label{fig:error_case_q73_o1}
\end{figure*}

\begin{figure*}[t]
\centering
\includegraphics[width=0.9\linewidth]{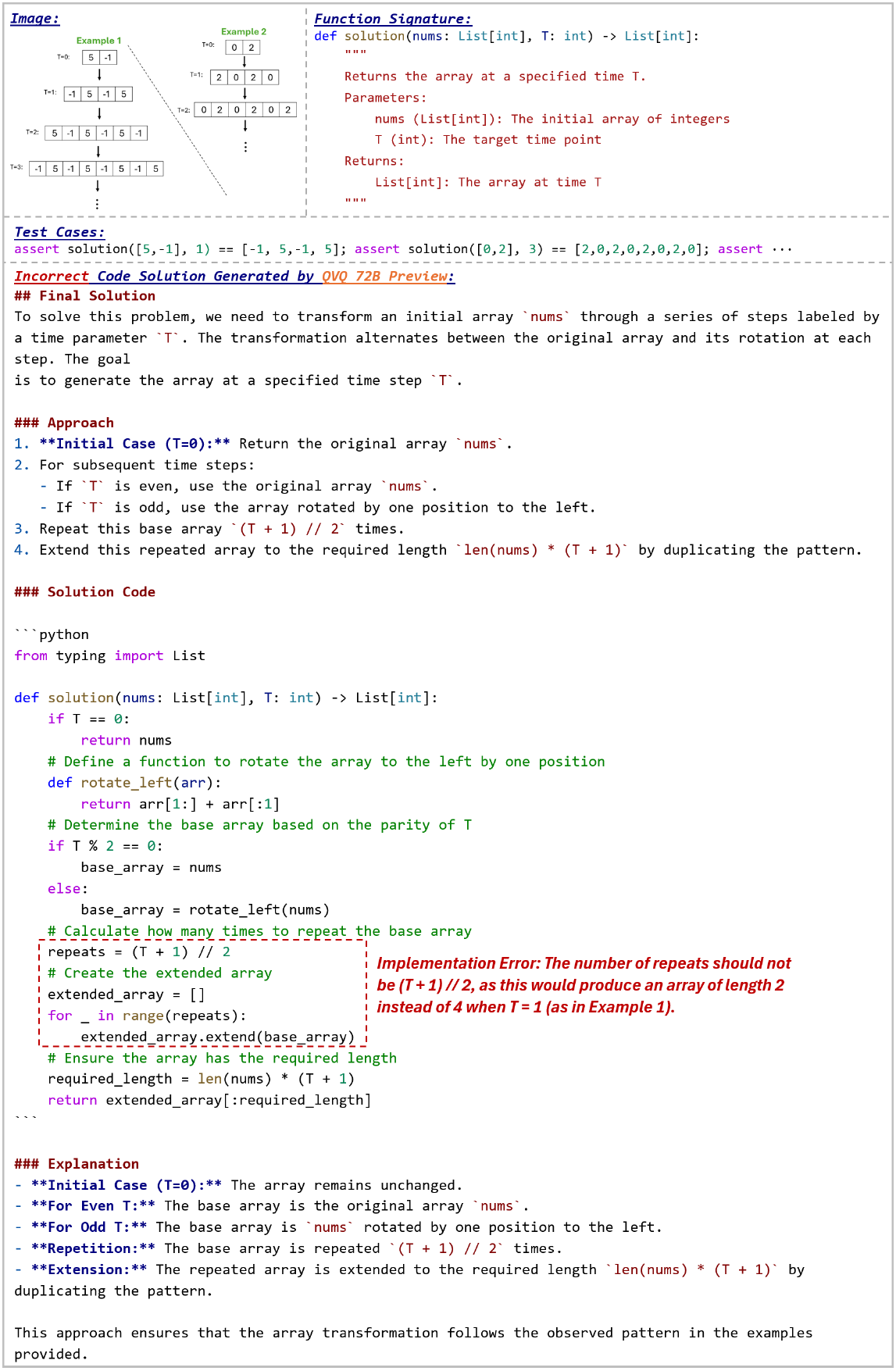}
\vspace{-0.5em}
\caption{Example case on the same task in Figure~\ref{fig:error_case_q73_sonnet}, demonstrated by QVQ 72B Preview under the \textit{V2C w/ CoT} setting.}
\label{fig:error_case_q73_qvq}
\end{figure*}

\begin{figure*}[t]
\centering
\includegraphics[width=\linewidth]{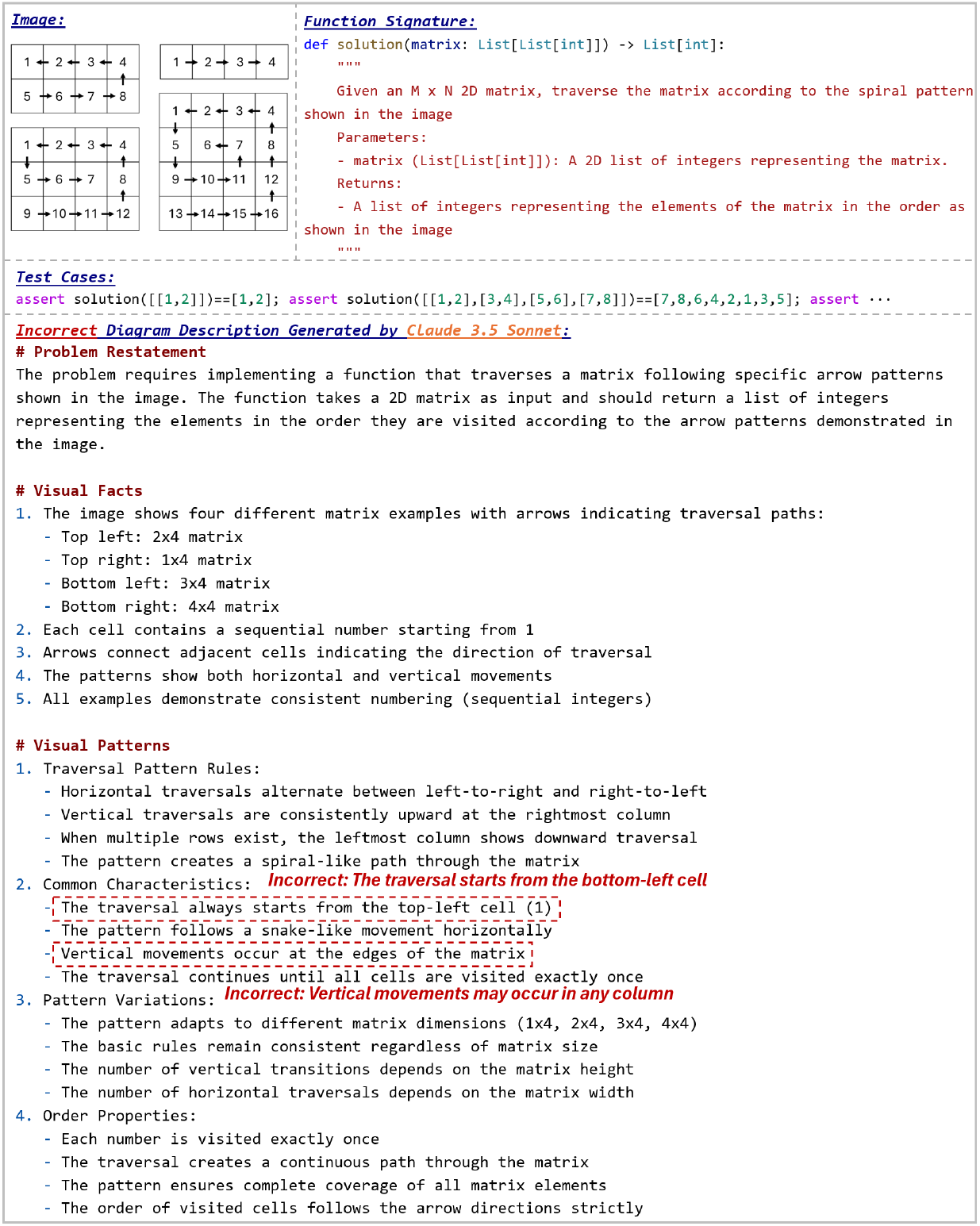}
\vspace{-0.5em}
\caption{Error case illustrating LMMs' difficulties in recognizing and reasoning about complex geometric arrangements, particularly spiral and circular patterns. The case shown for Claude 3.5 Sonnet is under the \textit{V2T2C w/ SC} setting.}
\label{fig:error_case_q18_sonnet}
\end{figure*}

\begin{figure*}[t]
\centering
\includegraphics[width=\linewidth]{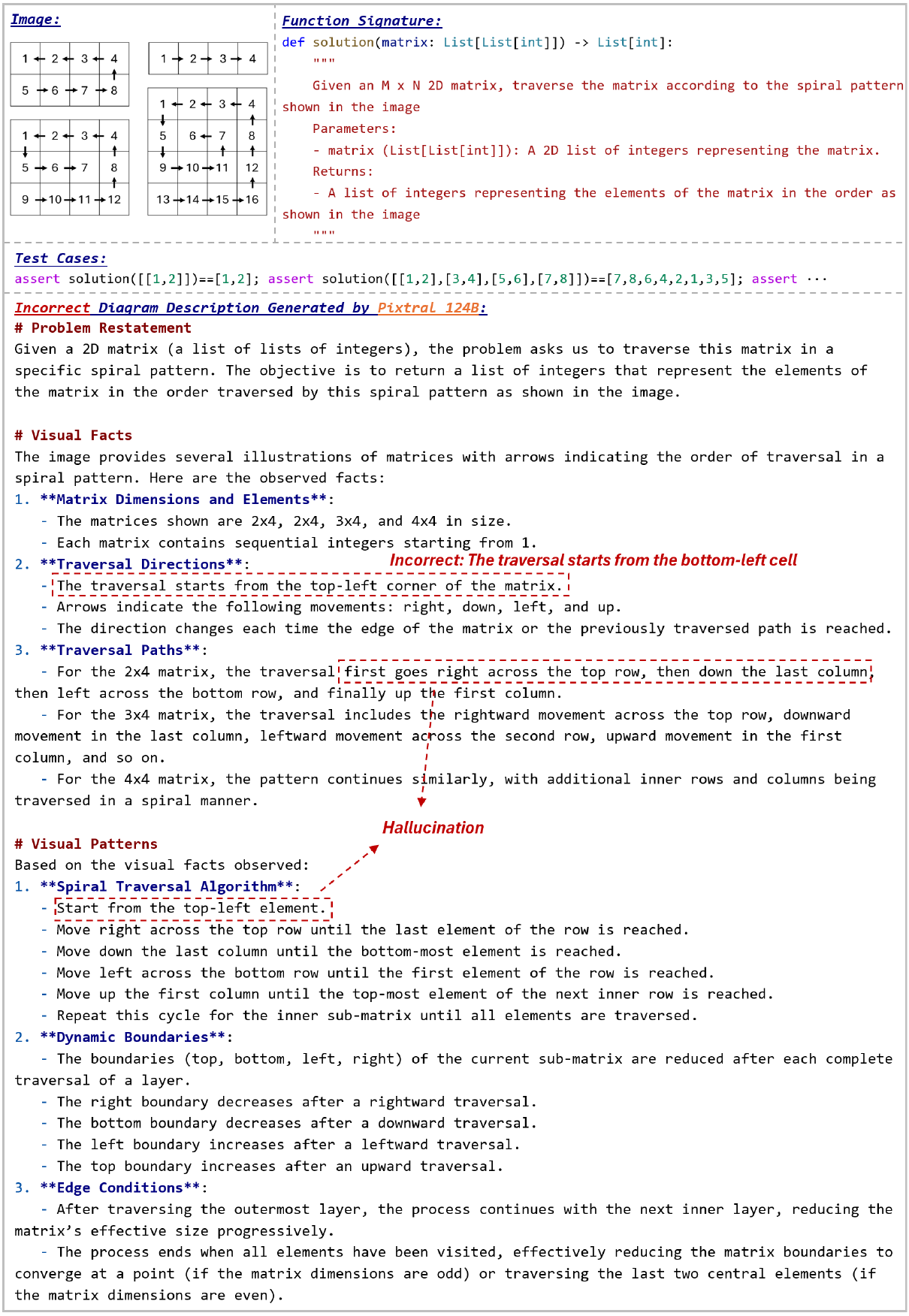}
\vspace{-0.5em}
\caption{Example case on the same task in Figure~\ref{fig:error_case_q18_sonnet}, demonstrated by Pixtral 124B under the \textit{V2T2C w/ SC} setting.}
\label{fig:error_case_q18_pixtral_large}
\end{figure*}

\begin{figure*}[t]
\centering
\includegraphics[width=\linewidth]{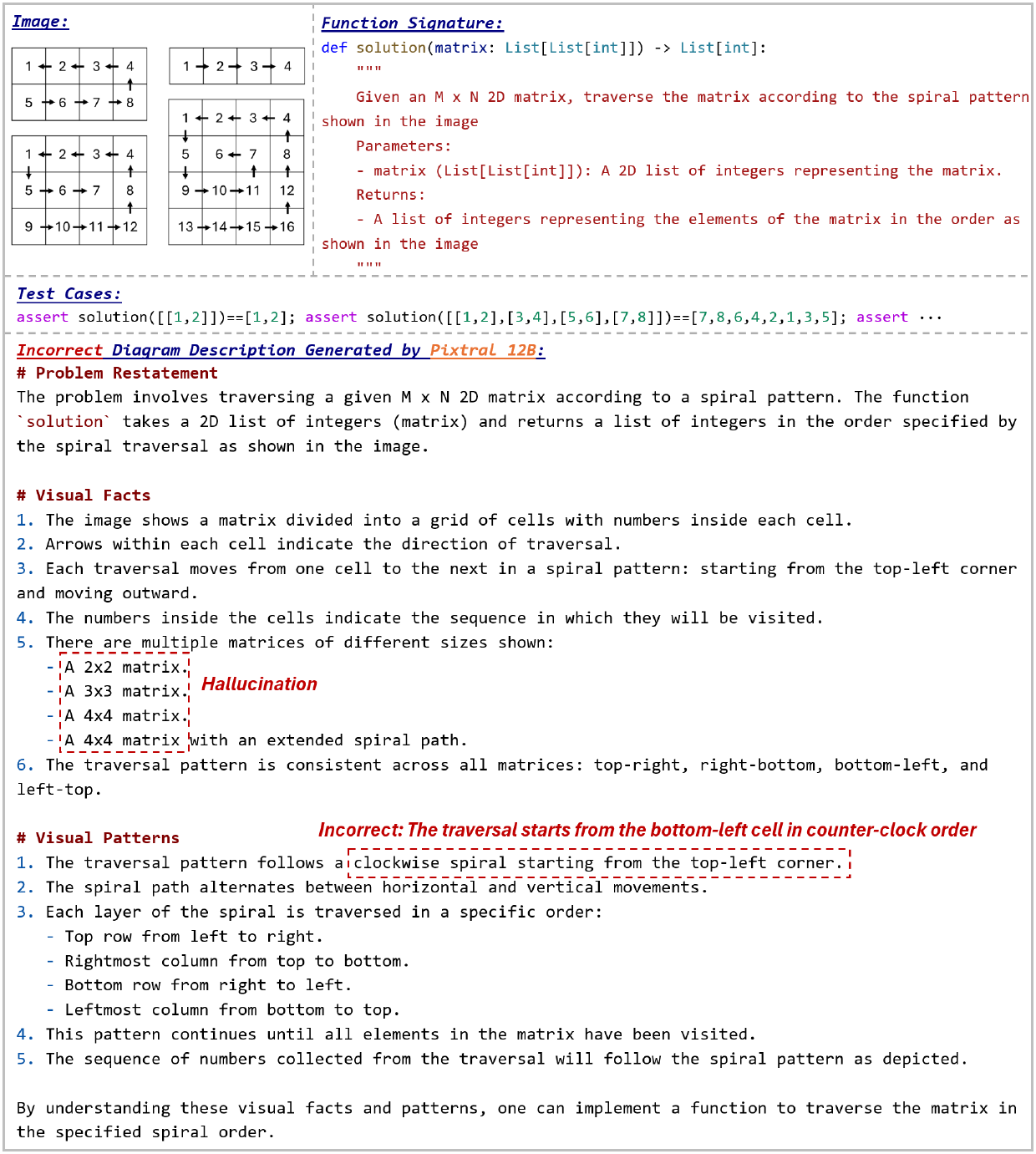}
\vspace{-0.5em}
\caption{Example case on the same task in Figure~\ref{fig:error_case_q18_sonnet}, demonstrated by Pixtral 12B under the \textit{V2T2C w/ SC} setting.}
\label{fig:error_case_q18_pixtral_12b}
\end{figure*}

\begin{figure*}[t]
\centering
\includegraphics[width=\linewidth]{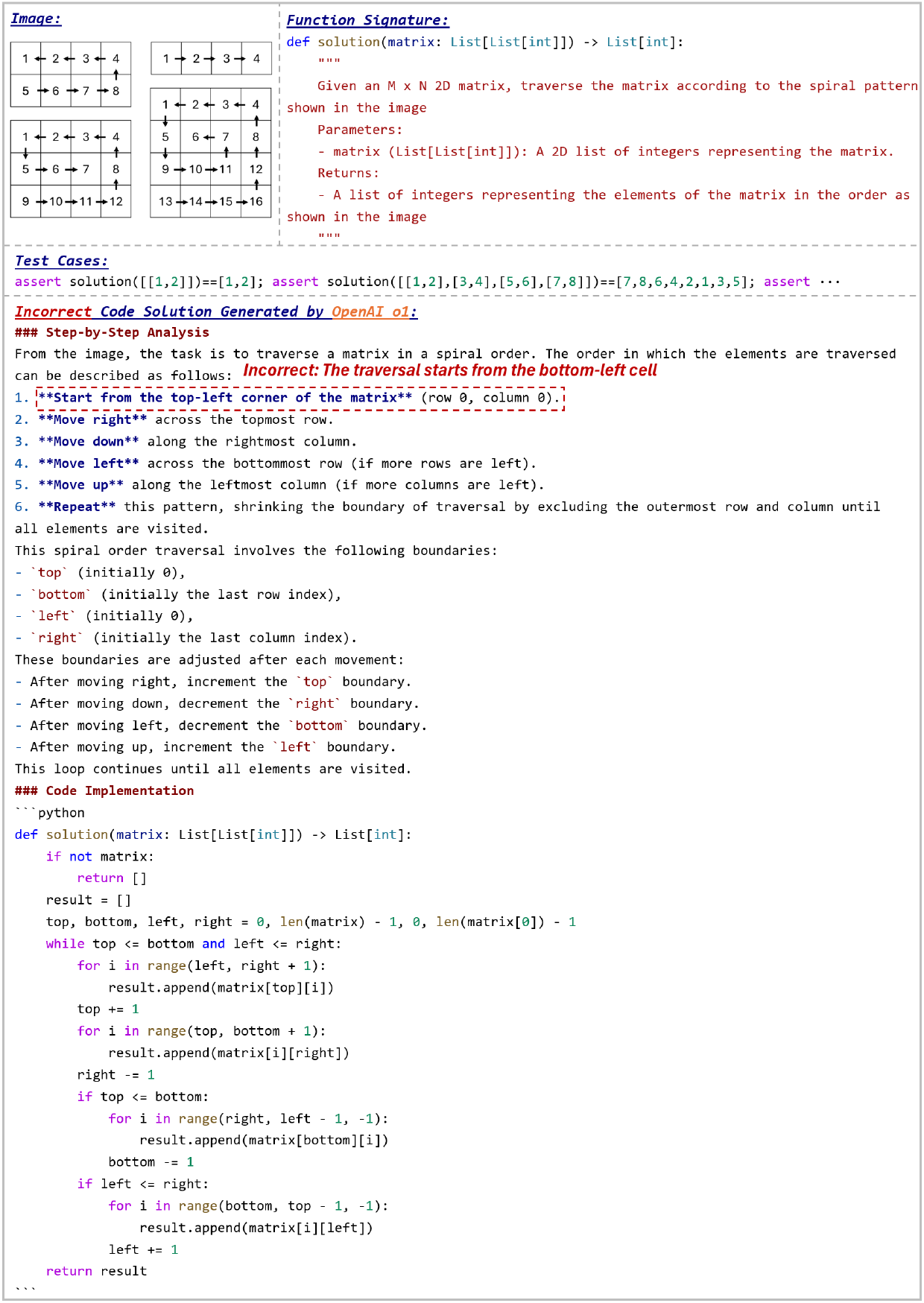}
\vspace{-0.5em}
\caption{Example case on the same task in Figure~\ref{fig:error_case_q18_sonnet}, demonstrated by OpenAI o1 under the \textit{V2C w/ CoT} setting.}
\label{fig:error_case_q18_o1}
\end{figure*}

\begin{figure*}[t]
\centering
\includegraphics[width=\linewidth]{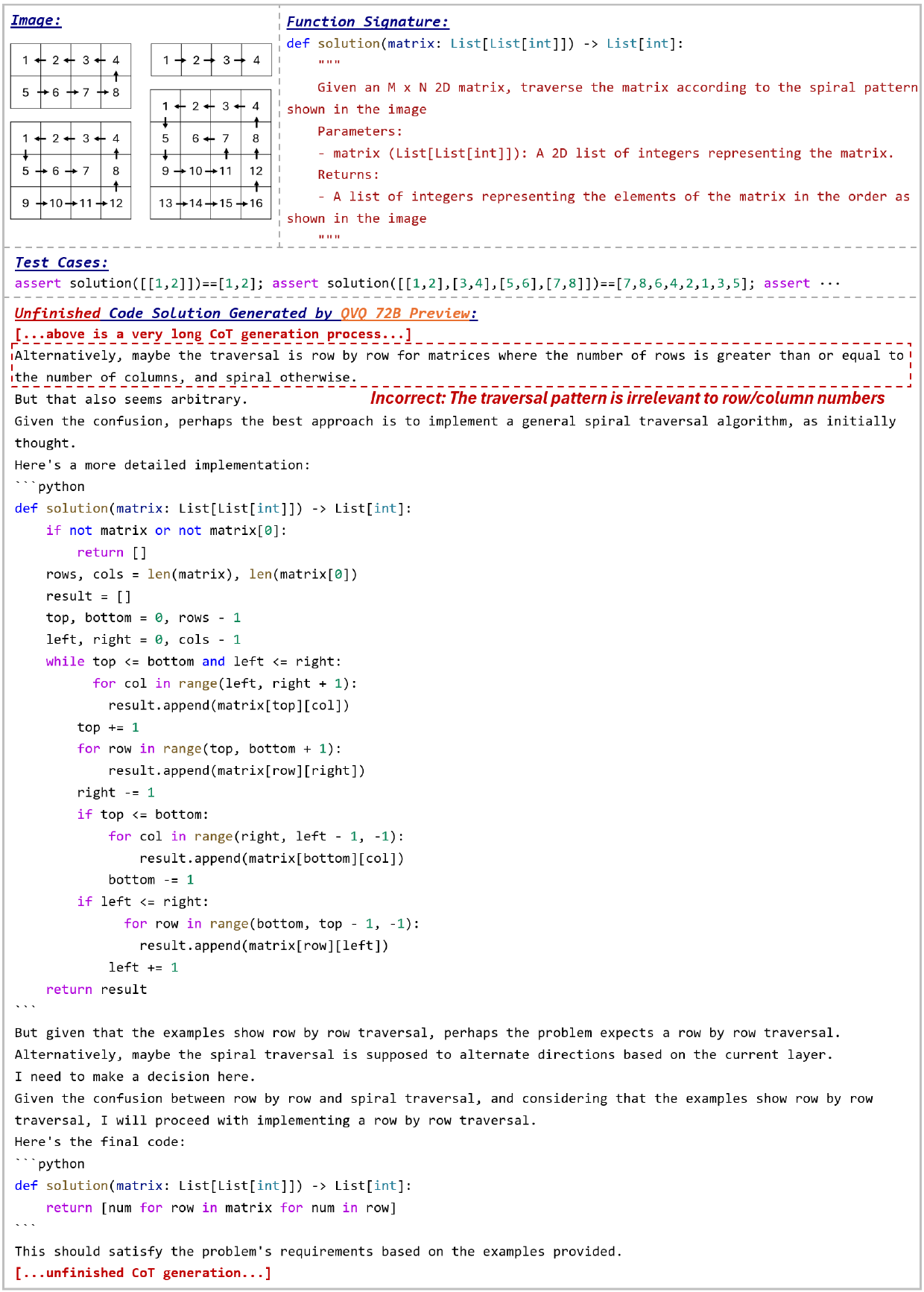}
\vspace{-0.5em}
\caption{Example case on the same task in Figure~\ref{fig:error_case_q18_sonnet}, demonstrated by QVQ 72B Preview under the \textit{V2C w/ CoT} setting.}
\label{fig:error_case_q18_qvq}
\end{figure*}

When examining specific capability aspects, we find that LMMs exhibit notable difficulty with diagrams involving dynamic patterns such as Spirals, Circular Arrangements, and Zigzags. Additionally, tasks requiring spatial transformations like Stacking, Translation, and Splitting, as well as mathematical operations such as Sorting and Absolute Value computations, pose significant challenges. We illustrate concrete error cases that highlight these challenges in Figure~\ref{fig:error_case_q63_sonnet} to~\ref{fig:error_case_q18_qvq}.

\subsection{Error Analysis by Task Difficulty}
We also investigate the correlation between LMM performance and task difficulty, using two key metrics. The first metric is the cyclomatic complexity~\citep{gill1991cyclomatic} of the human-annotated solution code for each task, which reflects the complexity of programming logic. The second metric is the token length of the human-annotated diagram descriptions, which indicates the difficulty of understanding the diagram from a textual perspective. These two metrics represent human-perceived difficulty in both visual comprehension and programmatic reasoning.

For measuring LMM performance, we use the averaged pass@3 score of the top five proprietary models under the \textit{V2T2C w/ SC} setting, with correlation results presented in Figure~\ref{fig:error_analysis_correlation}. Interestingly, the results suggest that LMM performance has little correlation with human-perceived difficulty, either in coding complexity or in visual description length, except for tasks with very high programming complexity or exceptionally long diagram descriptions.

Through a detailed case study, we find that many tasks in \ourbench are relatively easy for humans but remain challenging for LMMs, primarily because these models struggle to comprehend diagrams at a fundamental level. This limitation stems from their lack of basic visual perception and reasoning abilities, making it difficult to develop a precise metric that accurately captures LMM-perceived difficulty in our benchmark.

\begin{figure*}[t]
\centering
\includegraphics[width=\linewidth]{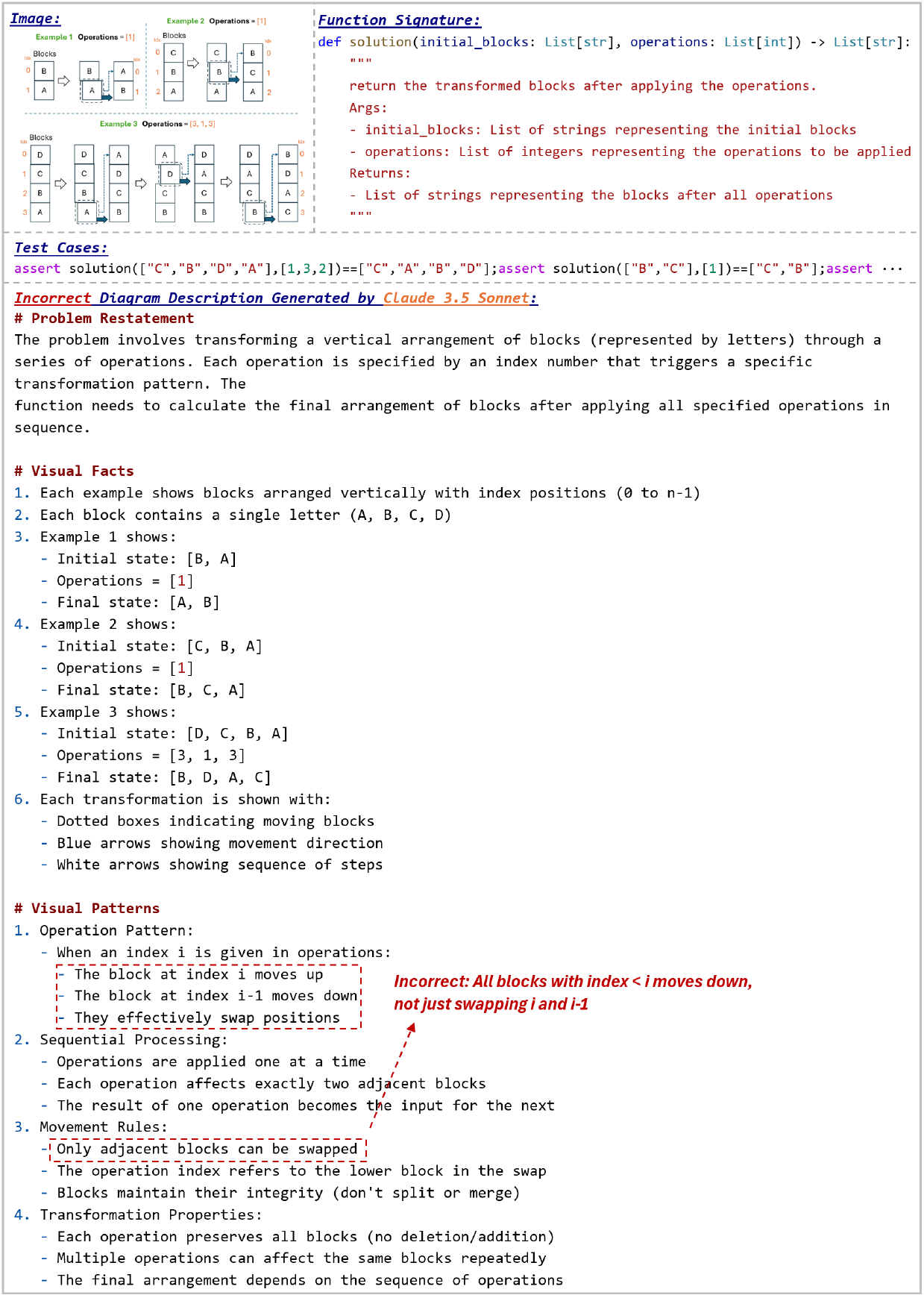}
\vspace{-0.5em}
\caption{Error case highlighting the gap between human intuition and LMM performance on seemingly straightforward visual reasoning tasks. The case shown for Claude 3.5 Sonnet is under the \textit{V2T2C w/ SC} setting.}
\label{fig:error_case_q84_sonnet}
\end{figure*}

\begin{figure*}[t]
\centering
\includegraphics[width=\linewidth]{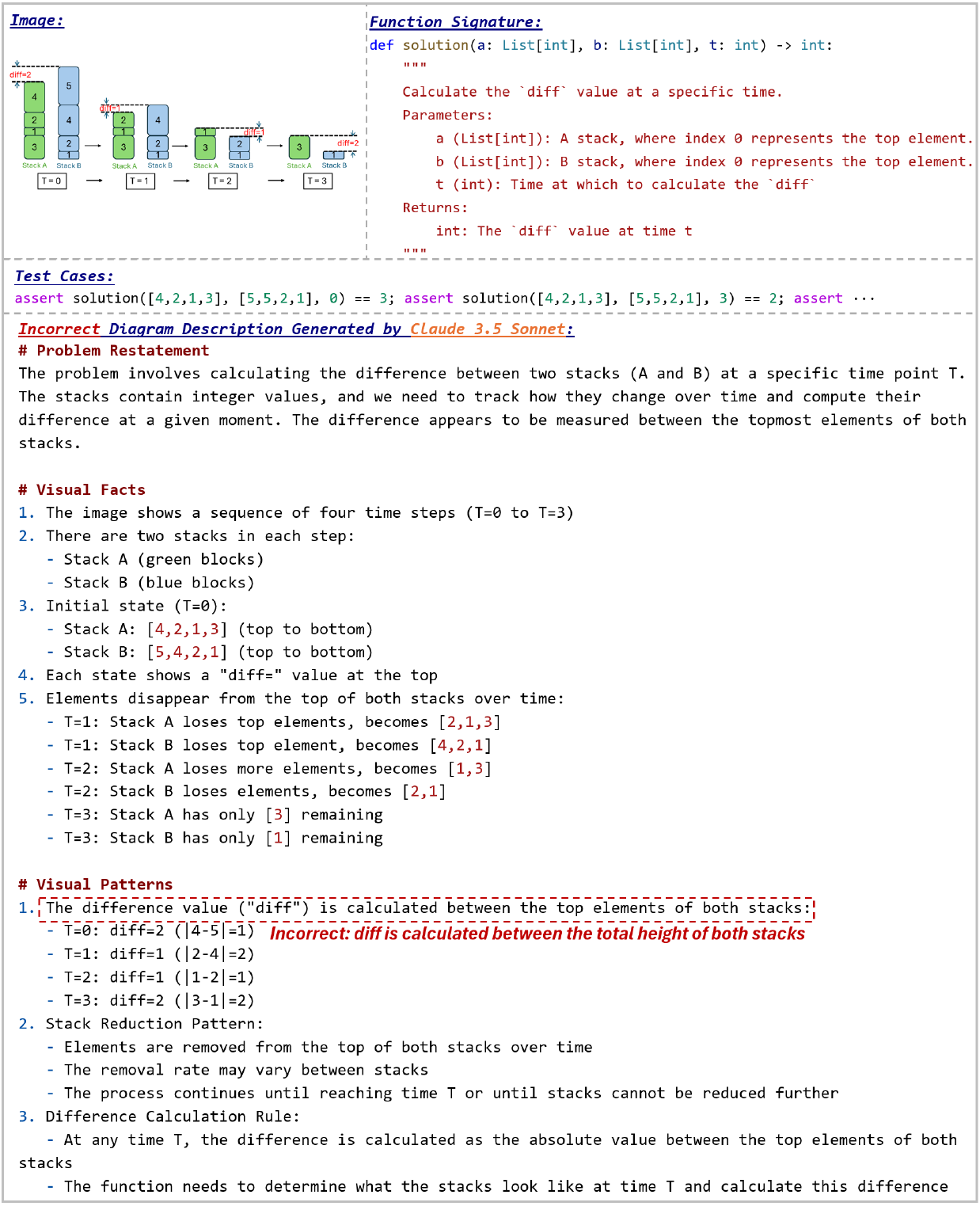}
\vspace{-0.5em}
\caption{Another case highlighting the gap between human intuition and LMM performance on seemingly straightforward visual reasoning tasks. The case shown for Claude 3.5 Sonnet is under the \textit{V2T2C w/ SC} setting.}
\label{fig:error_case_q47_sonnet}
\end{figure*}

The error cases in Figures~\ref{fig:error_case_q84_sonnet} and~\ref{fig:error_case_q47_sonnet} further illustrate this challenge. Tasks that appear trivial to humans often prove insurmountable for even the top-performing LMMs, highlighting the fundamental gap in their ability to interpret diagrams and reason visually.

\section{More Discussion on MMCode}
\label{appendix:mmcode}
MMCode~\citep{li2024mmcode} presents a coding dataset for evaluating LMMs' algorithmic problem-solving capabilities in visual contexts, comprising 3.5k questions crawled from competitive programming platforms. However, as explained in Appendix~\ref{appendix:data_collection}, the visual content in most coding challenges is redundant, with image information largely inferrable from textual descriptions. This redundancy is evident in MMCode's reported results, where performance on "language-only" inputs closely matches that of "vision + language" inputs.
In contrast, \ourbench is specifically designed to evaluate visual understanding and reasoning capabilities rather than general coding proficiency. Our benchmark ensures that visual context is integral to problem-solving. Experiments with five proprietary LMMs demonstrate a striking contrast: while providing only function signatures without diagrams or diagram descriptions results in 0\% pass rates across all models, the same models achieve over 90\% pass rates (Figure~\ref{fig:gt_description_to_code}) when given human-annotated diagram descriptions. This dramatic performance difference confirms the essential role of visual information in \ourbench. Furthermore, our difficulty analysis (Figure~\ref{fig:benchmark_statistics}) shows that the coding tasks maintain moderate complexity, enabling a focused assessment of visual reasoning abilities. Our evaluation pipeline also introduces a two-stage code generation process, allowing LMMs with lower coding proficiency to generate diagram descriptions while delegating the coding implementation to more capable models. These deliberate design choices clearly distinguish \ourbench from MMCode by placing visual reasoning at the forefront of evaluation.

\section{Other Considerations}
\paragraph{Environmental Considerations:} 
Our benchmark's challenging tasks typically require larger-sized multimodal models, which raises environmental concerns regarding computational costs. However, we believe the solution lies in improving training efficiency rather than simply scaling up model size. Our future work will focus on developing resource-efficient training methods while maintaining performance on our benchmark.

\paragraph{Language Coverage:}
Currently, our benchmark primarily focuses on English and Python, which may appear limiting. This choice was deliberate, as these languages are most prevalent in LMMs' training data and best demonstrate their capabilities. While this focus allows for deeper analysis, we acknowledge the importance of linguistic diversity. Our annotation pipeline is language-agnostic and can be extended to other programming languages in future iterations.

\paragraph{License and Distribution:}
Our benchmark consists of manually created tasks, drawing inspiration from Stack Overflow discussions and the MMCode dataset (which includes problems from platforms like Codeforces and LeetCode). We intend to distribute our code and data under a research-only license \textit{Creative Commons Non-Commercial (CC BY-NC)} to promote academic advancement.

\paragraph{Data Privacy and Protection:}
We can conclusively confirm that our dataset contains no personally identifiable information. All tasks were created from scratch by our team, with careful attention to privacy considerations. We maintained strict protocols during the creation process to ensure no sensitive information was included.

\paragraph{Computing Infrastructure:}
Our experimental setup utilized a computing node equipped with 8 NVIDIA A800 GPUs, primarily for LMM inference. Despite running multiple inference passes (1 greedy decode and 6 repeated sampling), the computational overhead remained manageable due to our focused dataset of 253 high-quality tasks.

\paragraph{Demographic of Annotators:}
The annotation team consisted of four highly qualified individuals -- postgraduate and doctoral students specializing in computer science, each with over four years of Python programming experience. Their participation is entirely voluntary and research-motivated, with no monetary compensation involved. All annotators explicitly consented to participate in this academic endeavor. While this arrangement worked well for our academic setting, we acknowledge that paid annotation might be necessary for larger-scale or commercial projects.

\end{document}